\pdfoutput=1
\documentclass[twoside,11pt]{article}

\usepackage{jmlr2e}

\usepackage[usenames, dvipsnames]{color}
\usepackage{amsmath,amsfonts,amssymb}
\usepackage{graphicx} 
\usepackage{cancel}
\usepackage{verbatim}
\usepackage{url}
\usepackage{xcolor}
\usepackage{multirow}
\usepackage{caption}
\usepackage{subcaption}
\usepackage{algpseudocode}
\usepackage{algorithm}
\usepackage{cancel}
\usepackage{mathtools}
\usepackage{pdflscape}
\usepackage[backgroundcolor=white,linecolor=red,bordercolor=white,textsize=tiny,textwidth=10mm]{todonotes}

\algnewcommand{\Initialize}[1]{%
	\State \textbf{Initialize:}
	\Statex \hspace*{\algorithmicindent}\parbox[t]{.8\linewidth}{\raggedright #1}
}

\graphicspath{{./figures/}}

\usepackage{bookmark}
\usepackage{cleveref}
\usepackage{hyperref}       

\usepackage[acronym]{glossaries}

\newglossaryentry{rmse}{%
	name={rmse},%
	description={root mean squared error},%
	first={root mean squared errors (RMSE)},%
	firstplural={root mean squared errors (RMSEs)},%
	text={RMSE},%
	plural={RMSEs}
}

\newglossaryentry{nll}{%
	name={nll},%
	description={negative log loss},%
	first={negative log loss (NLL)},%
	firstplural={negative log loss (NLLs)},%
	text={NLL},%
	plural={NLLs}
}

\newglossaryentry{gp}
{%
  name={GP},%
  description={Gaussian Process},%
  first={Gaussian Process (GP)},%
  firstplural={Gaussian Processes (GPs)},%
  text={GP}%
}

\newglossaryentry{dgp}
{%
  name={DGP},%
  description={deep Gaussian Process},%
  first={deep Gaussian Process (DGP)},%
  firstplural={deep Gaussian Processes (DGPs)},%
  text={DGP}%
}

\newglossaryentry{mlp}
{%
	name={MLP},%
	description={multi layer perceptron},%
	first={multi layer perceptron (MLP)},%
	firstplural={multi layer perceptrons (MLPs)},%
	text={MLP}%
}

\newglossaryentry{gplvm}
{%
  name={GPLVM},%
  description={Gaussian Process Latent Variable Model},%
  first={Gaussian Process Latent Variable Model (GPLVM)},%
  firstplural={Gaussian Process Latent Variable Models (GPLVMs)},%
  text={GPLVM}%
}

\newglossaryentry{gpssm}
{%
  name={GPSSM},%
  description={Gaussian Process State Space Model},%
  first={Gaussian Process State Space Model (GPSSM)},%
  firstplural={Gaussian Process State Space Models (GPSSMs)},%
  text={GPSSM}%
}

\newglossaryentry{vfe}
{%
  name={VFE},%
  description={Variational Free Energy},%
  first={variational free energy (VFE)},%
  text={VFE}%
}

\newglossaryentry{ep}
{%
  name={EP},%
  description={Expectation Propagation},%
  first={Expectation Propagation (EP)},%
  text={EP}%
}

\newcommand{\kuu}{\mathbf{K}_\mathbf{uu}}

\newcommand{\norm}{\mathcal{N}}
\newcommand{\normnat}{\tilde{\mathcal{N}}}

\newcommand{\xvec}{\mathbf{x}}
\newcommand{\fvec}{\mathbf{f}}
\newcommand{\gvec}{\mathbf{g}}
\newcommand{\uvec}{\mathbf{u}}
\newcommand{\ftest}{f_{\ne \uvec, \fvec}}

\newcommand{\yvec}{\mathbf{y}}
\newcommand{\mvec}{\mathbf{m}}

\newcommand{\Rvec}{\mathbf{R}}
\newcommand{\Svec}{\mathbf{S}}

\newcommand{\dd}{\mathrm{d}}

\def\E{{\mathbb E}}

\usepackage{enumitem}

\newenvironment{enumerate*}%
  {\begin{enumerate}[leftmargin=5mm]%
    \setlength{\itemsep}{1pt}%
    \setlength{\parskip}{1pt}}%
  {\end{enumerate}}

\allowdisplaybreaks[4]

\newcommand*{\mathcolor}{}
\def\mathcolor#1#{\mathcoloraux{#1}}
\newcommand*{\mathcoloraux}[3]{%
  \protect\leavevmode
  \begingroup
    \color#1{#2}#3%
  \endgroup
}

\makeatletter
\newcommand{\mypm}{\mathbin{\mathpalette\@mypm\relax}}
\newcommand{\@mypm}[2]{\ooalign{%
		\raisebox{.1\height}{$#1+$}\cr
		\smash{\raisebox{-.6\height}{$#1-$}}\cr}}
\makeatother

\newcommand{\argmin}{\operatornamewithlimits{argmin}}
\newcommand{\argmax}{\operatornamewithlimits{argmax}}

\ShortHeadings{Unifying Gaussian Process Approximations}{Bui, Yan and Turner}
\firstpageno{1}

\begin{document}

\title{A Unifying Framework for  Gaussian Process Pseudo-Point Approximations using Power Expectation Propagation}

\author{\name Thang D. Bui \email tdb40@cam.ac.uk\AND
		\name Josiah Yan \email josiah.yan@gmail.com\AND
		\name Richard E.~Turner \email ret26@cam.ac.uk\\
       \addr Computational and Biological Learning Lab, Department of Engineering\\
University of Cambridge, Trumpington Street, Cambridge, CB2 1PZ, UK}


\maketitle

\begin{abstract}

Gaussian processes (GPs) are flexible distributions over functions that enable high-level assumptions about unknown functions to be encoded in a parsimonious, flexible and general way. Although elegant, the application of GPs is limited by computational and analytical intractabilities that arise when  data are sufficiently numerous or when employing non-Gaussian models. Consequently, a wealth of GP approximation schemes have been developed over the last 15 years to address these key limitations. Many of these schemes employ a small set of pseudo data points to summarise the actual data. In this paper we develop a new pseudo-point approximation framework using Power Expectation Propagation (Power EP) that unifies a large number of these pseudo-point approximations. Unlike much of the previous venerable work in this area, the new framework is built on standard methods for approximate inference (variational free-energy, EP and Power EP methods) rather than employing approximations to the probabilistic generative model itself. In this way all of approximation is performed at `inference time' rather than at `modelling time' resolving awkward philosophical and empirical questions that trouble previous approaches. Crucially, we demonstrate that the new framework includes new pseudo-point approximation methods that outperform current approaches on regression and classification tasks.

%
%
%

\end{abstract}


\section{Introduction}

\glspl{gp} are powerful nonparametric distributions over continuous functions that are routinely deployed in probabilistic modelling for applications including regression and classification \citep{RasWil05}, representation learning \citep{Law05}, state space modelling \citep{WanFleHer05}, active learning \citep{HouHusGhaLen11}, reinforcement learning \citep{Dei10}, black-box optimisation \citep{SnoLarAda12}, and numerical methods \citep{MahHen2015}. \glspl{gp} have many elegant theoretical properties, but their use in probabilistic modelling is greatly hindered by analytic and computational intractabilities. A large research effort has been directed at this fundamental problem resulting in the development of a plethora of sparse approximation methods that can sidestep these intractabilities \citep{Csa02, CsaOpp02, SchTre02, SeeWilLaw03, QuiRas05, SneGha06, Sne07, NaiHol08, Tit09a, FigLaz09, AlvLueTitLaw10, QiAbdMin10, BuiTur14, FriCheRas14, Mch14, HenMatGha15, HerHer16, MatHenTurGha16}


This paper develops a general sparse approximate inference framework based upon Power Expectation Propagation (PEP) \citep{Min04} that unifies many of these approximations, extends them significantly, and provides improvements in practical settings. In this way, the paper provides a complementary perspective to the seminal review of \citet{QuiRas05} viewing sparse approximations through the lens of approximate {\it inference}, rather than approximate {\it generative models}.

The paper begins by reviewing several frameworks for sparse approximation focussing on the \gls{gp} regression and classification setting (\cref{sec:gprc}). It then lays out the new unifying framework and the relationship to existing techniques (\cref{sec:pep}). Readers whose focus is to understand the new framework might want to move directly to this section.
Finally, a thorough experimental evaluation is presented in section~\ref{sec:exp}.


\section{Pseudo-point Approximations for GP Regression and Classification\label{sec:gprc}}

This section provides a concise introduction to \gls{gp} regression and classification and then reviews several pseudo-point based sparse approximation schemes for these models. For simplicity, we first consider a supervised learning setting in which the training set comprises $N$ $D$-dimensional input and scalar output pairs $\{\xvec_n, y_n\}_{n=1}^{N}$ and the goal is to produce probabilistic predictions for the outputs corresponding to novel inputs. A non-linear function, $f(\xvec)$, can be used to parameterise the probabilistic mapping between inputs and outputs, $p(y_n | f,\xvec_n,\theta)$. Typical choices for the probabilistic mapping are Gaussian $p(y_n | f,\xvec_n,\theta) = \mathcal{N}(y_n; f(\xvec_n), \sigma_y^2)$ for the regression setting ($y_n \in \mathbb{R}$) and Bernoulli $p(y_n | f,\xvec_n,\theta) = \mathcal{B}(y_n; \Phi(f(\xvec_n)))$ with a sigmoidal link function $\Phi(f)$ for the binary classification setting ($y_n \in \{0,1\}$). Whilst it is possible to specify the non-linear function $f$ via an explicit parametric form, a more flexible and elegant approach employs a \gls{gp} prior over the functions directly, $p(f|\theta) = \mathcal{GP}(f;0, k_{\theta}(\cdot,\cdot))$, here assumed without loss of generality to have a zero mean-function and a covariance function $k_{\theta}(\xvec,\xvec')$. This class of probabilistic models has a joint distribution 
\begin{align}
p(f,\yvec | \theta ) = p (f|\theta) \prod_{n=1}^N p(y_n | f(\xvec_n),\theta)
\end{align} 
where we have collected the observations into the vector $\yvec$ and suppressed the inputs on the left hand side to lighten the notation.

This model class contains two potential sources of intractability. First, the possibly non-linear likelihood function can introduce analytic intractabilities that require approximation. Second, the \gls{gp} prior entails an $\mathcal{O}(N^3)$ complexity that is computationally intractable for many practical problems. These two types of intractability can be handled by combining standard approximate inference methods with pseudo-point approximations that summarise the full Gaussian process via $M$ pseudo data points leading to an $\mathcal{O}(NM^2)$ cost. The main approaches of this sort can be characterised in terms of two parallel frameworks that are described in the following sections.

\subsection{Sparse GP Approximation via Approximate Generative Models \label{sec:genmod}}

The first framework begins by constructing a new generative model that is similar to the original, so that inference in the new model might be expected to produce similar results, but which has a special structure that supports efficient computation. Typically this approach involves approximating the Gaussian process prior as it is the origin of the cubic cost. If there are analytic intractabilities in the approximate model, as will be the case in e.g.~classification or state-space models, then these will require approximate inference to be performed in the approximate model.

The seminal review by Qui{\~n}onero-Candela and Rasmussen \citep{QuiRas05} reinterprets a family of approximations in terms of this unifying framework. The \gls{gp} prior is approximated by identifying a small set of $M \le N$ pseudo-points $\uvec$, here assumed to be disjoint from the training function values $\fvec$ so that $f = \{ \uvec, \fvec, \ftest \}$. The \gls{gp} prior is then decomposed using the product rule
\begin{align}
p(f|\theta) = p(\uvec|\theta) p(\fvec|\uvec,\theta) p(\ftest|\fvec, \uvec,\theta).\end{align} 
Of central interest is the relationship between the pseudo-points and the training function values $p(\fvec|\uvec,\theta)  = \mathcal{N}(\fvec;\mathbf{K}_{\fvec\uvec} \mathbf{K}^{-1}_{\uvec\uvec} \uvec, \mathbf{D}_{\fvec\fvec})$ where $\mathbf{D}_{\fvec\fvec} = \mathbf{K}_{\fvec\fvec} - \mathbf{Q}_{\fvec\fvec}$ and $\mathbf{Q}_{\fvec\fvec} = \mathbf{K}_{\fvec\uvec} \mathbf{K}^{-1}_{\uvec\uvec} \mathbf{K}_{\uvec\fvec}$. Here we have introduced matrices corresponding to the covariance function's evaluation at the pseudo-input locations $\{ \mathbf{z}_m \}_{m=1}^M$, so that $[\mathbf{K}_{\uvec\uvec}]_{mm'} = k_{\theta}(\mathbf{z}_m,\mathbf{z}_{m'})$ and similarly for the covariance between the pseudo-input and data locations $[\mathbf{K}_{\uvec\fvec}]_{mn} = k_{\theta}(\mathbf{z}_m,\mathbf{x}_{n})$. Importantly, this term saddles learning with a cubic complexity cost.
Computationally efficient approximations can be constructed by simplifying these dependencies between the pseudo-points and the data function values $q(\fvec|\uvec,\theta) \approx p(\fvec|\uvec,\theta)$. In order to benefit from these efficiencies at prediction time as well, a second approximation is made whereby the pseudo-points form a bottleneck between the data function values and test function values $p(\ftest|\uvec,\theta) \approx p(\ftest|\fvec, \uvec,\theta)$. Together, the two approximations result in an approximate prior process,
 \begin{align}
q(f|\theta) = p(\uvec|\theta) q(\fvec|\uvec,\theta) p(\ftest|\uvec,\theta).
\end{align} 
We can now compactly summarise a number of previous approaches to GP approximation as special cases of the choice 
\begin{align}
q(\fvec|\uvec,\theta) = \prod_{b=1}^B  \mathcal{N}(\fvec_b;\mathbf{K}_{\fvec_b,\uvec} \mathbf{K}^{-1}_{\uvec\uvec} \uvec,\alpha \mathbf{D}_{\fvec_b,\fvec_b})
\end{align}
 where $b$ indexes $B$ disjoint blocks of data-function values. The Deterministic Training Conditional (DTC) approximation uses $\alpha \rightarrow 0$; the Fully Independent Training Conditional (FITC) approximation uses $\alpha =1$ and $B=N$; the Partially Independent Training Conditional (PITC) approximation uses $\alpha=1$ \citep{QuiRas05, SchTre02}. 
 
In a moment we will consider inference in the modified models, before doing so we note that it is possible to construct more flexible modified prior processes using the inter-domain approach that places the pseudo-points in a different domain from the data, defined by a linear integral transform  $g(z) = \int w(z,z')f(z') \text{d} z'$. Here the window $w(z,z')$ might be a Gaussian blur or a wavelet transform. The pseudo-points are now placed in the new domain $g = \{ \uvec, \gvec_{\ne \uvec}\}$ where they induce a potentially more flexible Gaussian process in the old domain $f$ through the linear transform \cite[see][for FITC]{FigLaz09}. The expressions in this section still hold, but the covariance matrices involving pseudo-points are modified to take account of the transform,
\begin{align}
[\mathbf{K}_{\uvec\uvec}]_{mm'} = \int w(\mathbf{z}_m,\mathbf{z}) k_{\theta}(\mathbf{z},\mathbf{z}') w(\mathbf{z}',\mathbf{z}_{m'}) \mathbf{d}\mathbf{z} \mathbf{d}\mathbf{z} ', \;\;
[\mathbf{K}_{\uvec\fvec}]_{mn} = \int w(\mathbf{z}_m,\mathbf{z}) k_{\theta}(\mathbf{z},\mathbf{x}_n)  \mathbf{d}\mathbf{z}.
\end{align}

Having specified modified prior processes, these can be combined with the original likelihood function to produce a new generative model. In the case of point-wise likelihoods we have
\begin{align}
q(\yvec,f|\theta) = q(f|\theta) \prod_{n=1}^N p(y_n| f(\xvec_n),\theta).
\end{align} 
Inference and learning can now be performed using the modified model using standard techniques. Due to the form of the new prior process, the computational complexity is $\mathcal{O}(NM^2)$ (for testing, $N$ becomes the number of test data points, assuming dependencies between the test-points are not computed).\footnote{It is assumed that the maximum size of the blocks is not greater than the number of pseudo-points $\text{dim}(\fvec_b) \le M$.} 
For example, in the case of regression, the posterior distribution over function values $f$ (necessary for inference and prediction) has a simple analytic form
\begin{align}
q(f |\yvec,\theta) = \mathcal{GP}(f;\mu_{ f | \yvec},\Sigma_{f | \yvec}),\;\;
\mu_{f | \yvec}  = \mathbf{Q}_{f \fvec} \overline{\mathbf{K}}_{\mathbf{ff}}^{-1} \yvec, \;\;
\Sigma_{f | \yvec} = \mathbf{K}_{ff} - \mathbf{Q}_{f \fvec} \overline{\mathbf{K}}_{\mathbf{ff}}^{-1} \mathbf{Q}_{\fvec f} \label{eqn:fitc_post}
\end{align}
where $\overline{\mathbf{K}}_{\mathbf{ff}} = \mathbf{Q}_{\mathbf{ff}} + \mathrm{blkdiag} (\{ \alpha_b \mathbf{D}_{\mathbf{f}_b\mathbf{f}_b} \}_{b=1}^B) + \sigma_y^2\mathrm{I}$ and $\mathrm{blkdiag}$ builds a block-diagonal matrix from its inputs. One way of understanding the origin of the computational gains is that the new generative model corresponds to a form of factor analysis in which the $M$ pseudo-points determine the $N$ function values at the observed data (as well as at potential test locations) via a linear Gaussian relationship. This results in low rank (sparse) structure in $\overline{\mathbf{K}}_{\mathbf{ff}}$ that can be exploited through the matrix inversion and determinant lemmas.
 In the case of regression, the new model's marginal likelihood also has an analytic form that allows the hyper-parameters, $\theta$, to be learned via optimisation
\begin{align}
	\small
    \log q(\yvec|\theta) &= - \frac{N}{2}\log(2\pi) - \frac{1}{2}\log|\overline{\mathbf{K}}_{\mathbf{ff}}| - \frac{1}{2}\mathbf{y}^{\intercal} \overline{\mathbf{K}}_{\mathbf{ff}}^{-1} \mathbf{y}.\label{eqn:fitc_energy}
\end{align}

The approximate generative model framework has attractive properties. The cost of inference, learning, and prediction has been reduced from $\mathcal{O}(N^3)$ to $\mathcal{O}(NM^2)$ and in many cases accuracy can be maintained with a relatively small number of pseudo-points. The pseudo-point input locations can be optimised by maximising the new model's marginal likelihood \citep{SneGha06}. When $M=N$ and the pseudo-points and observed data inputs coincide, then FITC and PITC are exact which appears reassuring. However, the framework is philosophically challenging as the elegant separation of model and (approximate) inference has been lost. Are we allowed in an online inference setting, for example, to add new pseudo-points as more data are acquired and the complexity of the underlying function is revealed? This seems sensible, but effectively changes the modelling assumptions as more data are seen. Devout Bayesians might then demand that we perform model averaging for coherence. Similarly, if the pseudo-input locations are optimised, the principled non-parametric model has suddenly acquired $MD$ parameters and with them all of the concomitant issues of parametric models including overfitting and optimisation difficulties \citep{BauWilRas16}. As the pseudo-inputs are considered part of the model, the Bayesians might then suggest that we place priors over the pseudo-inputs and perform full blown probabilistic inference over them. 

These awkward questions arise because the generative modelling interpretation of pseudo-data entangles the assumptions made about the data with the approximations required to perform inference. Instead, the modelling assumptions (which encapsulate prior understanding of the data) should remain decoupled from inferential assumptions (which leverage structure in the posterior for tractability). In this way pseudo-data should be introduced when we seek to perform computationally efficient approximate inference, leaving the modelling assumptions unchanged as we refine and improve approximate inference. Indeed, even under the generative modelling perspective, for analytically intractable likelihood functions an additional approximate inference step is required, begging the question; why not handle computational and analytic intractabilities together at inference time?  


%

\subsection{Sparse GP Approximation via Approximate Inference: VFE}\label{sec:VFE}

The approximate generative model framework for constructing sparse approximations is philosophically troubling. In addition, learning pseudo-point input locations via optimisation of the model likelihood can perform poorly e.g.~for DTC it is prone to overfitting even for $M \ll N$ \citep{Tit09a}. This motivates a more direct approach that commits to the true generative model and performs all of the necessary approximation at inference time.

Perhaps the most well known approach in this vein is Titsias's beautiful sparse \gls{vfe} method \citep{Tit09a}. The original presentation of this work employs finite variable sets and an augmentation trick that arguably obscures its full elegance. Here instead we follow the presentation in \citet{MatHenTurGha16} and lower bound the marginal likelihood using a distribution $q(f)$ over the entire infinite dimensional function,
\begin{align}
\log p(\yvec|\theta) = \log \int  p(\yvec,f|\theta) \text{d}f \ge   \int q(f) \log \frac {p(\yvec,f|\theta)} {q(f)} \text{d}f = \E_{q(f)} \left[ \log \frac {p(\yvec,f|\theta)} {q(f)} \right]= \mathcal{F}(q, \theta). \nonumber
\end{align}
The \gls{vfe} bound can be written as the difference between the model log-marginal likelihood and the KL divergence between the variational distribution and the true posterior $\mathcal{F}(q, \theta) = \log p(\yvec|\theta) - \mathrm{KL} ( q(f) || p(f | \yvec,\theta) )$. The bound is therefore saturated when $q(f) = p(f|\yvec,\theta)$, but this is intractable. Instead, pseudo-points are made explicit, $f = \{ \uvec, f_{\ne \uvec} \}$, 
and an approximate posterior distribution used of the following form $q(f) = q(\uvec, f_{\ne \uvec}|\theta) = p(f_{\ne \uvec} | \uvec,\theta) q(\uvec)$. Under this approximation, the set of variables $f_{\ne \uvec}$ do not experience the data directly, but rather only through the pseudo-points, as can be seen by comparison to the true posterior $p(f | \yvec,\theta) = p(f_{\ne \uvec} | \yvec, \uvec,\theta) p(\uvec|\yvec,\theta)$. Importantly, the form of the approximate posterior causes a cancellation of the prior conditional term, which gives rise to a bound with $\mathcal{O}(NM^2)$ complexity,
\begin{align}
\mathcal{F}(q, \theta)&= \E_{q(f|\theta)} \left[ \log  \frac {p(\yvec|f,\theta) \cancel{p(f_{\neq \uvec}|\uvec,\theta)} p(\uvec|\theta)} {\cancel{p(f_{\neq \uvec}|\uvec,\theta)} q(\uvec)} \right] \nonumber\\
&=  \sum_n \E_{q(f|\theta)} \left[ \log p(y_n| \mathrm{f}_n,\theta) \right] - \mathrm{KL} ( q(\uvec) || p(\uvec|\theta) ).\nonumber
\end{align}
For regression with Gaussian observation noise, the calculus of variations can be used to find the optimal approximate posterior Gaussian process over pseudo-data $q^{\text{opt}}(f|\theta) = p(f_{\ne \uvec} | \uvec,\theta) q^{\text{opt}}(\uvec)$ which has the form
\begin{align}
q^{\text{opt}}(f|\theta) = \mathcal{GP}(f;\mu_{ f | \yvec},\Sigma_{f | \yvec}),\;\;
\mu_{f | \yvec}  = \mathbf{Q}_{f \fvec} \tilde{\mathbf{K}}_{\mathbf{ff}}^{-1} \yvec, \;\;
\Sigma_{f | \yvec} = \mathbf{K}_{ff} - \mathbf{Q}_{f \fvec} \tilde{\mathbf{K}}_{\mathbf{ff}}^{-1} \mathbf{Q}_{\fvec f} \label{eqn:vfe_post}
 \end{align}
%
where $\tilde{\mathbf{K}}_{\mathbf{ff}} =  \mathbf{Q}_{\mathbf{ff}}  + \sigma_y^2\mathrm{I}$.  This process is identical to that recovered when performing exact inference under the DTC approximate regression generative model \citep{Tit09a} (see equation \ref{eqn:fitc_post} as $\alpha \rightarrow 0$). In fact DTC was originally derived using a related KL argument \citep{Csa02,SeeWilLaw03}. The optimised free-energy is
%
\begin{align}
    \mathcal{F}(q^{\text{opt}}, \theta) &= - \frac{N}{2}\log(2\pi) - \frac{1}{2}\log|\tilde{\mathbf{K}}_{\mathbf{ff}}| - \frac{1}{2}\mathbf{y}^{\intercal} \tilde{\mathbf{K}}_{\mathbf{ff}}^{-1} \mathbf{y} -  \frac{1}{2\sigma^2_y} \mathrm{trace} (\mathbf{K}_{\fvec\fvec} -\mathbf{Q}_{\mathbf{ff}})\label{eqn:vfe_energy}.
\end{align}
Notice that the free-energy has an additional trace term as compared to the marginal likelihood obtained from the DTC generative model approach (see equation \ref{eqn:fitc_energy} as $\alpha \rightarrow 0$). The trace term is proportional to the sum of the variances of the training function values given the pseudo-points, $p(\fvec|\uvec)$, it thereby encourages pseudo-input locations that explain the observed data well. This term acts as a regulariser that prevents overfitting which plagues the generative model formulation of DTC. 

The \gls{vfe} approach can be extended to non-linear models including classification \citep{HenMatGha15}, latent variable models \citep{TitLaw10} and state space models \citep{FriCheRas14, Mch14} by restricting $q(\uvec)$ to be Gaussian and optimising its parameters. Indeed, this uncollapsed form of the bound can be beneficial in the context of regression too as it is amenable to stochastic optimisation  \citep{HenFusLaw13}. Additional approximation is sometimes required to compute any remaining intractable non-linear integrals, but these are often low-dimensional. For example, when the likelihood depends on only one latent function value, as is typically the case for regression and classification, the bound requires only 1D integrals $\E_{q(f_n)} \left[ \log p(y_n| \mathrm{f}_n,\theta) \right]$ that can be evaluated using quadrature \citep{HenMatGha15}, for example. 

The \gls{vfe} approach can also be extended to employ inter-domain variables \citep{AlvLueTitLaw10,TobBuiTur15,MatHenTurGha16}. The approach considers the augmented generative model $p(f,g|\theta)$ where to remind the reader the auxiliary process is defined by a linear integral transformation, $g(z) = \int w(z,z')f(z') \text{d} z'$. Variational inference is now performed over both latent processes $q(f,g) = q(f, \uvec, g_{\ne \uvec}|\theta) = p(f,g_{\ne \uvec} | \uvec,\theta) q(\uvec)$. Here the pseudo-data have been placed into the auxiliary process with the idea being that they can induce richer dependencies in the original domain that model the true posterior more accurately. In fact, if the linear integral transformation is parameterised then the transformation can be learned so that it approximates the posterior more accurately.

A key concept underpinning the \gls{vfe} framework is that the pseudo-input locations (and the parameters of the inter-domain transformation, if employed) are purely parameters of the approximate posterior, hence the name `variational parameters'. This distinction is important as it means, for example, that we are free to add pseudo-data as more structure is revealed the underlying function without altering the modelling assumptions (e.g. see \cite{BuiNguTur17} for an example in online inference). Moreover, since the pseudo-input locations are variational parameters, placing priors over them is unnecessary in this framework. Unlike the model parameters, optimisation of variational parameters is automatically protected from overfitting as the optimisation is minimising the KL divergence between the approximate posterior and the true posterior. Indeed, although the DTC posterior is recovered in the regression setting, as we have seen the free-energy is {\it not} equal to the log-marginal likelihood of the DTC generative model, containing an additional term that substantially improves the quality of the optimised pseudo-point input locations. 

The fact that the form of the DTC approximation can be recovered from a direct approximate inference approach and that this new perspective leads to superior pseudo-input optimisation, raises the question; can this also be done for FITC and PITC?


\subsection{Sparse GP Approximation via Approximate Inference: EP}
\gls{ep} is a deterministic inference method \citep{Min01} that is known to outperform \gls{vfe} methods in \gls{gp} classification when unsparsified fully-factored approximations $q(\fvec) = \prod_{n} q_n(\mathrm{f}_n)$ are used \citep{NicRas08}. Motivated by this observation, EP has been combined with the approximate generative modelling approach to handle non-linear likelihoods \citep{NaiHol08,HerHer16}. This begs the question: can the sparsification and the non-linear approximation be handled in a single \gls{ep} inference stage, as for \gls{vfe}? Astonishingly Csat\'{o} and Opper not only developed such a method in 2002 \citep{CsaOpp02}, predating much of the work mentioned above, they showed that it is equivalent to applying the FITC approximation and running EP if further approximation is required. In our view, this is a central result, but it appears to have been largely overlooked by the field. Snelson was made aware of it when writing his thesis \citep{Sne07}, briefly acknowledging Csat\'{o} and Opper's contribution. \citet{QiAbdMin10} extended Csat\'{o} and Opper's work to utilise inter-domain pseudo-points and they additionally recognised that the EP energy function at convergence is equal to the FITC log-marginal likelihood approximation. Interestingly, no additional term arises as it does when the \gls{vfe} approach generalised the DTC generative model approach. 
We are unaware of other work in this vein. 

It is hard to be known for certain why these important results are not widely known, but a contributing factor is that the exposition in these papers is largely at Marr's algorithmic level \citep{Daw98}, and does not focus on the computational level making them challenging to understand. Moreover, Csat\'{o} and Opper's paper was written before EP was formulated in a general way and the presentation, therefore, does not follow what has become the standard approach. In fact, as the focus was online inference, Assumed Density Filtering \citep{KusBud00,ItoXio00} was employed rather than full-blown EP.
One of the main contributions of this paper is to provide a clear computational exposition including an explicit form of the approximating distribution and full details about each step of the EP procedure. 
In addition, to bringing clarity we make the following novel contributions:
\begin{itemize}

\item We show that a generalisation of \gls{ep} called Power EP can subsume the \gls{ep} and \gls{vfe} approaches (and therefore FITC and DTC) into a single unified framework. More precisely, the fixed points of Power EP yield the FITC and VFE posterior distribution under different limits and the Power EP marginal likelihood estimate (the negative `Power EP energy') recovers the FITC marginal likelihood and the VFE too. Critically the connection to the VFE method leans on the new interpretation of Titsias's approach \citep{MatHenTurGha16} outlined in the previous section that directly employs the approximate posterior over function values (rather than augmenting the model with pseudo-points). The connection therefore also requires a formulation of Power EP that involves KL divergence minimisation between stochastic processes.

\item We show how versions of PEP that are intermediate between the existing VFE and EP approaches can be derived, as well as mixed approaches that treat some data variationally and others using EP. We also show how PITC emerges from the same framework and how to incorporate inter-domain transforms. For regression with Gaussian observation noise, we obtain analytical expressions for the fixed points of Power EP in a general case that includes all of these extensions as well as the form of the Power EP marginal likelihood estimate at convergence that is useful for hyper-parameter and pseudo-input optimisation.

\item We consider (Gaussian) regression and probit classification as canonical models on which to test the new framework and demonstrate through exhaustive testing that versions of PEP intermediate between VFE and EP perform substantially better on average. 
The experiments also shed light on situations where VFE is to be preferred to EP and vice versa which is an important open area of research.
%

\end{itemize}

\begin{figure*}[!ht]
	\centering
	\includegraphics[width=\textwidth]{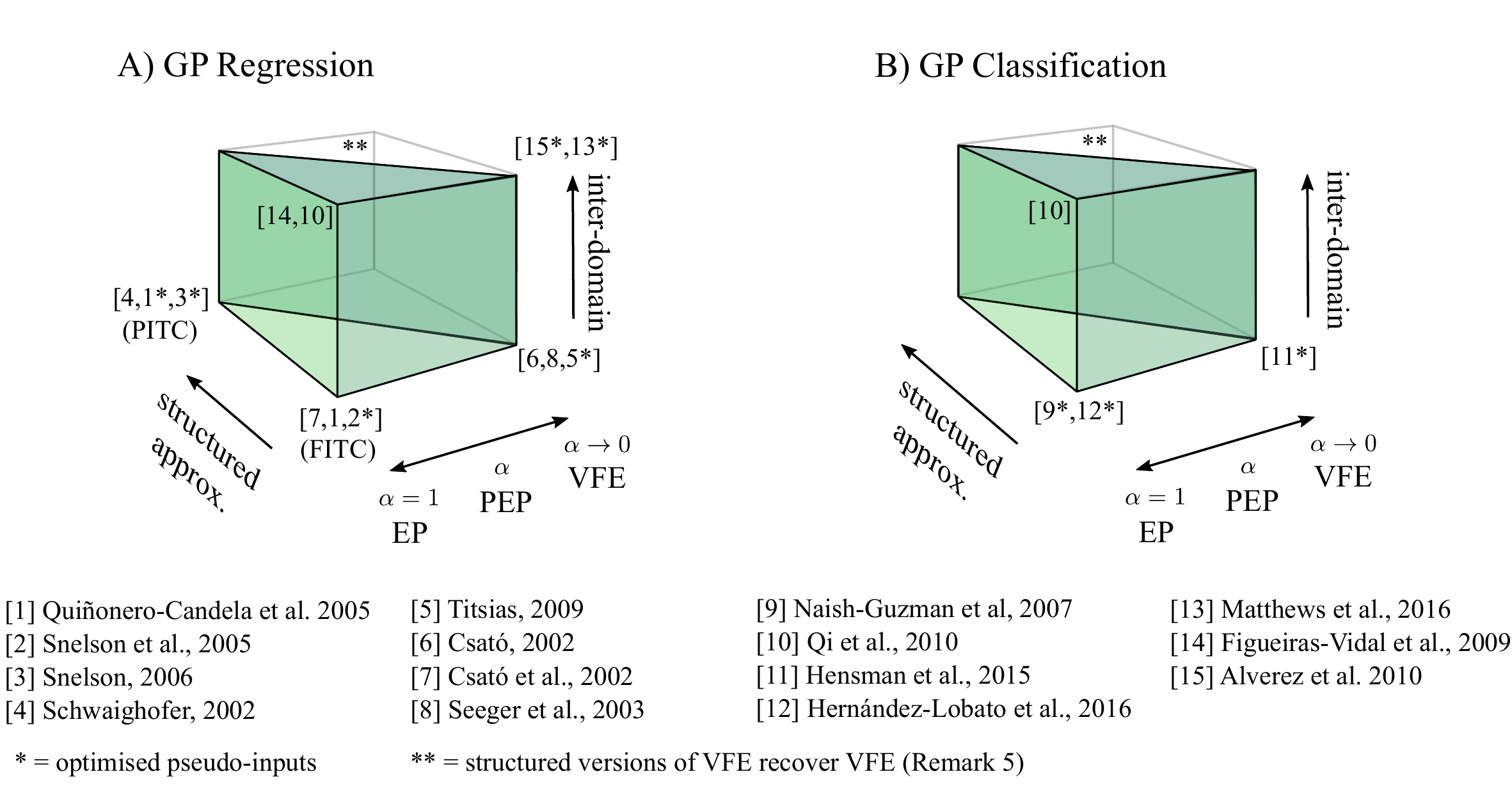}
	\caption{A unified view of pseudo-point GP approximations applied to A) regression, and B) classification. Every point in the algorithm polygons corresponds to a form of GP approximation. Previous algorithms correspond to labelled vertices. The new Power EP framework encompasses the three polygons, including their interior.\label{fig:cubes}}
\end{figure*}

Many of the new theoretical contributions described above are summarised in \cref{fig:cubes} along with their relationship to previous work.


\section{A New Unifying View using Power Expectation Propagation \label{sec:pep}}

In this section, we provide a new unifying view of sparse approximation using Power Expectation Propagation (PEP or Power EP) \citep{Min04}. We review Power EP, describe how to apply it for sparse GP regression and classification, and then discuss its relationship to existing methods.

\subsection{The Joint-Distribution View of Approximate Inference and Learning}
One way of understanding the goal of distributional inference approximations, including the VFE method, EP and Power EP,  is that they return an approximation of a tractable form to the model {\it joint-distribution} evaluated on the observed data. In the case of GP regression and classification, this means  $q^*(f|\theta) \approx p(f,\yvec| \theta)$ where $*$  is used to denote an unnormalised process. Why is the model joint-distribution a sensible object of approximation? The joint distribution can be decomposed into the product of the posterior distribution and the marginal likelihood, $p(f,\yvec| \theta) = p^*(f|\yvec,\theta) = p(f|\yvec, \theta) p(\yvec|\theta)$, the two inferential objects of interest. A tractable approximation to the joint can therefore be similarly decomposed $q^*(f|\theta) = Z q(f|\theta)$ into a normalised component that approximates the posterior $q(f|\theta) \approx p(f|\yvec, \theta) $ and the normalisation constant which approximates the marginal likelihood  $Z \approx p(\yvec|\theta)$. In other words, the approximation of the joint simultaneously returns approximations to the posterior and marginal likelihood.
In the current context tractability of the approximating family means that it is analytically integrable and that this integration can be performed with an appropriate computational complexity. We consider the approximating family comprising unnormalised GPs, $q^*(f|\theta) = Z \mathcal{GP}(f; m_\mathrm{f},V_{\mathrm{ff}'})$.  

The VFE approach can be reformulated in the new context using the un-normalised KL divergence \citep{ZhuRoh97} to measure the similarity between the approximation and the joint distribution
\begin{align}
\overline{\mathrm{KL}}( q^*(f|\theta) ||  p(f,\yvec| \theta) ) = \int q^*(f) \log \frac{q^*(f)}{p(f,\yvec| \theta) } \mathrm{d}f + \int \left ( p(f,\yvec| \theta) - q^*(f) \right)  \mathrm{d}f.  \label{eq:KLnorm}
\end{align}
The un-normalised KL divergence generalises the KL divergence to accommodate un-normalised densities. It is always non-negative and collapses back to the standard form when its arguments are normalised.  Minimising the un-normalised KL with respect to $q^*(f|\theta) = Z_{\mathrm{VFE}} q(f)$ encourages the approximation to match both the posterior and marginal-likelihood, and it yields analytic solutions 
\begin{align}
q^{\mathrm{opt}}(f) = \argmin_{q(f) \in \mathcal{Q}} \mathrm{KL} (q(f) || p(f|\yvec,\theta)),  \;\; \text{and} \;\; Z_{\mathrm{VFE}}^{\mathrm{opt}} = \exp(\mathcal{F}(q^{\mathrm{opt}}(f), \theta)).
\end{align}
That is, the standard variational free-energy approximation to the posterior and marginal likelihood is recovered. One of the pedagogical advantages of framing VFE in this way is that approximation of the posterior and marginal likelihood are committed to upfront, in contrast to the traditional derivation which begins by targeting approximation of the marginal likelihood, but shows that approximation of the posterior emerges as an essential part of this scheme (see section \ref{sec:VFE}). A disadvantage is that optimisation of hyper-parameters must logically proceed by optimising the marginal likelihood approximation, $Z_{\mathrm{VFE}}^{\mathrm{opt}}$, and at first sight therefore appears to necessitate different objective functions for $q^*(f|\theta)$ and $\theta$ (unlike the standard view which uses a single objective from the beginning). However, it is easy to show that maximising the single objective $p(\yvec|\theta) - \overline{\mathrm{KL}}( q^*(f|\theta) ||  p(f,\yvec| \theta) ) $ directly for both $q^*(f|\theta)$ and $\theta$ is equivalent and that this also recovers the standard VFE method (see appendix \ref{app:KL}).

%
\subsection{The Approximating Distribution Employed by Power EP}
\label{subsec:pep}
Power EP also approximates the joint-distribution employing an approximating family whose form mirrors that of the target,
 \begin{align}
   p^*(f|\yvec,\theta)= p(f|\yvec,\theta)p(\yvec|\theta) = p(f|\theta)\prod_n p(y_n|f,\theta) \approx p(f|\theta) \prod_n t_n(\uvec)= q^*(f|\theta). \label{eq:post}
 \end{align}
Here, the approximation retains the exact prior, but each likelihood term in the exact posterior, $p(y_n|\mathrm{f}_n,\theta)$, is approximated by a simple factor $t_n(\uvec)$ that is assumed Gaussian. These simple factors will be iteratively refined by the PEP algorithm such that they will capture the effect that each true likelihood has on the posterior. 


%
%

Before describing the details of the PEP algorithm, it is illuminating to consider an alternative interpretation of the approximation. Together, the approximate likelihood functions specify an un-normalised Gaussian over the pseudo-points that can be written $\prod_n t_n(\uvec) = \mathcal{N}(\tilde{\mathbf{y}}; \tilde{\mathbf{W}}\uvec, \tilde{\Sigma})$ (assuming that the product of these factors is normalisable which may not be the case for heavy tailed likelihoods, for example).

\begin{figure*}[!ht]
	\centering
		 \includegraphics[width=\textwidth]{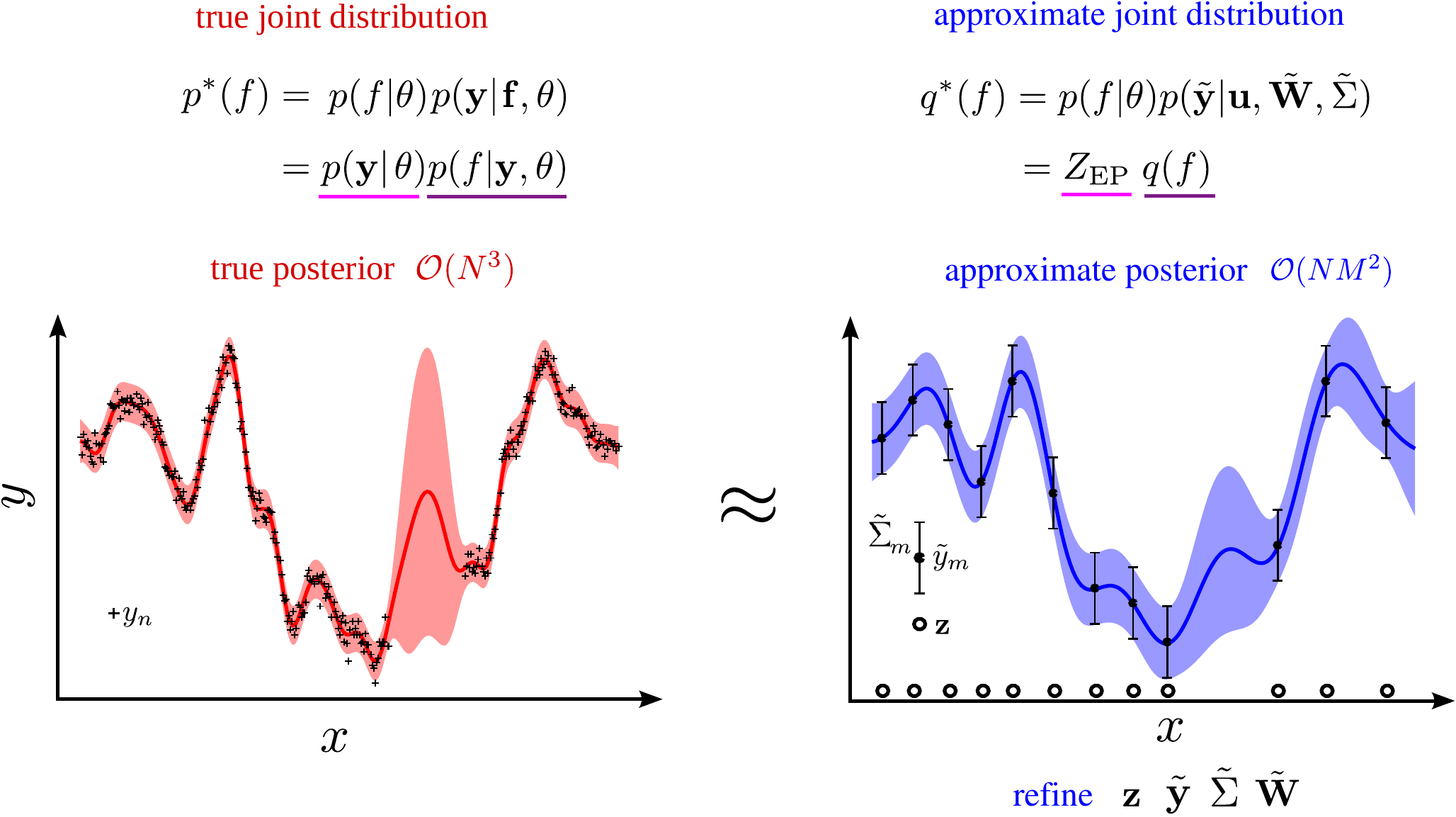}
	\caption{Perspectives on the approximating family. The true joint distribution over the unknown function $f$ and the $N$ data points $\yvec$ (top left) comprises the GP prior and an intractable likelihood function. This is approximated by a surrogate regression model with a joint distribution  over the function $f$ and $M$ surrogate data points $\tilde{\mathbf{y}}$ (top right). The surrogate regression model employs the same GP prior, but uses a Gaussian likelihood function $p(\tilde{\mathbf{y}}| \uvec, \tilde{\mathbf{W}}, \tilde{\Sigma})=\mathcal{N}(\tilde{\mathbf{y}}; \tilde{\mathbf{W}}\uvec, \tilde{\Sigma})$. The intractable true posterior (bottom left) is approximated by refining the surrogate data $\tilde{\mathbf{y}}$ their input locations $\mathbf{z}$ and the parameters of the surrogate model $\tilde{\mathbf{W}}$ and $\tilde{\Sigma}$. \label{fig:approximation-as-regression}}
\end{figure*}

The approximate posterior above can therefore be thought of as the (exact) GP posterior resulting from a surrogate regression problem with surrogate observations $\tilde{\mathbf{y}}$ that are generated from linear combinations of the pseudo-points and additive surrogate noise $\tilde{\mathbf{y}} =  \tilde{\mathbf{W}} \uvec + \tilde{\Sigma}^{1/2} \mathbf{\epsilon} $. We note that the pseudo-points $\uvec$ live on the latent function (or an inter-domain transformation thereof) and the surrogate observations $\tilde{\mathbf{y}}$ will not generally lie on the latent function. The surrorate observations and the pseudo-points are therefore analogous to the data $\yvec$ and the function values $\fvec$ in a normal Gaussian Process regression problem, respectively. To make the paper more specific on this point, we have defined parameters forthe surrogate regression problem explicitly in appendix \ref{app:surrogate}. The PEP algorithm will implicitly iteratively refine $\{ \tilde{\mathbf{y}}, \tilde{\mathbf{W}} , \tilde{\Sigma}\}$ such that exact inference in the simple surrogate regression model returns a posterior and marginal likelihood estimate that is `close' to that returned by performing exact inference in the intractable complex model (see \cref{fig:approximation-as-regression}). 
   
 \subsection{The EP Algorithm}
 
One method for updating the approximate likelihood factors $t_n(\uvec)$ is to minimise the unnormalised KL Divergence between the joint distribution and each of the distributions formed by replacing one of the likelihoods by the corresponding approximating factor \citep{LiHerTur15},  
 \begin{align}
 \argmax_{t_n(\uvec)} \overline{\mathrm{KL}}\left [p(f,\yvec| \theta) \middle \vert \middle \vert  \frac{p(f,\yvec| \theta) t_n(\uvec)} { p(y_n|\mathrm{f}_n,\theta)} \right] =   \argmax_{t_n(\uvec)}  \overline{\mathrm{KL}}[p^*_{\setminus n}(f) p(y_n|\mathrm{f}_n,\theta) || p^*_{\setminus n}(f) t_n(\uvec)].
 \end{align}
 Here we have introduced the leave-one-out joint $p^*_{\setminus n}(f) = p(f,\yvec| \theta)  / p(y_n|\mathrm{f}_n,\theta)$ which makes clear that the minimisation will cause the approximate factors to approximate the likelihoods in the context of the leave-one-out joint. Unfortunately, such an update is still intractable. Instead, EP approximates this idealised procedure by replacing the exact leave-one-out joint on both sides of the KL by the approximate leave-one-out joint (called the cavity) $p^*_{\setminus n}(f) \approx q^*_{\setminus n}(f) = q^*(f)/t_n(\uvec)$.  Not only does this improve tractability, but it also means that the new procedure effectively refines the approximating distribution directly at each stage, rather than setting the component parts in isolation,
 \begin{align}
   \overline{\mathrm{KL}}([q^*_{\setminus n}(f) p(y_n|\mathrm{f}_n,\theta) || q^*_{\setminus n}(f) t_n(\uvec)] = \overline{\mathrm{KL}}([q^*_{\setminus n}(f) p(y_n|\mathrm{f}_n,\theta) || q^*(f)].
 \end{align}
 However, the updates for the approximating factors are now coupled and so the updates must now be iterated, unlike in the idealised procedure. In this way, EP iteratively refines the approximate factors or surrogate likelihoods so that the GP posterior of the surrogate regression task best approximates the posterior of the original regression/classification problem.
   
  \subsection{The Power EP Algorithm}
Power EP is, algorithmically, a mild generalisation of the EP algorithm that instead removes (or includes) a fraction $\alpha$ of the approximate (or true) likelihood functions in the following steps:
\begin{enumerate*}
\item \textbf{Deletion}: compute the cavity distribution by removing a fraction of one approximate factor, $q^*_{\setminus n}(f|\theta) \propto  q^*(f|\theta) / t_n^\alpha(\uvec)$.
\item \textbf{Projection}: first, compute the tilted distribution by incorporating a corresponding fraction of the true likelihood into the cavity, $\tilde{p}(f) = q^*_{\setminus n}(f|\theta) p^\alpha(y_n|\mathrm{f}_n)$. Second, project the tilted distribution onto the approximate posterior using the KL divergence for un-normalised densities, 
\begin{align}
q^*(f|\theta) \leftarrow \argmin_{q^*(f|\theta) \in \mathcal{Q}} \overline{\mathrm{KL}}(\tilde{p}(f) || q^*(f|\theta)).
\end{align}
 Here $\mathcal{Q}$ is the set of allowed $q^*(f|\theta)$ defined by \cref{eq:post}.
\item \textbf{Update}: compute a new fraction of the approximate factor by dividing the new approximate posterior by the cavity, $t_{n,\mathrm{new}}^{\alpha}(\uvec) = q^*(f|\theta) / q^*_{\setminus n}(f|\theta)$, and incorporate this fraction back in to obtain the updated factor, $t_{n}(\uvec) = t_{n,\mathrm{old}}^{1-\alpha}(\uvec) t^{\alpha}_{n,\mathrm{new}}(\uvec)$.
\end{enumerate*}
The above steps are iteratively repeated for each factor that needs to be approximated. Notice that the procedure only involves one likelihood factor to be handled at a time. In the case of analytically intractable likelihood functions, this often requires only low dimensional integrals to be computed. In other words, PEP has transformed a high dimensional intractable integral that is hard to approximate into a set of low dimensional intractable integrals that are simpler to approximate. The procedure is not, in general guaranteed to converge but we did not observe any convergence issues in our experiments. Furthermore, it can be shown to be numerically stable when the factors are log-concave (as in \gls{gp} regression and classification without pseudo-data) \citep{See08}.

If Power EP converges, the fractional updates are equivalent to running the original EP procedure, but replacing the KL minimisation with an alpha-divergence minimisation \citep{ZhuRoh95,Min05},
 \begin{align}
   \overline{\mathrm{D}}_{\alpha} [p^*(f) || q^*(f) ] = \frac{1}{\alpha(1-\alpha)} \int \left [ \alpha p^*(f) + (1-\alpha) q^*(f) -  p^*(f)^{\alpha}  q^*(f)^{1-\alpha} \right] \mathrm{d}f.
 \end{align}
When $\alpha = 1$, the alpha-divergence is the inclusive KL divergence $\overline{\mathrm{D}}_1[p^*(f) || q^*(f) ] = \overline{\mathrm{KL}}[p^*(f) || q^*(f) ]$ recovering EP as expected from the PEP algorithm. As $\alpha \to 0$ the exclusive KL divergence is recovered, $\overline{\mathrm{D}}_{\to 0}[p^*(f) || q^*(f) ] = \overline{\mathrm{KL}}[q^*(f) || p^*(f) ]$, and since minimising a set of local exclusive KL divergences is equivalent to minimising a single global exclusive KL divergence \citep{Min05}, the Power EP solution is the minimum of a variational free-energy (see appendix \ref{app:inclusive_kl} for more details). In the current case, we will now show explicitly that these cases of Power EP recover FITC and Titsias's VFE solution respectively.


\subsection{General Results for Gaussian Process Power EP}

This section describes the Power EP steps in finer detail showing the complexity is $\mathcal{O}(NM^2)$ and laying the ground work for the equivalence relationships. The appendix \ref{app:full_deriv} includes a full derivation. 

We start by defining the approximate factors to be in natural parameter form, making it simple to combine and delete them, $t_n(\uvec) = \normnat(\uvec; z_n,\mathbf{T}_{1, n}, \mathbf{T}_{2, n}) = z_n \exp( \uvec^\intercal\mathbf{T}_{1, n} - \frac{1}{2} \uvec^\intercal \mathbf{T}_{2, n} \uvec)$. We consider full rank $\mathbf{T}_{2, n}$, but will show that the optimal form is rank 1. 
The parameterisation means the approximate posterior over the pseudo-points has natural parameters $\mathbf{T}_{1, \uvec} = \sum_n \mathbf{T}_{1, n}$ and $\mathbf{T}_{2, \uvec} = \mathbf{K}_{\mathbf{uu}}^{-1} + \sum_n \mathbf{T}_{2, n}$ inducing an approximate posterior, $q^*(f|\theta) = \mathcal{Z}_{\mathrm{PEP}} \mathcal{GP}(f; m_\mathrm{f}, V_\mathrm{ff'})$. Here and in what follows, the dependence on the hyperparameters $\theta$ will be suppressed to lighten the notation. The mean and covariance functions of the approximate posterior are
\begin{align}
&m_\mathrm{f} = \mathbf{K}_{\mathrm{f}\uvec}\mathbf{K}_{\uvec\uvec}^{-1} \mathbf{T}_{2, \uvec}^{-1} \mathbf{T}_{1, \uvec}; \;\;\;\; V_\mathrm{ff'} = \mathbf{K}_{\mathrm{f}\mathrm{f}'} - \mathbf{Q}_{\mathrm{f}\mathrm{f}'} + \mathbf{K}_{\mathrm{f}\uvec}\mathbf{K}_{\uvec\uvec}^{-1} \mathbf{T}_{2, \uvec}^{-1} \mathbf{K}_{\uvec\uvec}^{-1}\mathbf{K}_{\uvec\mathrm{f'}}.
\end{align}
{\bf Deletion:} The cavity for data point $n$, $q^*_{\setminus n}(f) \propto  q^*(f) / t_n^\alpha(\uvec)$, has a similar form to the posterior, but the natural parameters are modified by the deletion step, $\mathbf{T}^{\setminus n}_{1, \uvec} = \mathbf{T}_{1, \uvec} - \alpha\mathbf{T}_{1, n}$ and $\mathbf{T}^{\setminus n}_{2, \uvec} = \mathbf{T}_{2, \uvec} - \alpha\mathbf{T}_{2, n} $, yielding the following mean and covariance functions
\begin{align}
&m_\mathrm{f}^{\setminus n} = \mathbf{K}_{\mathrm{f}\uvec}\mathbf{K}_{\uvec\uvec}^{-1} \mathbf{T}^{\setminus n, -1}_{2, \uvec} \mathbf{T}^{\setminus n}_{1, \uvec}; \;\;\;\; V_\mathrm{ff'}^{\setminus n} = \mathbf{K}_{\mathrm{f}\mathrm{f}'} - \mathbf{Q}_{\mathrm{f}\mathrm{f}'} + \mathbf{K}_{\mathrm{f}\uvec}\mathbf{K}_{\uvec\uvec}^{-1} \mathbf{T}^{\setminus n, -1}_{2, \uvec} \mathbf{K}_{\uvec\uvec}^{-1}\mathbf{K}_{\uvec\mathrm{f'}}.
\end{align}
{\bf Projection:} The central step in Power EP is the projection. Obtaining the new approximate un-normalised posterior $q^*(f)$ by minimising $\overline{\mathrm{KL}}(\tilde{p}(f) || q^*(f))$ would na\"ively appear intractable. Fortunately,
\begin{remark}
Due to the structure of the approximate posterior, $q^*(f) = p(f_{\neq \uvec}|\uvec) q^*(\uvec)$, the objective, $\overline{\mathrm{KL}}(\tilde{p}(f) || q^*(f))$ is minimised when $\textcolor{NavyBlue}{\E_{\tilde{p} (f)} [\phi(\uvec)] = \E_{q^*(\uvec)} [\phi(\uvec)]}$, where $\phi(\uvec) = \{\uvec, \uvec\uvec^\intercal\}$ are the sufficient statistics, that is when the moments at the pseudo-inputs are matched. \label{rem:mm}
\end{remark}
This is the central result from which computational savings are derived. Furthermore, this moment matching condition would appear to necessitate computation of a set of integrals to find the zeroth, first and second moments. However, the technique known as `differentiation under the integral sign'\footnote{In this case, the dominated convergence theorem can be used to justify the interchange of integration and differentiation \citep[see e.g.][]{Bro86}.} provides a useful shortcut that only requires one integral to compute the log-normaliser of the tilted distribution, $\log \tilde{Z}_n = \log \E_{q^*_{\setminus n}(f)} [p^\alpha(y_n|\mathrm{f}_n)]$, before differentiating w.r.t.~the cavity mean to give
\begin{align}
    & \mathbf{m}_\uvec = \mathbf{m}^{\setminus n}_{\uvec} + \mathbf{V}^{\setminus n}_{\uvec \mathrm{f}_n} \frac{\dd \log \tilde{Z}_n}{\dd m^{\setminus n}_{\mathrm{f}_n}}; & \mathbf{V}_\uvec = \mathbf{V}^{\setminus n}_{\uvec} + \mathbf{V}^{\setminus n}_{\uvec \mathrm{f}_n} \frac{\dd^2 \log \tilde{Z}_n}{\dd (m^{\setminus n}_{\mathrm{f}_n} )^2} \mathbf{V}^{\setminus n}_{\mathrm{f}_n\uvec}. & \label{eqn:mm}
\end{align}
{\bf Update:} Having computed the new approximate posterior, the approximate factor $t_{n,\mathrm{new}}(\uvec) = q^*(f) / q^*_{\setminus n}(f)$ can be straightforwardly obtained, resulting in,
\begin{align}
\small
\mathbf{T}_{1,n,\mathrm{new}} = \mathbf{V}_\uvec^{-1}\mathbf{m}_\uvec - (\mathbf{V}^{\setminus n}_\uvec)^{-1}\mathbf{m}^{\setminus n}_\uvec,\; 
\mathbf{T}_{2,n,\mathrm{new}} = \mathbf{V}_\uvec^{-1} - (\mathbf{V}^{\setminus n}_\uvec)^{-1}, \; 
z_n^\alpha = \tilde{Z}_n \mathrm{e}^{\mathcal{G}(q_*^{\setminus n}(\uvec)) - \mathcal{G}(q^*(\uvec))}, \nonumber
\end{align}
\sloppy where we have defined the log-normaliser as the functional $\mathcal{G}(\normnat(\uvec; z,\mathbf{T}_{1}, \mathbf{T}_{2})) = \log \int \normnat(\uvec; z,\mathbf{T}_{1}, \mathbf{T}_{2}) \dd\uvec$. Remarkably, these results and \cref{eqn:mm} reveals that $\mathbf{T}_{2,n,new}$ is a rank-1 matrix. As such, the minimal and simplest way to parameterise the approximate factor is $t_{n}(\uvec) = z_n\norm(\mathbf{K}_{\mathrm{f}_n\uvec}\mathbf{K}_{\uvec\uvec}^{-1}\uvec; g_n, v_n)$, where $g_n$ and $v_n$ are scalars, resulting in a significant memory saving and $\mathcal{O}(NM^2)$ cost.

In addition to providing the approximate posterior after convergence, Power EP also provides an approximate log-marginal likelihood for model selection and hyper-parameter optimisation,
\begin{align}
\log \mathcal{Z}_{\mathrm{PEP}} = \log \int p(f) \prod_n t_n(\uvec) \dd f = \mathcal{G}(q^*(\uvec)) - \mathcal{G}(p^*(\uvec)) + \sum_n \log z_n. \label{eqn:logZ}
\end{align}
Armed with these general results, we now consider the implications for Gaussian Process regression.


\subsection{Gaussian Regression case\label{sec:fitc}}
When the model contains Gaussian likelihood functions, closed-form expressions for the Power EP approximate factors at convergence can be obtained and hence the approximate posterior: 
\begin{align}
t_n(\uvec) = \norm(\mathbf{K}_{\mathrm{f}_n\uvec} \mathbf{K}_{\mathbf{uu}}^{-1} \uvec; y_n, \alpha D_{\mathrm{f}_n\mathrm{f}_n} + \sigma_y^2), \;\;
q(\uvec) = \norm(\uvec; \mathbf{K}_{\uvec\mathrm{f}}\overline{\mathbf{K}}_{\mathbf{ff}}^{-1} \yvec, \mathbf{K}_{\uvec\uvec} - \mathbf{K}_{\uvec\mathbf{f}}\overline{\mathbf{K}}_{\mathbf{ff}}^{-1}\mathbf{K}_{\mathbf{f}\uvec}) \nonumber
\end{align}
where $\overline{\mathbf{K}}_{\mathbf{ff}} = \mathbf{Q}_{\mathbf{ff}} + \alpha \mathrm{diag} (\mathbf{D}_{\mathbf{ff}}) + \sigma_y^2\mathrm{I}$ and $\mathbf{D}_{\mathbf{ff}} = \mathbf{K}_{\mathbf{ff}} - \mathbf{Q}_{\mathbf{ff}}$ as defined in \cref{sec:gprc}. These analytic expressions can be rigorously proven to be the stable fixed point of the Power EP procedure using remark \ref{rem:mm}. Briefly, assuming the factors take the form above, the natural parameters of the cavity $q^*_{\setminus n}(\uvec)$ become,
\begin{align}
&\mathbf{T}^{\setminus n}_{1, \uvec} = \mathbf{T}_{1, \uvec} - \alpha \gamma_n y_n \mathbf{K}_{\mathrm{f}_n\uvec} \mathbf{K}_{\mathbf{uu}}^{-1} , \;\;\;
\mathbf{T}^{\setminus n}_{2, \uvec} = \mathbf{T}_{2, \uvec} -  \alpha \gamma_n \mathbf{K}_{\mathbf{uu}}^{-1} \mathbf{K}_{\uvec \mathrm{f}_n} \mathbf{K}_{\mathrm{f}_n\uvec} \mathbf{K}_{\mathbf{uu}}^{-1}, \;\;\; 
\end{align}
where $\gamma_n^{-1} = \alpha D_{\mathrm{f}_n\mathrm{f}_n} + \sigma_y^2$. The subtracted quantities in the equations above are exactly the contribution the likelihood factor makes to the cavity distribution (see remark \ref{rem:mm}) so  
$ \int q^*_{\setminus n}(f) p^\alpha(y_n|\mathrm{f}_n) \text{d}f_{\ne \uvec} = q^*_{\setminus n}(\uvec) \int p(\mathrm{f}_n|\uvec) p^\alpha(y_n|\mathrm{f}_n) \text{d} \mathrm{f}_n \propto q^*(\uvec)$. 
Therefore, the posterior approximation remains unchanged after an update and the form for the factors above is the fixed point. Moreover, the approximate log-marginal likelihood is also analytically tractable,
\begin{align}
\small   \log \mathcal{Z}_{\mathrm{PEP}} &= - \frac{N}{2}\log(2\pi) - \frac{1}{2}\log|\overline{\mathbf{K}}_{\mathbf{ff}}| - \frac{1}{2}\mathbf{y}^{\intercal} \overline{\mathbf{K}}_{\mathbf{ff}}^{-1} \mathbf{y} - \frac{1-\alpha}{2\alpha} \sum_n \log\left(1+ \alpha D_{\mathrm{f}_n\mathrm{f}_n}/ \sigma^2_y \right)\nonumber.
\end{align}
We now look at special cases and the correspondence to the methods discussed in \cref{sec:gprc}.
\begin{remark}
When $\alpha = 1$ [EP], the Power EP posterior becomes the FITC posterior in \cref{eqn:fitc_post} and the Power EP approximate marginal likelihood becomes the FITC marginal likelihood in \cref{eqn:fitc_energy}. In other words, the FITC approximation for GP regression is, surprisingly, equivalent to running an EP algorithm for sparse GP posterior approximation to convergence. 
\end{remark}
\begin{remark}
As $\alpha \to 0$ the approximate posterior and approximate marginal likelihood are identical to that of the VFE approach in \cref{eqn:vfe_post,eqn:vfe_energy} \citep{Tit09a}. This result uses the limit: $\lim_{x \to 0} x^{-1} \log(1+x) = 1$. So FITC and Titsias's VFE approach employ the same form of pseudo-point approximation, but refine it in different ways.
\end{remark}

\begin{remark}
For fixed hyper-parameters, a single pass of Power EP is sufficient for convergence in the regression case.
\end{remark}


\subsection{Extensions: Structured, Inter-domain and Multi-power Power EP Approximations}
The framework can now be generalised in three orthogonal directions:
\begin{enumerate}
\item enable structured approximations to be handled that retain more dependencies in the spirit of PITC (see section \ref{sec:genmod})
\item incorporate inter-domain pseudo-points thereby adding further flexibility to the form of the approximate posterior 
\item employ different powers $\alpha$ for each factor (thereby enabling e.g.~VFE updates to be used for some data points and EP for others).
\end{enumerate}
Given the groundwork above, these three extensions are straightforward. In order to handle structured approximations,  we take inspiration from PITC and partition the data into $B$ disjoint blocks $\yvec_b = \{ y_n \}_{n \in \mathcal{B}_b}$ (see section \ref{sec:genmod}). Each PEP factor update will then approximate an entire block which will contain a set of data points, rather than just a single one. This is related to a form of EP approximation that has recently been used to distribute Monte Carlo algorithms across many machines \citep{GelVehJylRobChoCun14, XuLakTehZhu14}. 
In order to handle inter-domain variables, we define a new domain via a linear transform $g(\xvec) = \int \text{d}\xvec' W(\xvec,\xvec') f(\xvec')$ which now contains the pseudo-points $g =\{g_{\ne \uvec},\uvec \} $. Choices for $W(\xvec,\xvec')$ include Gaussians or wavelets.
These two extensions mean that the approximation becomes,
\begin{align}
p(f,g) \prod_b p(\yvec_b|f) \approx p(f,g) \prod_b t_b(\uvec) = q^*(f).
\end{align}
Power EP is then performed using private powers $\alpha_b$ for each data block, which is the third generalisation mentioned above. Analytic solutions are again available (covariance matrices now incorporate the inter-domain transform)
\begin{align}
& t_b(\uvec) = \norm(\mathbf{K}_{\mathbf{f}_b\uvec} \mathbf{K}_{\mathbf{uu}}^{-1} \uvec; \yvec_b, \alpha_{b} \mathbf{D}_{\mathbf{f}_b\mathbf{f}_b} + \sigma_y^2\mathrm{I}), \;\;\;\; 
q(\uvec) = \norm(\uvec; \mathbf{K}_{\uvec\mathrm{f}}\overline{\mathbf{K}}_{\mathbf{ff}}^{-1} \yvec, \mathbf{K}_{\uvec\uvec} - \mathbf{K}_{\uvec\mathbf{f}}\overline{\mathbf{K}}_{\mathbf{ff}}^{-1}\mathbf{K}_{\mathbf{f}\uvec}) & \nonumber
\end{align}
where $\overline{\mathbf{K}}_{\mathbf{ff}} = \mathbf{Q}_{\mathbf{ff}} + \mathrm{blkdiag} (\{ \alpha_b \mathbf{D}_{\mathbf{f}_b\mathbf{f}_b} \}_{b=1}^B) + \sigma_y^2\mathrm{I}$ and $\mathrm{blkdiag}$ builds a block-diagonal matrix from its inputs. The approximate log-marginal likelihood can also be obtained in closed-form,
\begin{align}
	\small
    \log \mathcal{Z}_{\mathrm{PEP}} &= - \frac{N}{2}\log(2\pi) - \frac{1}{2}\log|\overline{\mathbf{K}}_{\mathbf{ff}}| - \frac{1}{2}\mathbf{y}^{\intercal} \overline{\mathbf{K}}_{\mathbf{ff}}^{-1} \mathbf{y} +  \sum_b \frac{1-\alpha_b}{2\alpha_b} \log\left(\mathrm{I}+\alpha_b \mathbf{D}_{\mathbf{f}_b\mathbf{f}_b}/ \sigma^2_y \right).\nonumber
\end{align}
\begin{remark}
When $\alpha_b = 1$ and $W(\xvec,\xvec') = \delta(\xvec-\xvec')$ the structured Power EP posterior becomes the PITC posterior and the Power EP approximate marginal likelihood becomes the PITC marginal likelihood. Additionally, when $B=N$ we recover FITC as discussed in \cref{sec:fitc}.
\end{remark}
\begin{remark}
When $\alpha_b \rightarrow 0$ and $W(\xvec,\xvec') = \delta(\xvec-\xvec')$ the structured Power EP posterior and approximate marginal likelihood becomes identical to the VFE approach \citep{Tit09a}. This is a result of the equivalence of local and global exclusive KL divergence minimisation. See appendix \ref{app:inclusive_kl} for more details and \cref{fig:cubes} for more relationships.
\end{remark}

\subsection{Classification}
For classification, the non-Gaussian likelihood prevents an analytic solution. As such, the iterative Power EP procedure is required to obtain the approximate posterior. The projection step requires computation of the log-normaliser of the tilted distribution, $\log \tilde{Z}_n = \log \E_{q^*_{\setminus n}(f)} [p^\alpha(y_n|f)] = \log \E_{q^*_{\setminus n}(\mathrm{f}_n)} [\Phi^\alpha(y_n\mathrm{f}_n)]$. For general $\alpha$, this quantity is not available in closed form\footnote{except for special cases, e.g.~when $\alpha=1$ and $\Phi(x)$ is the probit inverse link function, $\Phi(x) = \int_{-\infty}^x \norm(a; 0, 1) \dd a$.}. However, it involves a one-dimensional expectation of a non-linear function of a normally-distributed random variable and, therefore, can be approximated using numerical methods, e.g.~Gauss-Hermite quadrature. This procedure gives an approximation to the expectation, resulting in an approximate update for the posterior mean and covariance. The approximate log-marginal likelihood can also be obtained and used for hyper-parameter optimisation. As $\alpha\to 0$, it becomes the variational free-energy used in \citep{HenMatGha15} which employs quadrature for the same purpose. 
These relationships are shown in \cref{fig:cubes} which also shows that inter-domain transformations and structured approximations have not yet been fully explored in the classification setting. In our view, the inter-domain generalisation would be a sensible one to pursue and it is mathematically and algorithmically straightforward. The structured approximation variant is more complicated as it requires multiple non-linear likelihoods to be handled at each step of EP. This will require further approximation such as using Monte Carlo methods \citep{GelVehJylRobChoCun14, XuLakTehZhu14}. In addition, when $\alpha=1$, $M=N$ and the pseudo-points are at the training inputs, the standard EP algorithm for GP classification is recovered \citep[sec.~3.6]{RasWil05}.

Since the proposed Power EP approach is general, an extension to other likelihood functions is as simple as for VFE methods \citep{DezBon15}. For example, the multinomial probit likelihood can be handled in the same way as the binary case, where the log-normaliser of the tilted distribution can be computed using a $C$-dimensional Gaussian quadrature [$C$ is the number of classes] \citep{SeeJor04} or nested EP \citep{RiiJylVeh13}. 
\subsection{Complexity\label{sec:complexity}}
The computational complexity of all the regression and classification methods described in this section is $\mathcal{O}(NM^2)$ for training, and $\mathcal{O}(M^2)$ per test point for prediction. The training cost can be further reduced to $\mathcal{O}(M^3)$, in a similar vein to the uncollapsed \gls{vfe} approach \citep{HenFusLaw13, HenMatGha15}, by employing stochastic updates of the posterior and stochastic optimisation of the hyper-parameters using minibatches of data points \citep{HerHer16}. In particular, the Power EP update steps in \cref{subsec:pep} are repeated for only a small subset of training points and for only a small number of iterations. The approximate log-marginal likelihood in \cref{eqn:logZ} is then computed using this minibatch and optimised as if the Power EP procedure has converged. This approach results in a computationally efficient training scheme, at the cost of returning noisy hyper-parameter gradients. In practice, we find that the noise can be handled using stochastic optimisers such as Adam \citep{KinBa15}. In summary, given these advances the general PEP framework is as scalable as variational inference.

\section{Experiments\label{sec:exp}}
The general framework described above lays out a large space of potential inference algorithms suggesting many exciting directions for innovation. The experiments considered in the paper will investigate only one aspect of this space; how do algorithms that are intermediate between VFE ($\alpha=0$) and EP/FITC ($\alpha=1$) perform? Specifically, we will investigate how the performance of the inference scheme varies as a function of $\alpha$ and whether this depends on; the type of problem (classification or regression); the dataset (synthetic datasets, 8 real world regression datasets and 6 classification datasets); the performance metric (we compare metrics that require point-estimates to those that are uncertainty sensitive). An important by-product of the experiments is that they provide a comprehensive comparison between the VFE and EP approaches which has been an important area of debate in its own right.
The results presented below are compact summaries of a large number of experiments full details of which are included in the appendix \ref{app:extra_exp} (along with additional experiments). Python and Matlab implementations are available at \url{http://github.com/thangbui/sparseGP_powerEP}.

\subsection{Regression on Synthetic Datasets}
\begin{figure*}[!ht]
\centering
\includegraphics[width=\textwidth]{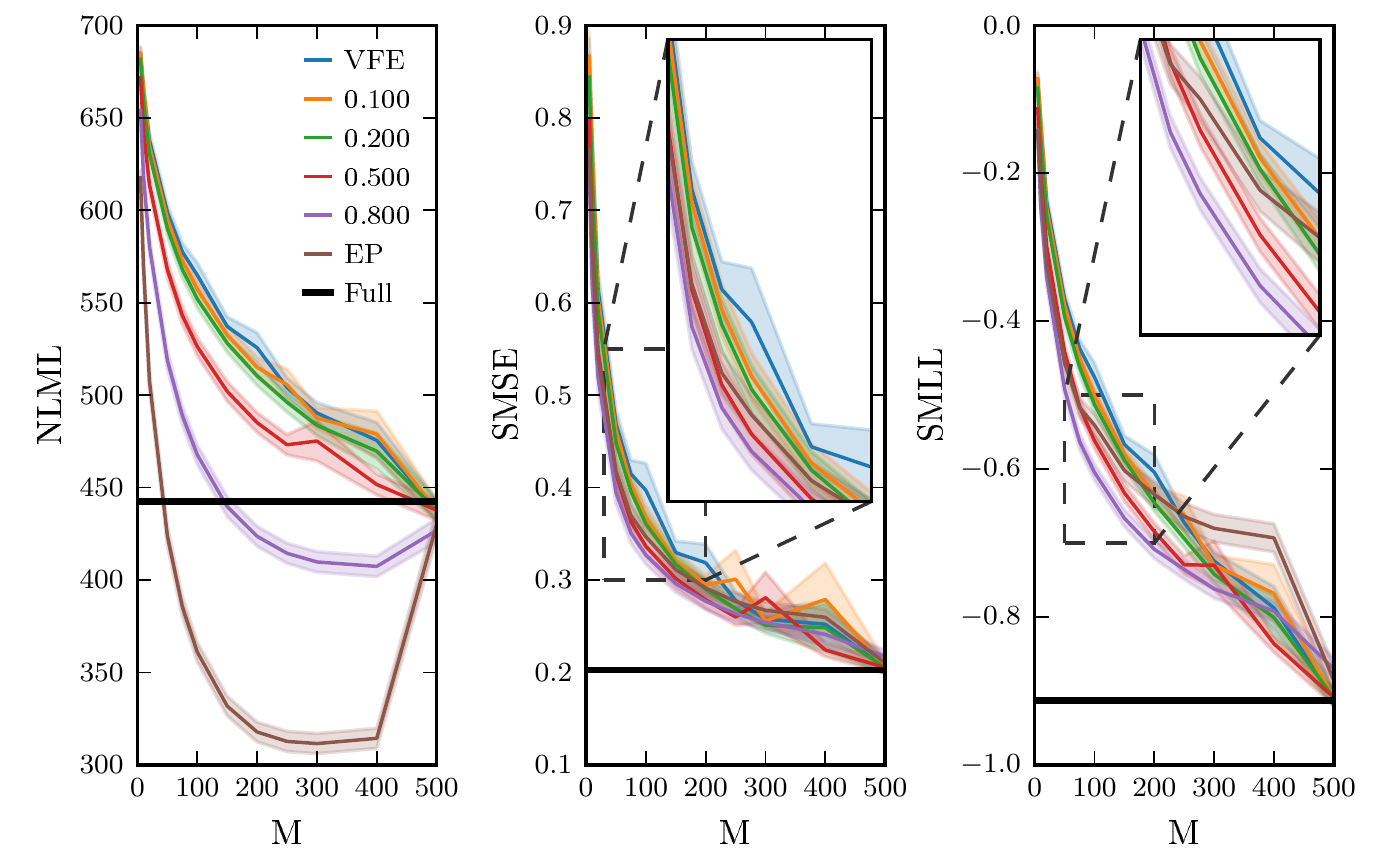}
\caption{The performance of various $\alpha$ values averaged over 30 trials. See text for more details\label{fig:res_toy}}
\end{figure*}

In the first experiment, we investigate the performance of the proposed Power EP method on toy regression datasets where ground truth is known. We vary $\alpha$ (from 0 VFE to 1 EP/FITC) and the number of pseudo-points (from 5 to 500). We use thirty datasets, each comprising 1000 data points with five input dimensions and one output dimension, that were drawn from a GP with an Automatic Relevance Determination squared exponential kernel. A 50:50 train/test split was used. The hyper-parameters and pseudo-inputs were found by optimising the PEP energy using L-BFGS with a maximum of 2000 function evaluations. The performances are compared using two metrics: standardised mean squared error (SMSE) and standardised mean log loss (SMLL) as described in \citep[page 23]{RasWil05}. The approximate negative log-marginal likelihood (NLML) for each experiment is also computed. The mean performance using Power EP with different $\alpha$ values and full GP regression is shown in \cref{fig:res_toy}. The results demonstrate that as M increases, the SMLL and SMSE of the sparse methods approach that of full GP. Power EP with $\alpha=0.8$ or $\alpha=1$ (EP) overestimates the log-marginal likelihood when intermediate numbers of pseudo-points are used, but the overestimation is markedly less when $M=N=500$. Importantly, however, an intermediate value of $\alpha$ in the range 0.5-0.8 seems to be best for prediction on average, outperforming both EP and VFE.
\subsection{Regression on Real-world Datasets\label{sec:reg_exp}}
The experiment above was replicated on 8 UCI regression datasets, each with 20 train/test splits. We varied $\alpha$ between $0$ and 1, and $M$ was varied between 5 and 200. Full details of the experiments along with extensive additional analysis is presented in the appendices. Here we concentrate on several key aspects. First we consider pairwise comparisons between VFE ($\alpha \to 0$), Power EP with $\alpha=0.5$ and EP/FITC ($\alpha=1$) on both the SMSE and SMLL evaluation metrics. Power EP with $\alpha=0.5$ was chosen because it is the mid-point between VFE and EP and because settings around this value empirically performed the best on average across all datasets, splits, numbers of inducing points, and evaluation metrics.

\begin{landscape}
\begin{figure}[!ht]
    \centering
    \includegraphics[width=\linewidth]{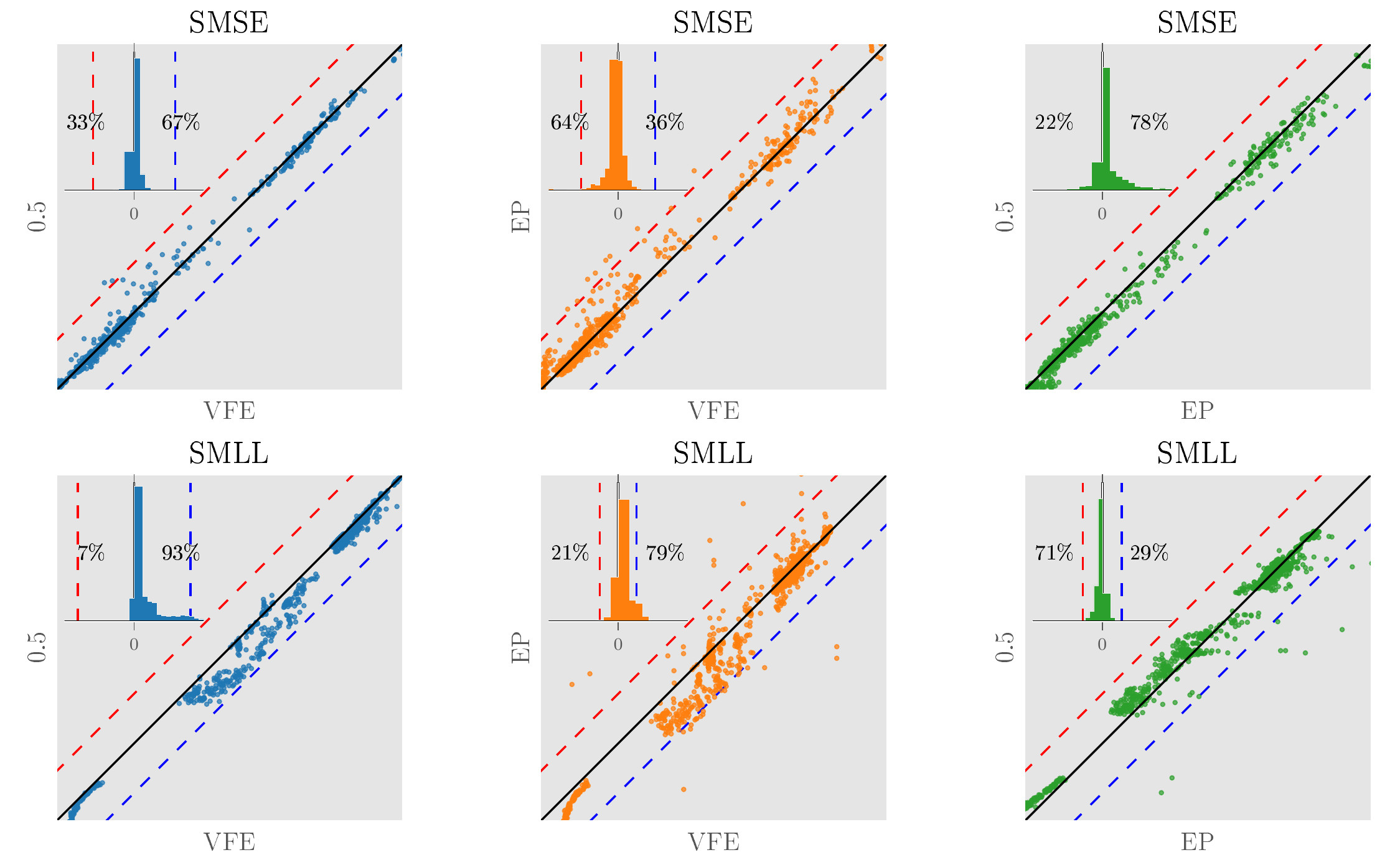}
	\caption{Pair-wise comparisons between Power EP with $\alpha=0.5$, EP ($\alpha=1$) and VFE ($\alpha\to 0$), evaluated on several regression datasets and various settings of $M$. Each coloured point is the result for one split. Points that are below the diagonal line illustrate the method on the $y$-axis is better than the method on the $x$-axis. The inset diagrams show the histograms of the difference between methods ($x$-value $-$ $y$-value), and the counts of negative and positive differences. Note that this indicates pairwise ranking of the two methods. Positive differences mean the $y$-axis method is better than the $x$-axis method and vice versa. For example, the middle, bottom plot shows EP is on average better than VFE.\label{fig:reg_scatter_all}}

\end{figure}
\end{landscape}

In \cref{fig:reg_scatter_all}A we plot (for each dataset, each split and each setting of $M$) the evaluation scores obtained using one inference algorithm (e.g.~PEP $\alpha=0.5$) against the score obtained using another (e.g.~VFE $\alpha=0$). In this way, points falling below the identity line indicate experiments where the method on the y-axis outperformed the method on the x-axis. These results have been collapsed by forming histograms of the difference in the performance of the two algorithms, such that mass to the right of zero indicates the method on the y-axis outperformed that on the x-axis. The proportion of mass on each side of the histogram, also indicated on the plots, shows in what fraction of experiments one method returns a more accurate result than the other. This is a useful summary statistic, linearly related to the average rank, that we will use to unpack the results. The average rank is insensitive to the magnitude of the performance differences and readers might worry that this might give an overly favourable view of a method that performs the best frequently, but only by a tiny margin, and when it fails it does so catastrophically. However, the histograms indicate that the methods that win most frequently tend also to `win big' and `lose small', although EP is a possible exception to this trend (see the outliers below the identity line on the bottom right-hand plot). 

A clear pattern emerges from these plots. First PEP $\alpha=0.5$ is the best performing approach on the SMSE metric, outperforming VFE 67\% of the time and EP 78\% of the time. VFE is better than EP on the SMSE metric 64\% of the time. Second, EP performs the best on the SMLL metric, outperforming VFE 93\% of the time and PEP $\alpha=0.5$ 71\% of the time. PEP $\alpha=0.5$ outperforms VFE in terms of the SMLL metric 93\% of the time. 

\begin{figure*}[t!]
    \centering
    \begin{subfigure}[t]{0.5\textwidth}
        \centering
        \includegraphics[trim={0.5cm 0.5cm 0.5cm 0cm},clip,width=\textwidth]{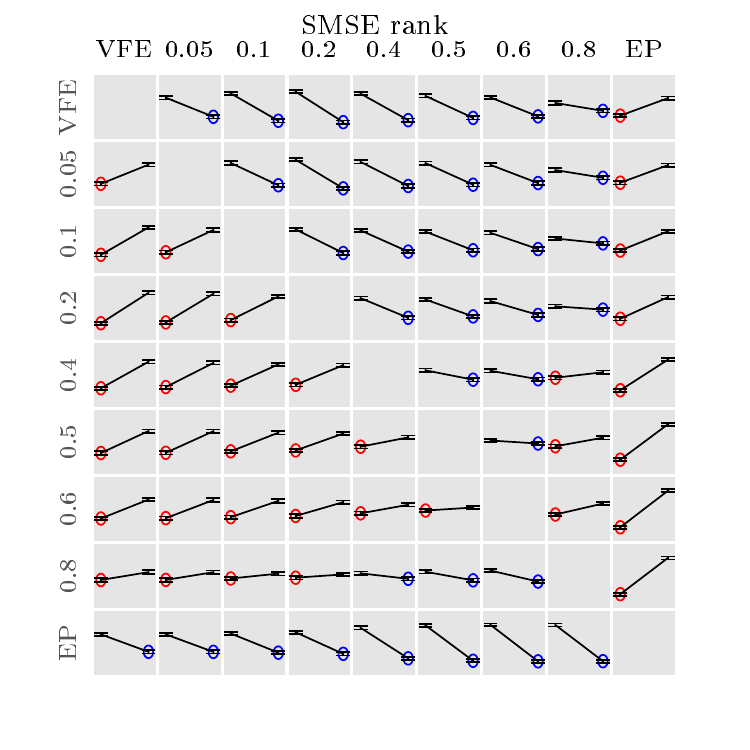}
    \end{subfigure}%
    ~ 
    \begin{subfigure}[t]{0.5\textwidth}
        \centering
        \includegraphics[trim={0.5cm 0.5cm 0.5cm 0cm},clip,width=\textwidth]{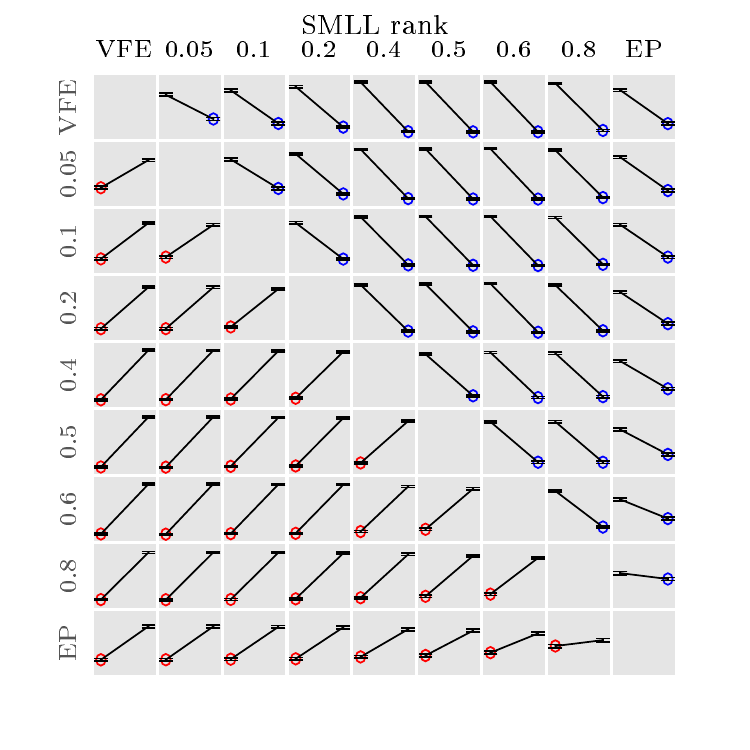}
    \end{subfigure}
    \includegraphics[trim={0.5cm 0cm 0cm 0cm},clip,width=\textwidth]{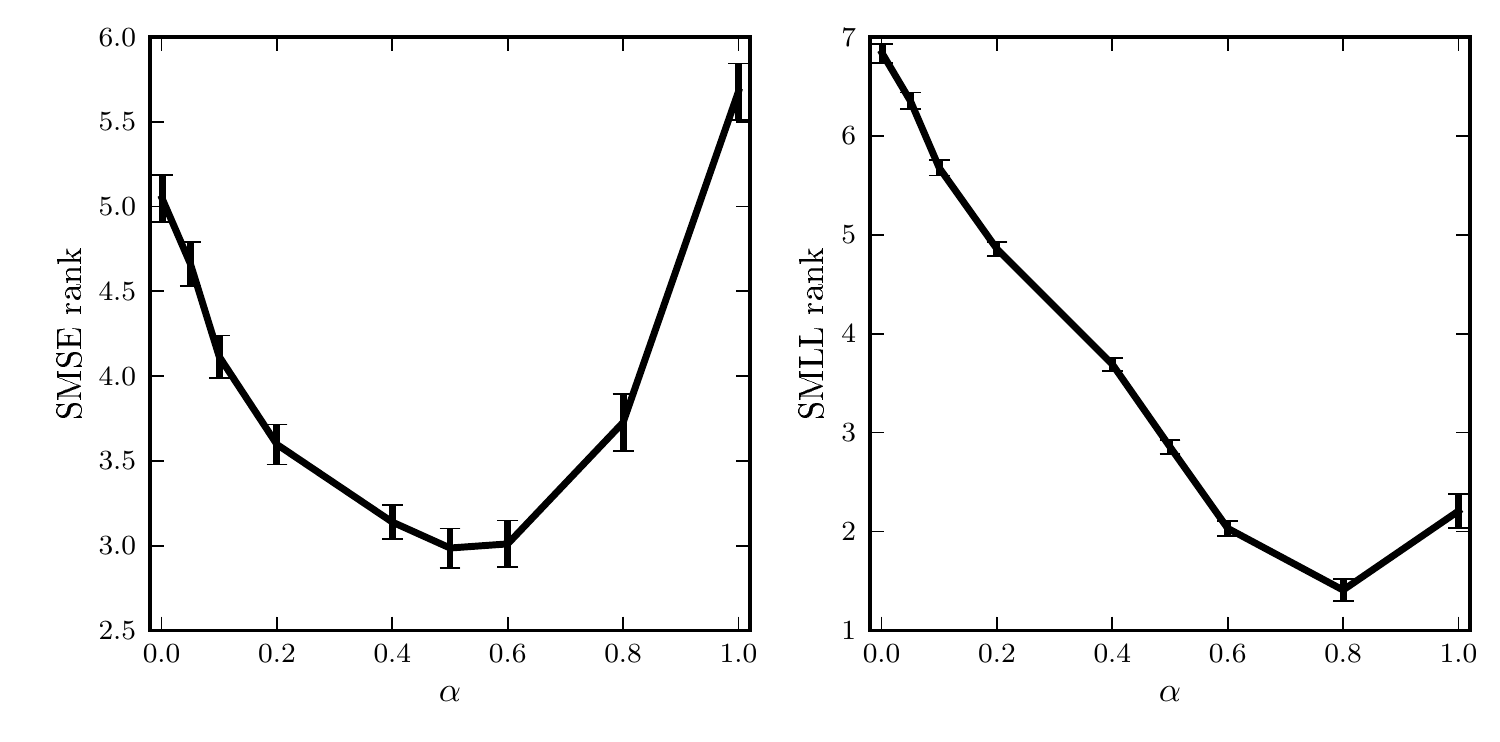}
    \caption{Average ranking of various $\alpha$ values in the regression experiment, lower is better. Top plots show the pairwise comparisons. Red circles denote rows being better than the corresponding columns, and blue circles mean vice versa. Bottom plots show the ranks of all methods when being compared together. Intermediate $\alpha$ values (not EP or VFE) are best on average.\label{fig:reg_rank}}
\end{figure*}

These pairwise rank comparisons have been extended to other values of $\alpha$ in \cref{fig:reg_rank}A. Here, each row of the figure compares one approximation with all others. Horizontal bars indicate that the methods have equal average rank.  Upward sloping bars indicate the method shown on that row has lower average rank (better performance), and downward sloping bars indicate higher average rank (worse performance). The plots show that PEP $\alpha=0.5$ outperforms all other methods on the SMSE metric, except for PEP $\alpha=0.6$ which is marginally better. EP is outperformed by all other methods, and VFE only outperforms EP on this metric.  On the other hand, EP is the clear winner on the SMLL metric, with performance monotonically decreasing with $\alpha$ so that VFE is the worst. 

The same pattern of results is seen when we simultaneously compare all of the methods, rather than considering sets of pairwise comparisons.  The average rank plots shown in \cref{fig:reg_scatter_all}B were produced by sorting the performances of the 8 different approximating methods for each dataset, split, and number of pseudo-points $M$ and assigning a rank. These ranks are then averaged over all datasets and their splits, and settings of $M$.  PEP $\alpha=0.5$ is the best for the SMSE metric, and the two worst methods are EP and VFE. PEP $\alpha=0.8$ is the best for the SMLL metric, with EP and PEP $\alpha=0.6$ not far behind (when EP performs poorly it can do so with a large magnitude, explaining the discrepancy with the pairwise ranks).

There is some variability between individual datasets, but the same general trends are clear: For MSE $\alpha = 0.5$ is better than VFE on 6/8 datasets and EP on 8/8 datasets, whilst VFE is better than EP on 3 datasets (the difference on the others being small). For NLL EP is better than $\alpha=0.5$ on 5/8 datasets and VFE on 7/8 datasets,  whilst  $\alpha=0.5$ is better than VFE on 8/8 datasets. Performance tends to increase for all methods as a function of the number of pseudo-points M. The interaction between the choice of $M$ and the best performing inference method is often complex and variable across datasets making it hard to give precise advice about selecting $\alpha$ in an $M$ dependent way.

In summary, we make the following recommendations based on these results for GP regression problems. For a MSE loss, we recommend using $\alpha=0.5$. For a NLL we recommend using EP.  It is possible that more fine grained recommendations are possible based upon details of the dataset and the computational resources available for processing, but further work will be needed to establish this.

\subsection{Binary Classification\label{sec:classification}}
We also evaluated the Power EP method on 6 UCI classification datasets, each has 20 train/test splits. The details of the datasets are included in appendix \ref{app:classification}. The datasets are all roughly balanced, and the most imbalanced is \texttt{pima} with 500 positive and 267 negative data points. Again $\alpha$ was varied between 0 and 1, and $M$ was varied between 10 and 100. We adopt the experimental protocol discussed in \cref{sec:complexity}, including: (i) not waiting for Power EP to converge before making hyper-parameter updates, (ii) using minibatches of data points for each Power EP sweep, (iii) parallel factor updates. The Adam optimiser was used with default hyper-parameters to handle the noisy gradients produced by these approximations \citep{KinBa15}. We also implemented the VFE approach of \cite{HenMatGha15} and include this in the comparison to the PEP methods. The VFE approach should be theoretically identical to PEP with small $\alpha$, however, we note that the results can be slightly different due to differences in the implementation -- optimisation for VFE vs.~the iterative PEP procedure and we also note that each step of PEP only gets to see a tiny fraction of each data point when $\alpha$ is small which can slow the learning speed. Similar to the regression experiment, we compare the methods using the pairwise ranking plots on the test error and negative log-likelihood (NLL) evaluation metrics.

\begin{landscape}
\begin{figure}[!ht]
    \centering
    \includegraphics[width=\linewidth]{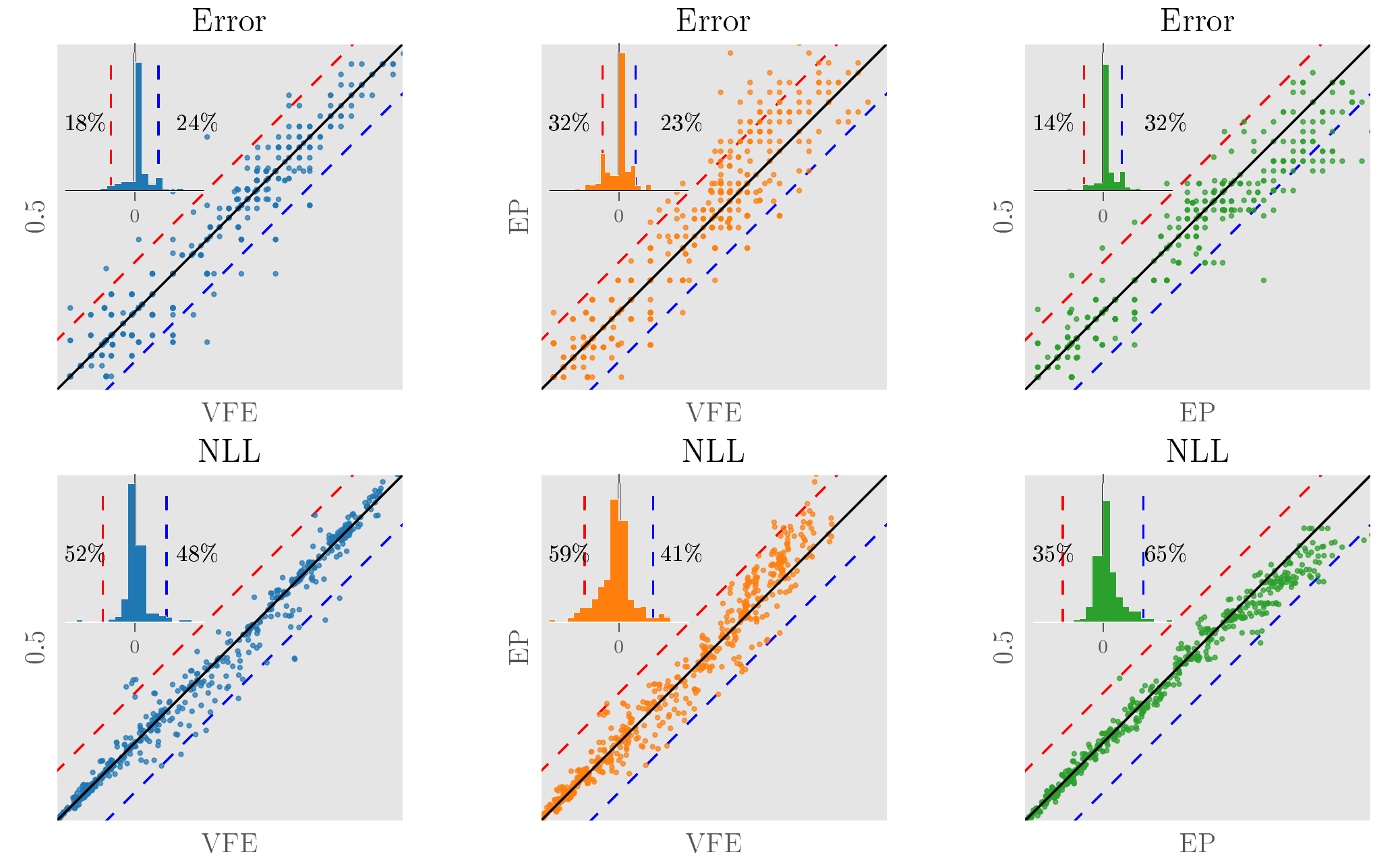}
	\caption{Pair-wise comparisons between Power EP with $\alpha=0.5$, EP ($\alpha=1$) and VFE ($\alpha\to 0$), evaluated on several classification datasets and various settings of $M$. Each coloured point is the result for one split. Points that are below the diagonal line illustrate the method on the $y$-axis is better than the method on the $x$-axis. The inset diagrams show the histograms of the difference between methods ($x$-value $-$ $y$-value), and the counts of negative and positive differences. Note that this indicates pairwise ranking of the two methods. Positive differences means the $y$-axis method is better than the $x$-axis method and vice versa. \label{fig:cla_scatter_all}}
\end{figure}
\end{landscape}

In \cref{fig:cla_scatter_all}, we plot (for each dataset, each split and each setting of $M$) the evaluation scores using one inference algorithm against the score obtained using another [see \cref{sec:reg_exp} for a detailed explanation of the plots]. In contrast to the regression results in \cref{sec:reg_exp}, there are no clear-cut winners among the methods. The test error results show that PEP $\alpha=0.5$ is marginally better than VFE and EP, while VFE edges EP out in this metric. Similarly, all methods perform comparably on the NLL scale, except with PEP $\alpha=0.5$ outperforming EP by a narrow magin (65\% of the time vs.~35\%) 

\begin{figure*}[t!]
    \centering
    \begin{subfigure}[t]{0.5\textwidth}
        \centering
        \includegraphics[trim={0.5cm 0.5cm 0.5cm 0cm},clip,width=\textwidth]{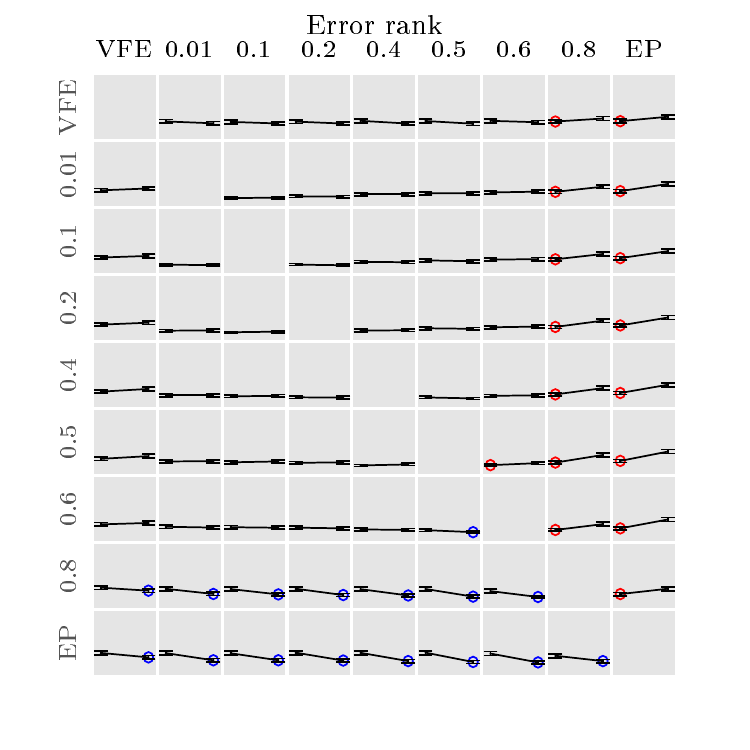}
    \end{subfigure}%
    ~ 
    \begin{subfigure}[t]{0.5\textwidth}
        \centering
        \includegraphics[trim={0.5cm 0.5cm 0.5cm 0cm},clip,width=\textwidth]{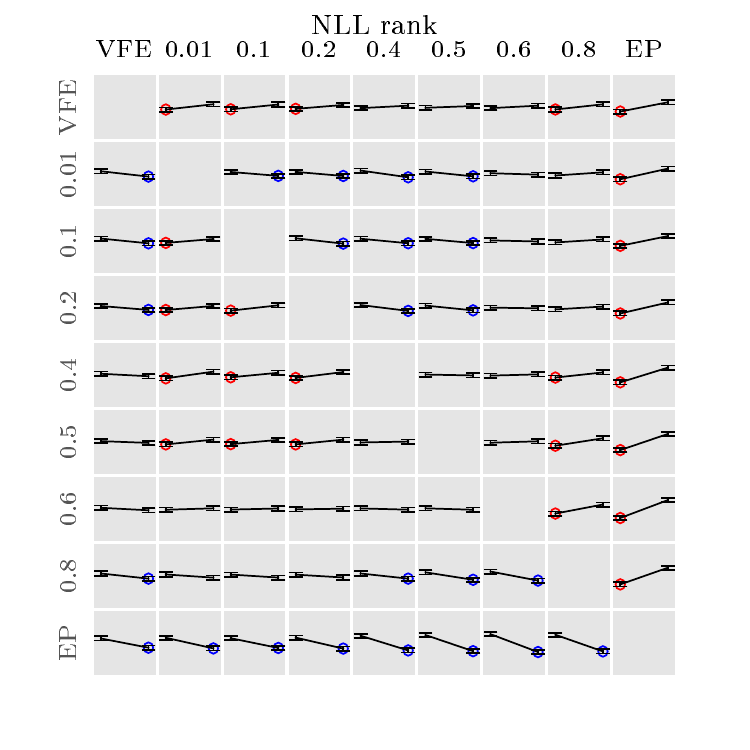}
    \end{subfigure}
    \includegraphics[trim={0.5cm 0cm 0cm 0cm},clip,width=\textwidth]{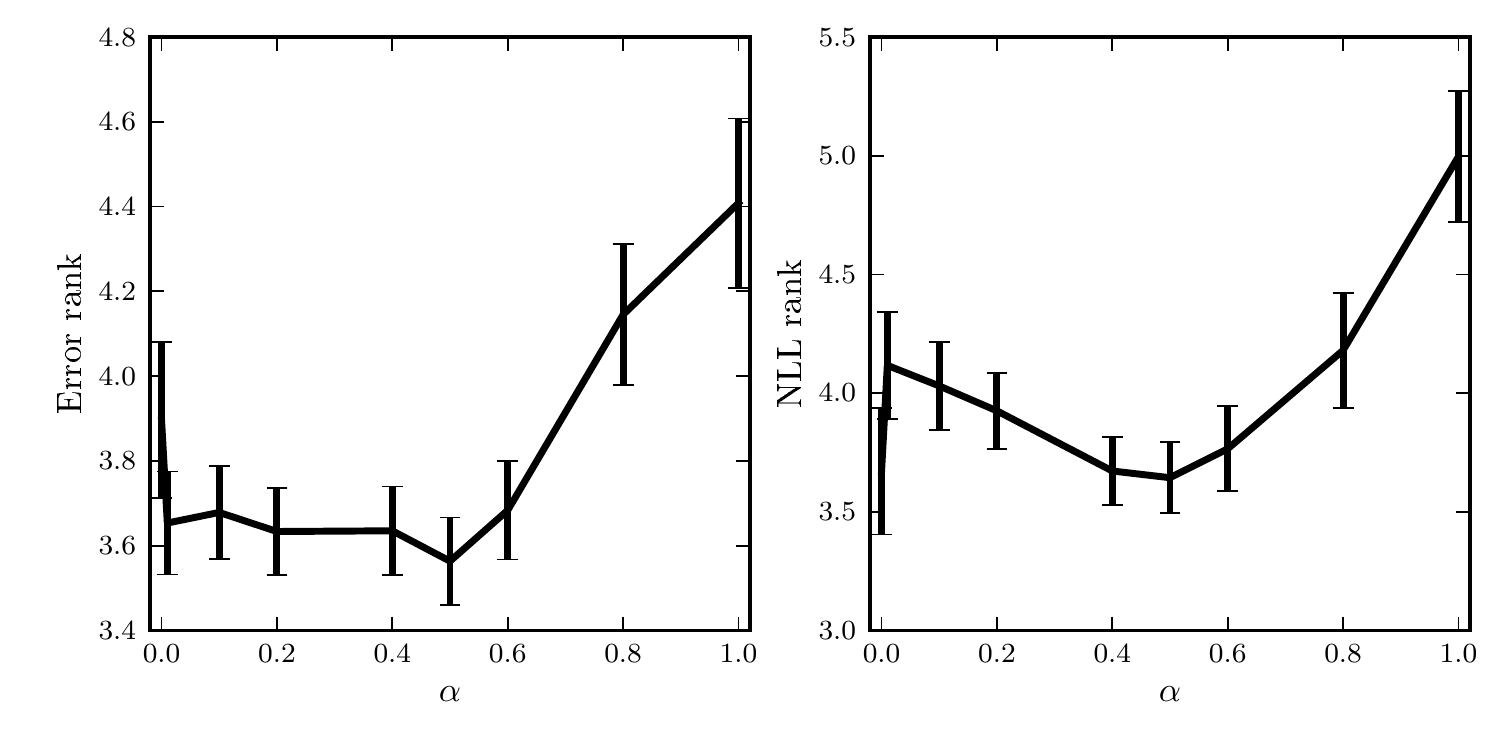}
    \caption{Average ranking of various $\alpha$ values in the classification experiment, lower is better. Top plots show the pairwise comparisons. Red circles denote rows being better than the corresponding columns, and blue circles mean vice versa. Bottom plots show the ranks of all methods when being compared together. Intermediate $\alpha$ values (not EP or VFE) are best on average.\label{fig:cla_rank}}
\end{figure*}

We repeat the pairwise comparison above to all methods and show the results in \cref{fig:cla_rank}. The plots show that there is no conlusive winner on the test error metric, and VFE, PEP $\alpha=0.4$ and PEP $\alpha=0.5$ have a slight edge over other $\alpha$ values on the NLL metric. Notably, methods corresponding to bigger $\alpha$ values, such as PEP $\alpha=0.8$ and EP, are outperformed by all other methods. Similar to the regression experiment, we observe the same pattern of results when all methods are simultaneously compared, as shown in \cref{fig:cla_rank}. However, the large errorbars suggest the difference between the methods is small in both metrics.



There is some variability between individual datasets, but the general trends are clear and consistent with the pattern noted above. For test error, PEP $\alpha=0.5$ is better than VFE on 1/6 dataset and is better than EP on 3/6 datasets (the differences on the other datasets are small). VFE outperforms EP on 2/6 datasets, while EP beats VFE on only 1/6 datasets. For NLL, PEP $\alpha=0.5$ only clearly outperforms VFE on 1/6 dataset, but is worse compared to VFE on 1 dataset (the other 4 datasets have no clear winner). PEP $\alpha=0.5$ is better than EP on 5/6 datasets and EP is better on the remaining dataset). EP is only better than VFE on 2/6 datasets, and is outperformed by VFE on the other 4/6 datasets. The finding that PEP and VFE are slightly better than EP on the NLL metric is surprising as we expected EP perform the best on the uncertainty sensitive metric (just as was discovered in the regression case). The full results are included in the appendices (see figs 25, 26 and 27). Similar to the regression case, we observe that as $M$ increases, the performance tends to be better for all methods and the differences between the methods tend to become smaller, but we have not found evidence for systematic sensitivity to the nature of the approximation.

In summary, we make the following recommendations based on these results for GP classification problems. For a raw test error loss and for NLL, we recommend using $\alpha=0.5$ (or  $\alpha=0.4$). It is possible that more fine grained recommendations are possible based upon details of the dataset and the computational resources available for processing, but further work will be needed to establish this.

%
%
%
\section{Discussion}

It is difficult to identify precisely where the best approximation methods derive their advantages, but here we will speculate. Since the negative variational free-energy is a lower-bound on the log-marginal likelihood it has the enviable theoretical guarantee that pseudo-input optimisation is always guaranteed to improve the estimate of the log marginal likelihood and the posterior (as measured by the inclusive KL). The negative EP energy, in contrast, is not generally a lower bound which can mean that pseudo-input optimisation drives the solution to the point where the EP energy over-estimates the log marginal likelihood the most, rather than to the point where the marginal likelihood and/or posterior estimate is best. For this reason, we believe that variational methods are likely to be better than EP if the goal is to derive accurate marginal likelihood estimates, or accurate predictive distributions, for fixed hyper-parameter settings. For hyper-parameter optimisation, things are less clear cut since variational methods are biased away from the maximal marginal likelihood, towards hyper-parameter settings for which the posterior approximation is accurate. Often this bias is severe and also creates local-optima \citep{TurSah11}. So, although EP will generally also be biased away from the maximal marginal likelihood and potentially towards areas of over-estimation, it can still outperform variational methods. Superposed onto these factors, is a general trend for variational methods to minimise MSE / classification error-rate and EP methods to minimise negative log-likelihood, due to the form of their respective energies (the variational free-energy includes the average training MSE in the regression case, for example). Intermediate methods will blend the strengths and weaknesses of the two extremes. It is interesting that values of $\alpha$ around a half are arguably the best performing on average. Similar empirical conclusions have been made elsewhere \cite{Min05, HerLiRowHerBuiTur16, DepHerDosUdl16}. In this case, the alpha-divergence interpretation of Power EP shows that it is minimising the Hellinger distance whose square root is a valid distance metric. Further experimental and theoretical work is required to clarify these issues.

The results presented above employed (approximate) type-II maximum likelihood fitting of the hyper-parameters.  This estimation method is known in some circumstances to overfit the data. It is therefore conceivable therefore that pseudo-point approximations, which have a tendency to encourage under-fitting due to their limited representational capacity, could be beneficial due to them mitigating overfitting. We do not believe that this is a strong effect in the experiments above. For example, in the synthetic data experiments the NLML, SMSE and SMLL obtained from fitting the unapproximated GP were similar to those obtained using the GP from which the data were generated, indicating that overfitting is not a strong effect (see fig.~9 in the appendix). It is true that EP and $\alpha=0.8$ over-estimates the marginal likelihood in the synthetic data experiments, but this is a distinct effect from over-fitting which would, for example, result in overconfident predictions on the test dataset. The SMSE and SMLL on the training and test sets, for example, are similar which is indicative of a well-fit model. It would be interesting to explore distributional hyper-parameter estimates \citep[see e.g.][]{PiiVeh17} that employ these pseudo-point approximations.

One of the features of the approximate generative models introduced in section \ref{sec:genmod} for regression, is that they contain input-dependent noise, unlike the original model. Many datasets contain noise of this sort and so approximate models like FITC and PITC, or models in which the observation noise is explicitly modelled are arguably more appropriate than the original unapproximated regression model \citep{Sne07,SauHenVehLaw16}. Motivated by this train of reasoning, \citet{Tit09a} applied the variational free-energy approximation to the FITC generative model an approach that was later generalised by \citet{HoaHoaLow16} to encompass a more general class of input dependent noise, including Markov structure \citep{LowYuCheJei15}. Here the insight is that the resulting variational lower bound separates over data points \citep{HenFusLaw13} and is, therefore, amenable to stochastic optimisation using minibatches unlike the marginal likelihood. 
In a sense, these approaches unify the approximate generative modelling approach, including the FITC and PITC variants, with the variational free-energy methods. Indeed, one approach is to posit the desired form of the optimal variational posterior, and to work backwards from this to construct the generative model implied \citep{HoaHoaLow16}.
However, these approaches are quite different from the one described in this paper where FITC and PITC are shown to emerge in the context of approximating the original unapproximated GP regression model using Power EP. Indeed, if the goal really is to model input dependent noise, it is not at all clear that generative models like FITC are the most sensible. For example, FITC uses a single set of hyper-parameters to describe the variation of the underlying function and the input dependent noise.

\section{Conclusion}
This paper provided a new unifying framework for GP pseudo-point approximations based on Power EP that subsumes many previous approaches including FITC, PITC, DTC, Titsias's VFE method, Qi et al's EP method, and inter-domain variants. It provided a clean computational perspective on the seminal work of Csat\'{o} and Opper that related FITC to EP, before extending their analysis significantly to include a closed form Power EP marginal likelihood approximation for regression, connections to PITC, and further results on classification and GPSSMs. The new framework was used to devise new algorithms for GP regression and GP classification. Extensive experiments indicate that intermediate values of Power EP with the power parameter set to $\alpha=0.5$ often outperform the state-of-the-art EP and VFE approaches. The new framework suggests many interesting directions for future work in this area that we have not explored, for example, extensions to online inference, combinations with special structured matrices (e.g.~circulant and Kronecker structure), Bayesian hyper-parameter learning, and applications to richer models. The current work has only scratched the surface, but we believe that the new framework will form a useful theoretical foundation for the next generation of GP approximation schemes.

\acks{The authors would like to thank Prof.~Carl Edward Rasmussen, Nilesh Tripuraneni, Matthias Bauer, James Hensman, and Hugh Salimbeni for insightful comments and discussion.
TDB thanks Google for funding his European Doctoral Fellowship.
RET thanks EPSRC grants EP/G050821/1, EP/L000776/1 and EP/M026957/1.}

\bookmarksetup{startatroot}
\appendix
\section{A unified objective for un-normalised KL variational free-energy methods} \label{app:KL}

Here we show that performing variational inference by optimising the un-normalised KL naturally leads to a single objective for both the approximation to the joint distribution, $q^*(f|\theta)$ and the hyper-parameters $\theta$. 

The un-normalised KL is given by
\begin{align}
\overline{\mathrm{KL}}( q^*(f|\theta) ||  p(f,\yvec| \theta) ) = \int q^*(f|\theta) \log \frac{q^*(f|\theta)}{p(f,\yvec| \theta) } \mathrm{d}f + \int \left ( p(f,\yvec| \theta) - q^*(f|\theta) \right)  \mathrm{d}f.
\end{align}
This is intractable as it includes the marginal likelihood $p(\yvec| \theta)  = \int p(f,\yvec| \theta) \mathrm{d}f$. However, since we are interested in minimising this objective with respect to $q^*(f|\theta)$ we can ignore the intractable term,
\begin{align}
\argmin_{q^*(f|\theta)} \overline{\mathrm{KL}}( q^*(f|\theta) ||  p(f,\yvec| \theta) ) &= \argmax_{q^*(f|\theta)} \left ( p(\yvec|\theta) - \overline{\mathrm{KL}}( q^*(f|\theta) ||  p(f,\yvec| \theta) ) \right) \\ &= \argmax_{q^*(f|\theta)}  \left( \int q^*(f|\theta) \log \frac{p(f,\yvec| \theta)}{q^*(f|\theta) } \mathrm{d}f +  \int q^*(f|\theta)  \mathrm{d}f \right).
\end{align}
In other words, we have turned the unnormalised KL into a tractable lower-bound of the marginal likelihood $\mathcal{G}(q^*(f|\theta),\theta) = p(\yvec|\theta) - \overline{\mathrm{KL}}( q^*(f|\theta) ||  p(f,\yvec| \theta) ) $. The structure of this new lower-bound can be understood by decomposing the approximation to the joint distribution into a normalised posterior approximation $q(f|\theta)$  and an approximation to the marginal likelihood, $Z_{\mathrm{VFE}}$, that is $q^*(f|\theta) = Z_{\mathrm{VFE}} q(f|\theta)$.
\begin{align}
\mathcal{G}(Z_{\mathrm{VFE}} q(f|\theta),\theta)  =   Z_{\mathrm{VFE}} \left (1 - \log Z_{\mathrm{VFE}} +   \int q(f|\theta)  \log \frac{p(f,\yvec| \theta)}{q(f|\theta) } \mathrm{d}f \right)
\end{align}
We can see that optimising the lower-bound with respect to $\theta$ is equivalent to optimising the standard variational free-energy $\mathcal{F}(q(f|\theta), \theta) = \int q(f|\theta)  \log \frac{p(f,\yvec| \theta)}{q(f|\theta) } \mathrm{d}f$. Moreover, optimising for $Z_{\mathrm{VFE}}$ recovers $Z_{\mathrm{VFE}}^{\mathrm{opt}} = \exp(\mathcal{F}(q(f|\theta), \theta))$. Substituting this back into the bound 
\begin{align}
\mathcal{G}(Z^{\mathrm{opt}}_{\mathrm{VFE}} q(f|\theta),\theta)  =  Z_{\mathrm{VFE}}^{\mathrm{opt}} = \exp(\mathcal{F}(q(f|\theta), \theta)).
\end{align}
In other words, the new collapsed bound is just the exponential of the original variational free-energy and optimising the collapsed bound for $\theta$ is equivalent to optimising the approximation to the marginal likelihood.

\section{Global and local inclusive KL minimisations}
\label{app:inclusive_kl}
In this section, we will show that optimising a single global inclusive KL-divergence, $\mathrm{KL}(q||p)$, is equivalent to optimising a sum of a set of local inclusive KL-divergence, $\mathrm{KL}(q||\tilde{p})$, where $p$, $q$ and $\tilde{p}$ are the exact posterior, the approximate posterior and the tilted distribution accordingly. Without loss of generality, we assume that $p(\theta) = \prod_{n} f_n(\theta) \approx \prod_{n} t_n(\theta) = q(\theta)$, that is the exact posterior is a product of factors, $\{f_{n}(\theta)\}_{n}$, each of which is approximated by an approximate factor $t_n(\theta)$. Substituting these distributions into the global KL-divergence gives,
\begin{align}
\mathrm{KL}(q(\theta) || p(\theta)) &= \int \dd \theta q(\theta) \log \frac{q(\theta)}{p(\theta)}\nonumber\\
&= \int \dd \theta q(\theta) \log \frac{\prod_{n} t_n(\theta) }{\prod_{n} f_n(\theta)}\nonumber\\
&= \int \dd \theta q(\theta) \log \bigg[ \frac{\prod_{n} t_n(\theta) }{\prod_{n} f_n(\theta)} \frac{\prod_{n} \prod_{i\neq n} t_i(\theta)}{\prod_{n} \prod_{i\neq n} t_i(\theta)} \bigg] \nonumber\\
&= \int \dd \theta q(\theta) \log \frac{\prod_{n} [\prod_{i} t_i(\theta)]}{\prod_{n} [f_n(\theta) \prod_{i\neq n} t_i(\theta)]} \nonumber\\
&= \sum_n \int \dd \theta q(\theta) \log \frac{\prod_{i} t_i(\theta)}{[f_n(\theta) \prod_{i\neq n} t_i(\theta)}\nonumber\\
&= \sum_n \mathrm{KL} (q(\theta) || \tilde{p}_n(\theta)),
\end{align}
which means running the EP procedure, where we use $\mathrm{KL} (q(\theta) || \tilde{p}_n(\theta))$ in place of $\mathrm{KL} (\tilde{p}_n(\theta) || q(\theta))$, is {\it equivalent} to the VFE approach which optimises a single global KL-divergence, $\mathrm{KL}(q(\theta)||p(\theta))$.

\section{Some relevant linear algebra and function expansion identities}

The Woodbury matrix identity or Woodbury formula is:
\begin{align}
(A + UCV)^{-1} = A^{-1} - A^{-1} U ( C^{-1} + V A^{-1} U )^{-1} V A^{-1}. \label{t1}
\end{align}

In general, $C$ need not be invertible, we can use the Binomial inverse theorem,
\begin{align}
(A + UCV)^{-1} = A^{-1} - A^{-1} U C ( C + C V A^{-1} U C )^{-1} C V A^{-1}. \label{t2}
\end{align}

When $C$ is an identity matrix and $U$ and $V$ are vectors, the Woodbury identity can be shortened and become the Sherman-Morrison formula,
\begin{align}
(A + uv^\intercal)^{-1} = A^{-1} - \frac { A^{-1} u v^\intercal A^{-1} } { 1 + v^\intercal A^{-1} u }. \label{t3}
\end{align}

Another useful identity is the matrix determinant lemma,
\begin{align}
\mathrm{det}(A + uv^\intercal) =  ( 1 + v^\intercal A^{-1} u ) \mathrm{det}(A). \label{t4}
\end{align}

The above theorem can be extend for matrices $U$ and $V$,
\begin{align}
\mathrm{det}(A + UV^\intercal) =  \mathrm{det} ( \mathrm{I} + V^\intercal A^{-1} U ) \mathrm{det} (A). \label{t5}
\end{align}

We also make use of the following Maclaurin series,
\begin{align}
\exp(x) &= 1 + x + \frac{x^2}{2!} + \frac{x^3}{3!} + \cdots \label{expx}\\
\text{and}\;\; \log(1+x) &= x - \frac{x^2}{2} + \frac{x^3}{3} + \cdots.\label{logx}
\end{align}

\section{KL minimisation between Gaussian processes and moment matching}
The difficult step of Power-EP is the projection step, that is how to find the posterior approximation $q(f)$ that minimises the KL divergence, $\mathrm{KL} ( \tilde{p} (f) || q(f))$, where $\tilde{p}(f)$ is the tilted distribution. We have chosen the form of the approximate posterior
\begin{align}
q(f) = p( f_{ \neq \uvec } | \uvec ) q(\uvec) = p( f_{ \neq \uvec } | \uvec ) \frac{ \exp(\theta_\uvec^\intercal \phi(\uvec)) }{ \mathcal{Z}(\theta_\uvec) },
\end{align}
where $\mathcal{Z}(\theta_\uvec) = \int \exp(\theta_\uvec^\intercal \phi(\uvec)) \dd \uvec$ to ensure normalisation. We can then write the KL minimisation objective as follows,
\begin{align}
\mathcal{F}_\mathrm{KL} 
&= \mathrm{KL} ( \tilde{p} (f) || q (f) )\\
&= \int \tilde{p} (f) \log \frac{\tilde{p} (f)}{q (f)} \dd f\\
&= \langle \log \tilde{p} (f) \rangle_{\tilde{p} (f)} - \langle \log p( f_{ \neq \uvec } | \uvec ) \rangle_{\tilde{p} (f)} - \theta_\uvec^\intercal \langle \phi(\uvec) \rangle_{\tilde{p} (f)} + \log \mathcal{Z}(\theta_\uvec).
\end{align}
Since $p( f_{ \neq \uvec } | \uvec )$ is the prior conditional distribution, the only free parameter that controls our posterior approximation is $\theta_\uvec$. As such, to find $\theta_\uvec$ that minimises $F_\mathrm{KL}$, we find the gradient of $F_\mathrm{KL}$ w.r.t~$\theta_\uvec$ and set it to zero,
\begin{align}
0 = \frac{ \dd \mathcal{F}_\mathrm{KL} } { \dd \theta_\uvec } 
&= - \langle \phi(\uvec) \rangle_{\tilde{p} (f)} + \frac{ \dd  \log \mathcal{Z}(\theta_\uvec) } { \dd \theta_\uvec } \\
&= - \langle \phi(\uvec) \rangle_{\tilde{p} (f)} + \langle \phi(\uvec) \rangle_{q (u)},
\end{align}
therefore, $\textcolor{NavyBlue}{\langle \phi(\uvec) \rangle_{\tilde{p} (f)} = \langle \phi(\uvec) \rangle_{q (u)}}$. That is, though we are trying to perform the KL minimisation between two Gaussian processes, due to the special form of the posterior approximation, {\it \textcolor{NavyBlue}{it is sufficient to only match the moments at the inducing points $\uvec$}}.\footnote{We can show that this condition gives the minimum of $\mathcal{F}_{\mathrm{KL}}$ by computing the second derivative.}

\section{Shortcuts to the moment matching equations\label{sec:mm}}
The most crucial step in Power-EP is the moment matching step as discussed above. This step can be done analytically for the Gaussian case, as the mean and covariance of the approximate posterior can be linked to the cavity distribution as follows,
\begin{align}
\mathbf{m}_{\uvec} &= \mathbf{m}_{\uvec}^{\setminus n} + \mathbf{V}_{\uvec f}^{\setminus n} \frac{\dd \log \mathcal{Z}_{\mathrm{tilted}, n}}{\dd \mathbf{m}_{f}^{\setminus n}}, \\
\mathbf{V}_{\uvec} &= \mathbf{V}_{\uvec}^{\setminus n} + \mathbf{V}_{\uvec f}^{\setminus n} \frac{\dd^2 \log \mathcal{Z}_{\mathrm{tilted}, n}}{\dd \mathbf{m}_{f}^{\setminus n, 2}} \mathbf{V}_{f\uvec}^{\setminus n},
\end{align}
where $\mathcal{Z}_{\mathrm{tilted}, n}$ is the normaliser of the tilted distribution,
\begin{align}
\mathcal{Z}_{\mathrm{tilted}, n} 
&= \int q^{\setminus n}(f) p(y_n|f) \dd f\\
&= \int q^{\setminus n}(f) p(y_n|f_n) \dd f\\
&= \int q^{\setminus n}(f_n) p(y_n|f_n) \dd f_n.
\end{align}
In words, $\mathcal{Z}_{\mathrm{tilted}, n}$ only depends on the marginal distribution of the cavity process, $q^{\setminus n}(f_n)$, simplifying the moment matching equations above,
\textcolor{RubineRed}{
\begin{align}
\mathbf{m}_{\uvec} &= \mathbf{m}_{\uvec}^{\setminus n} + \mathbf{V}_{\uvec f_n}^{\setminus n} \frac{\dd \log \mathcal{Z}_{\mathrm{tilted}, n}}{\dd m_{f_n}^{\setminus n}}, \label{eq:mm1} \\
\mathbf{V}_{\uvec} &= \mathbf{V}_{\uvec}^{\setminus n} + \mathbf{V}_{\uvec f_n}^{\setminus n} \frac{\dd^2 \log \mathcal{Z}_{\mathrm{tilted}, n}}{\dd m_{f_n}^{\setminus n, 2}} \mathbf{V}_{f_n\uvec}^{\setminus n}. \label{eq:mm2}
\end{align}
}

We can rewrite the cross-covariance $\mathbf{V}_{\uvec f_n}^{\setminus n} = \mathbf{V}_{\uvec}^{\setminus n} \mathbf{K}_{\uvec\uvec}^{-1} \mathbf{K}_{\uvec f_n}$. We also note that, $m_{f_n}^{\setminus n} = \mathbf{K}_{f_n\uvec}\mathbf{K}_{\uvec\uvec}^{-1} \mathbf{m}_\uvec^{\setminus n}$, resulting in,
\begin{align}
\frac{\dd \log \mathcal{Z}_{\mathrm{tilted}, n}}{\dd \mathbf{m}_{\uvec}^{\setminus n}} &= \frac{\dd \log \mathcal{Z}_{\mathrm{tilted}, n}}{\dd m_{f_n}^{\setminus n}} \mathbf{K}_{\uvec\uvec}^{-1} \mathbf{K}_{\uvec f_n},\\
\frac{\dd \log \mathcal{Z}_{\mathrm{tilted}, n}}{\dd \mathbf{V}_{\uvec}^{\setminus n}} &= \mathbf{K}_{\uvec\uvec}^{-1} \mathbf{K}_{\uvec f_n} \frac{\dd^2 \log \mathcal{Z}_{\mathrm{tilted}, n}}{\dd m_{f_n}^{\setminus n, 2}} \mathbf{K}_{f_n\uvec }  \mathbf{K}_{\uvec\uvec}^{-1}.
\end{align}
Substituting these results back in eqs.~\ref{eq:mm1} and \ref{eq:mm2}, we obtain
\textcolor{NavyBlue}{
\begin{align}
\mathbf{m}_{\uvec} &= \mathbf{m}_{\uvec}^{\setminus n} + \mathbf{V}_{\uvec}^{\setminus n} \frac{\dd \log \mathcal{Z}_{\mathrm{tilted}, n}}{\dd \mathbf{m}_\uvec^{\setminus n}}, \label{eq:mm3}\\
\mathbf{V}_{\uvec} &= \mathbf{V}_{\uvec}^{\setminus n} + \mathbf{V}_{\uvec}^{\setminus n} \frac{\dd^2 \log \mathcal{Z}_{\mathrm{tilted}, n}}{\dd \mathbf{m}_\uvec^{\setminus n, 2}} \mathbf{V}_{\uvec}^{\setminus n}.\label{eq:mm4}
\end{align}
}

Therefore, using eqs.~\ref{eq:mm1} and \ref{eq:mm2}, or eqs.~\ref{eq:mm3} and \ref{eq:mm4} are equivalent in our approximation settings. 

\section{Full derivation of the Power-EP procedure\label{app:full_deriv}}
We provide the full derivation of the Power-EP procedure in this section. We follow the derivation in \citep{QiAbdMin10} closely, but provide a clearer exposition and details how to get to each step used in the implementation, and how to handle powered/fractional deletion and update in Power-EP.  
\subsection{Optimal factor parameterisation}

We start by defining the approximate factors to be in natural parameter form as this makes it simple to combine and delete them, $t_n(\uvec) = \normnat(\uvec; z_n,\mathbf{T}_{1, n}, \mathbf{T}_{2, n}) = z_n \exp( \uvec^\intercal\mathbf{T}_{1, n} - \frac{1}{2} \uvec^\intercal \mathbf{T}_{2, n} \uvec)$. We initially consider full rank $\mathbf{T}_{2, n}$, but will show that the optimal form is rank 1.

The next goal is to relate these parameters to the approximate GP posterior. The approximate posterior over the pseudo-outputs has natural parameters $\mathbf{T}_{1, \uvec} = \sum_n \mathbf{T}_{1, n}$ and $\mathbf{T}_{2, \uvec} = \mathbf{K}_{\mathbf{uu}}^{-1} + \sum_n \mathbf{T}_{2, n}$. This induces an approximate GP posterior with mean and covariance function,
\begin{align}
m_\mathrm{f} &= \mathbf{K}_{\mathrm{f}\uvec}\mathbf{K}_{\uvec\uvec}^{-1} \mathbf{T}_{2, \uvec}^{-1} \mathbf{T}_{1, \uvec} = \mathbf{K}_{\mathrm{f}\uvec} \gamma \\
V_\mathrm{ff'} &= \mathbf{K}_{\mathrm{f}\mathrm{f}'} - \mathbf{Q}_{\mathrm{f}\mathrm{f}'} + \mathbf{K}_{\mathrm{f}\uvec}\mathbf{K}_{\uvec\uvec}^{-1} \mathbf{T}_{2, \uvec}^{-1} \mathbf{K}_{\uvec\uvec}^{-1}\mathbf{K}_{\uvec\mathrm{f'}} = \mathbf{K}_{\mathrm{f}\mathrm{f}'} - \mathbf{K}_{\mathrm{f}\uvec} \beta \mathbf{K}_{\uvec\mathrm{f'}}.
\end{align}
where $\gamma$ and $\beta$ are likelihood-dependent terms we wish to store and update using PEP; $\gamma$ and $\beta$ fully specify the approximate posterior.

{\bf Deletion step:} The cavity for data point $n$, $q^{\setminus n}(f) \propto  q^*(f) / t_n^\alpha(\uvec)$, has a similar form to the posterior, but the natural parameters are modified by the deletion, $\mathbf{T}^{\setminus n}_{1, \uvec} = \mathbf{T}_{1, \uvec} - \alpha\mathbf{T}_{1, n}$ and $\mathbf{T}^{\setminus n}_{2, \uvec} = \mathbf{T}_{2, \uvec} - \alpha\mathbf{T}_{2, n} $, yielding a new mean and covariance function
\begin{align}
    m_\mathrm{f}^{\setminus n} &= \mathbf{K}_{\mathrm{f}\uvec}\mathbf{K}_{\uvec\uvec}^{-1} \mathbf{T}^{\setminus n, -1}_{2, \uvec} \mathbf{T}^{\setminus n}_{1, \uvec} = \mathbf{K}_{\mathrm{f}\uvec} \gamma^{\setminus n}\\
V_\mathrm{ff'}^{\setminus n} &= \mathbf{K}_{\mathrm{f}\mathrm{f}'} - \mathbf{Q}_{\mathrm{f}\mathrm{f}'} + \mathbf{K}_{\mathrm{f}\uvec}\mathbf{K}_{\uvec\uvec}^{-1} \mathbf{T}{\setminus n, -1}_{2, \uvec} \mathbf{K}_{\uvec\uvec}^{-1}\mathbf{K}_{\uvec\mathrm{f'}} = \mathbf{K}_{\mathrm{f}\mathrm{f}'} - \mathbf{K}_{\mathrm{f}\uvec} \beta^{\setminus n} \mathbf{K}_{\uvec\mathrm{f'}}.
\end{align}
{\bf Projection step:} The central step in Power EP is the projection step. Obtaining the new approximate unormalised posterior $q^*(f)$ such that $\mathrm{KL}(\tilde{p}(f) || q^*(f))$ is minimised would na\"ively appear intractable. Fortunately, as shown in the previous section, because of the structure of the approximate posterior, $q(f) = p(f_{\neq \uvec}|\uvec) q(\uvec)$, the objective, $\mathrm{KL}(\tilde{p}(f) || q^*(f))$ is minimised when $\textcolor{NavyBlue}{\E_{\tilde{p} (f)} [\phi(\uvec)] = \E_{q (u)} [\phi(\uvec)]}$, where $\phi(\uvec)$ are the sufficient statistics, that is when the moments at the pseudo-inputs are matched.
This is the central result from which computational savings are derived. Furthermore, this moment matching condition would appear to necessitate computation of a set of integrals to find the zeroth, first and second moments. Using results from the previous section simplifies and provides the following shortcuts,
\begin{align}
    \mathbf{m}_\uvec &= \mathbf{m}^{\setminus n}_{\uvec} + \mathbf{V}^{\setminus n}_{\uvec \mathrm{f}_n} \frac{\dd \log \tilde{Z}_n}{\dd m^{\setminus n}_{\mathrm{f}_n}} \label{eqn:pepmm1}\\
    \mathbf{V}_\uvec &= \mathbf{V}^{\setminus n}_{\uvec} + \mathbf{V}^{\setminus n}_{\uvec \mathrm{f}_n} \frac{\dd^2 \log \tilde{Z}_n}{\dd (m^{\setminus n}_{\mathrm{f}_n} )^2} \mathbf{V}^{\setminus n}_{\mathrm{f}_n\uvec}. \label{eqn:pepmm2}
\end{align}
where $\log \tilde{Z}_n = \log \E_{q^{\setminus n}(f)} [p^\alpha(y_n|\mathrm{f}_n)]$ is the log-normaliser of the tilted distribution.

{\bf Update step:} Having computed the new approximate posterior, the fractional approximate factor $t_{n,\mathrm{new}}(\uvec) = q^*(f) / q^{\setminus n}(f)$ can be straightforwardly obtained, resulting in,
\begin{align}
\mathbf{T}_{1,n,\mathrm{new}} &= \mathbf{V}_\uvec^{-1}\mathbf{m}_\uvec - \mathbf{V}^{\setminus n, -1}_\uvec\mathbf{m}^{\setminus n}_\uvec \label{eqn:update_mean}\\
\mathbf{T}_{2,n,\mathrm{new}} &= \mathbf{V}_\uvec^{-1} - \mathbf{V}^{\setminus n, -1}_\uvec \label{eqn:update_cov}\\
z_n^\alpha &= \tilde{Z}_n \exp(\mathcal{G}_{q_*^{\setminus n}(\uvec)} - \mathcal{G}_{q^*(\uvec)}),
\end{align}
where $\mathcal{G}_{\normnat(\uvec; z,\mathbf{T}_{1}, \mathbf{T}_{2})} = \int \normnat(\uvec; z,\mathbf{T}_{1}, \mathbf{T}_{2}) \dd\uvec$. Let $d_1 = \frac{\dd \log \tilde{Z}_n}{\dd m^{\setminus n}_{\mathrm{f}_n}}$ and $d_2 = \frac{\dd^2 \log \tilde{Z}_n}{\dd (m^{\setminus n}_{\mathrm{f}_n} )^2}$. Using \cref{t1} and \cref{eqn:pepmm2}, we have,
\begin{align}
\mathbf{V}_\uvec^{-1} - \mathbf{V}^{\setminus n, -1}_\uvec = - \mathbf{V}^{\setminus n, -1}_\uvec \mathbf{V}^{\setminus n}_{\uvec \mathrm{f}_n} \left[ d_2^{-1} + \mathbf{V}^{\setminus n}_{\mathrm{f}_n \uvec} \mathbf{V}^{\setminus n, -1}_\uvec \mathbf{V}^{\setminus n}_{\uvec \mathrm{f}_n} \right]^{-1} \mathbf{V}^{\setminus n}_{\mathrm{f}_n \uvec} \mathbf{V}^{\setminus n, -1}_\uvec \label{eqn:pep_inv1}
\end{align}
Let $v_n = \alpha (- d_2^{-1} - \mathbf{V}^{\setminus n}_{\mathrm{f}_n \uvec} \mathbf{V}^{\setminus n, -1}_\uvec \mathbf{V}^{\setminus n}_{\uvec \mathrm{f}_n})$, and $\mathbf{w}_n = \mathbf{V}^{\setminus n, -1}_\uvec \mathbf{V}^{\setminus n}_{\uvec \mathrm{f}_n}$. Combining \cref{eqn:pep_inv1} and \cref{eqn:update_cov} gives 
\begin{align}
\mathbf{T}_{2,n,\mathrm{new}} = \mathbf{w}_n \alpha v_n^{-1} \mathbf{w}_n^{\intercal} \label{eqn:pepT2update}
\end{align}
At convergence, we have $t_{n}(\uvec)^{\alpha} = t_{n, \mathrm{new}}(\uvec)$, hence $\mathbf{T}_{2,n} = \mathbf{w}_n v_n^{-1} \mathbf{w}_n^{\intercal}$. In words, $\mathbf{T}_{2,n}$ is  optimally a rank-1 matrix. Note that,
\begin{align}
\mathbf{w}_n 
&= \mathbf{V}^{\setminus n, -1}_\uvec \mathbf{V}^{\setminus n}_{\uvec \mathrm{f}_n} \\
&= (\mathbf{K}_{\uvec\uvec} - \mathbf{K}_{\uvec\uvec} \beta^{\setminus n} \mathbf{K}_{\uvec\uvec})^{-1} (\mathbf{K}_{\uvec\mathrm{f}_n} - \mathbf{K}_{\uvec\uvec} \beta^{\setminus n} \mathbf{K}_{\uvec\mathrm{f}_n})\\
&= \mathbf{K}_{\uvec\uvec}^{-1} (\mathrm{I} - \mathbf{K}_{\uvec\uvec} \beta^{\setminus n})^{-1} (\mathrm{I} - \mathbf{K}_{\uvec\uvec} \beta^{\setminus n}) \mathbf{K}_{\uvec\mathrm{f}_n}\\
&= \mathbf{K}_{\uvec\uvec}^{-1} \mathbf{K}_{\uvec\mathrm{f}_n}.
\end{align}
Using \cref{eqn:pepmm1} an \cref{eqn:pepT2update} gives,
\begin{align}
\mathbf{V}_\uvec^{-1}\mathbf{m}_\uvec 
&= (\mathbf{V}^{\setminus n, -1}_\uvec + \mathbf{w}_n \alpha v_n^{-1} \mathbf{w}_n^{\intercal}) (\mathbf{m}^{\setminus n}_{\uvec} + \mathbf{V}^{\setminus n}_{\uvec \mathrm{f}_n} d_1)\\
&= \mathbf{V}^{\setminus n, -1}_\uvec \mathbf{m}^{\setminus n}_{\uvec} + \mathbf{w}_n \alpha v_n^{-1} \mathbf{w}_n^{\intercal} \mathbf{m}^{\setminus n}_{\uvec}  + \mathbf{V}^{\setminus n, -1}_\uvec \mathbf{V}^{\setminus n}_{\uvec \mathrm{f}_n} d_1 + \mathbf{w}_n \alpha v_n^{-1} \mathbf{w}_n^{\intercal} \mathbf{V}^{\setminus n}_{\uvec \mathrm{f}_n} d_1
\end{align}
Substituting this result into \cref{eqn:update_mean},
\begin{align}
\mathbf{T}_{1,n,\mathrm{new}} 
&= \mathbf{V}_\uvec^{-1}\mathbf{m}_\uvec - \mathbf{V}^{\setminus n, -1}_\uvec\mathbf{m}^{\setminus n}_\uvec \\
&= \mathbf{w}_n \alpha v_n^{-1} \mathbf{w}_n^{\intercal} \mathbf{m}^{\setminus n}_{\uvec}  + \mathbf{V}^{\setminus n, -1}_\uvec \mathbf{V}^{\setminus n}_{\uvec \mathrm{f}_n} d_1 + \mathbf{w}_n \alpha v_n^{-1} \mathbf{w}_n^{\intercal} \mathbf{V}^{\setminus n}_{\uvec \mathrm{f}_n} d_1\\
&= \mathbf{w}_n  \alpha v_n^{-1} \left( \mathbf{w}_n^{\intercal} \mathbf{m}^{\setminus n}_{\uvec}  + d_1 v_n / \alpha + \mathbf{w}_n^{\intercal} \mathbf{V}^{\setminus n}_{\uvec \mathrm{f}_n} d_1 \right).
\end{align}
Let $\mathbf{T}_{1,n,\mathrm{new}} = \mathbf{w}_n \alpha v_n^{-1} g_n$, we obtain,
\begin{align}
g_n = -\frac{d_1}{d_2} + \mathbf{K}_{\mathrm{f}_n\uvec}\gamma^{\setminus n}.
\end{align}
At convergence, $\mathbf{T}_{1,n} = \mathbf{w}_n v_n^{-1} g_n$. Re-writing the form of the approximate factor using $\mathbf{T}_{1,n}$ and $\mathbf{T}_{2,n}$ at convergence,
\begin{align}
t_n(\uvec) &= \normnat(\uvec; z_n,\mathbf{T}_{1, n}, \mathbf{T}_{2, n}) \\
&= z_n \exp( \uvec^\intercal\mathbf{T}_{1, n} - \frac{1}{2} \uvec^\intercal \mathbf{T}_{2, n} \uvec)\\
&= z_n \exp( \uvec^\intercal\mathbf{w}_n v_n^{-1} g_n - \frac{1}{2} \uvec^\intercal  \mathbf{w}_n v_n^{-1} \mathbf{w}_n^{\intercal} \uvec)
\end{align}
As a result, the minimal and simplest way to parameterise the approximate factor is $t_{n}(\uvec) = \tilde{z}_n\norm(\mathbf{w}_n^{\intercal} \uvec; g_n, v_n) = \tilde{z}_n\norm(\mathbf{K}_{\mathrm{f}_n\uvec}\mathbf{K}_{\uvec\uvec}^{-1}\uvec; g_n, v_n)$, where $g_n$ and $v_n$ are scalars, resulting in a significant memory saving compared to the parameterisation using $\mathbf{T}_{1,n}$ and $\mathbf{T}_{2,n}$.

\subsection{Projection}
We now recall the update equations in the projection step (eqns.~\ref{eqn:pepmm1} and \ref{eqn:pepmm2}):

\begin{align}
    \mathbf{m}_\uvec &= \mathbf{m}^{\setminus n}_{\uvec} + \mathbf{V}^{\setminus n}_{\uvec \mathrm{f}_n} d_1,\\
    \mathbf{V}_\uvec &= \mathbf{V}^{\setminus n}_{\uvec} + \mathbf{V}^{\setminus n}_{\uvec \mathrm{f}_n} d_2 \mathbf{V}^{\setminus n}_{\mathrm{f}_n\uvec}.
\end{align}

Note that:
\begin{align}
\mathbf{m}_\uvec &= \mathbf{K}_{\uvec\uvec} \gamma, \\
\mathbf{V}_{\uvec} &= \mathbf{K}_{\uvec\uvec} - \mathbf{K}_{\uvec\uvec} \beta \mathbf{K}_{\uvec\uvec},
\end{align}
and
\begin{align}
\mathbf{m}_\uvec^{\setminus n} &= \mathbf{K}_{\uvec\uvec} \gamma^{\setminus n}, \\
\mathbf{V}_{\uvec}^{\setminus n} &= \mathbf{K}_{\uvec\uvec} - \mathbf{K}_{\uvec\uvec} \beta^{\setminus n} \mathbf{K}_{\uvec\uvec}.
\end{align}
Using these results, we can convert the update for the mean and covariance, $\mathbf{m}_\uvec$ and $\mathbf{V}_\uvec$, into an update for $\gamma$ and $\beta$,

\begin{align}
\gamma 
&= \mathbf{K}_{\uvec\uvec}^{-1} \mathbf{m}_\uvec \\
&= \mathbf{K}_{\uvec\uvec}^{-1} (\mathbf{m}_\uvec^{\setminus n} + \mathbf{V}^{\setminus n}_{\uvec \mathrm{f}_n} d_1)  \\
&= \gamma^{\setminus n} + \mathbf{K}_{\uvec\uvec}^{-1}  \mathbf{V}^{\setminus n}_{\uvec \mathrm{f}_n} d_1, \;\; \label{eqn:pepmmgamma}\text{and}\\
\beta
&= \mathbf{K}_{\uvec\uvec}^{-1} (\mathbf{K}_{\uvec\uvec} - \mathbf{V}_\uvec) \mathbf{K}_{\uvec\uvec}^{-1} \\
&= \mathbf{K}_{\uvec\uvec}^{-1} (\mathbf{K}_{\uvec\uvec} - \mathbf{V}^{\setminus n}_{\uvec} - \mathbf{V}^{\setminus n}_{\uvec \mathrm{f}_n} d_2 \mathbf{V}^{\setminus n}_{\mathrm{f}_n\uvec}) \mathbf{K}_{\uvec\uvec}^{-1} \\
&= \beta^{\setminus n} - \mathbf{K}_{\uvec\uvec}^{-1}\mathbf{V}^{\setminus n}_{\uvec \mathrm{f}_n} d_2 \mathbf{V}^{\setminus n}_{\mathrm{f}_n\uvec} \mathbf{K}_{\uvec\uvec}^{-1}\label{eqn:pepmmbeta}
\end{align} 

\subsection{Deletion step}

Finally, we present how deletion might be accomplished. One direct approach to this step is to divide out the cavity from the cavity, that is,
\begin{align}
q^{\setminus n}(f) \propto \frac{q(f)}{t_{n}^{\alpha}(\uvec)} = \frac{p(f_{\neq \uvec} | \uvec) q(\uvec)} {t_{n}^{\alpha}(\uvec)} = p(f_{\neq \uvec} | \uvec) q^{\setminus n}(\uvec).
\end{align}
Instead, we use an alternative using the KL minimisation as used in \citep{QiAbdMin10}, by realising that doing this will result in an identical outcome as the direct approach since the factor and distributions are Gaussian. Furthermore, we can re-use results from the projection and inclusion steps, by simply swapping the quantities and negating the site approximation variance. In particular, we present projection and deletion side-by-side, to facilitate the comparison, 
\begin{align}
    &\text{Projection:} \;\;\;\; q(f) \approx q^{\setminus n} (f) p(y_n|\mathrm{f}_n)& \\
    &\text{Deletion:} \;\;\;\;\;\; q^{\setminus n}(f) \propto q(f) \frac{1}{t_n^{\alpha}(\uvec)}&
\end{align}

The projection step minimises the KL between the LHS and RHS while moment matching, to get $q(f)$. We would like to do the same for the deletion step to find $q^{\setminus n}(f)$, and thus reuse the same moment matching results for $\gamma$ and $\beta$ with some modifications. 

Our task will be to reuse Equations \ref{eqn:pepmmgamma} and \ref{eqn:pepmmbeta}, the moment matching equations in $\gamma$ and $\beta$. We have two differences to account for. Firstly, we need to change any uses of the parameters of the cavity distribution to the parameters of the approximate posterior, $\mathbf{V}^{\setminus n}_{\uvec \mathrm{f}_n}$ to $\mathbf{V}_{\uvec \mathrm{f}_n}$, $\gamma^{\setminus n}$ to $\gamma$ and $\beta^{\setminus n}$ to $\beta$. This is the equivalent of re-deriving the entire projection operation while swapping the symbols (and quantities) for the cavity and the full distribution. Secondly, the derivatives $d_1$ and $d_2$ are different here, as 

\begin{align}
    \log \tilde{Z}_n &= \log \int{q(f) \frac{1}{t_n^{\alpha}(\uvec)}} \dd f
\end{align}

Now, we note

\begin{align}
    \frac{1}{t_n(\uvec)} &\propto \frac{1}{\mathcal{N}^{\alpha}(\mathbf{w}_n^\intercal \uvec; g_n, v_n)} \\
    &\propto \frac{1}{\exp{\left( -\frac{\alpha}{2} v_n^{-1} \left(\mathbf{w}_n^\intercal \uvec - g_n \right)^2 \right)}} \\
    &= \exp{\left( \frac{1}{2} \alpha v_n^{-1} \left(\mathbf{w}_n^\intercal \uvec - g_n \right)^2 \right)} \\
    &\propto \mathcal{N}(\mathbf{w}_n^\intercal \uvec; g_n, -v_n/\alpha)
\end{align}

Then we obtain the derivatives of $\log \tilde{Z}_n$

\begin{align}
    \tilde{d}_2 = \frac{\dd^2 \log \tilde{Z}_n} {\dd m_{\mathrm{f}_n}^2} &= - \left[ \mathbf{K}_{\mathrm{f}_n, \uvec} \mathbf{K}_{\uvec, \uvec}^{-1} \mathbf{K}_{\uvec, \mathrm{f}_n} - \mathbf{K}_{\mathrm{f}_n, \uvec} \beta \mathbf{K}_{\uvec, \mathrm{f}_n} - v_n/\alpha \right]^{-1} \\
    \tilde{d}_1 = \frac{\dd \log \tilde{Z}_n} {\dd m_{\mathrm{f}_n}} &= (\mathbf{K}_{\mathrm{f}_n, \uvec} \gamma - g_n) \tilde{d}_2
\end{align}

Putting the above results together, we obtain,

\begin{align}
\gamma^{\setminus n}
&= \gamma + \mathbf{K}_{\uvec\uvec}^{-1}  \mathbf{V}_{\uvec \mathrm{f}_n} \tilde{d}_1, \;\; \text{and}\\
\beta^{\setminus n}
&= \beta - \mathbf{K}_{\uvec\uvec}^{-1}\mathbf{V}_{\uvec \mathrm{f}_n} \tilde{d}_2 \mathbf{V}_{\mathrm{f}_n\uvec} \mathbf{K}_{\uvec\uvec}^{-1}
\end{align} 

\subsection{Summary of the PEP procedure}

We summarise here the key steps and equations that we have obtained, that are used in the implementation:

\begin{enumerate}
\item Initialise the parameters: $\{g_n = 0\}_{n=1}^{N}$, $\{v_n = \infty \}_{n=1}^{N}$, $\gamma = \mathbf{0}_{M\times1}$ and $\beta = \mathbf{0}_{M\times M}$

\item Loop through all data points until convergence:

\begin{enumerate}
\item Deletion step: find $\gamma^{\setminus n}$ and $\beta^{\setminus n}$
\begin{align}
\gamma^{\setminus n}
&= \gamma + \mathbf{K}_{\uvec\uvec}^{-1}  \mathbf{V}_{\uvec \mathrm{f}_n} \tilde{d}_1, \;\; \text{and}\\
\beta^{\setminus n}
&= \beta - \mathbf{K}_{\uvec\uvec}^{-1}\mathbf{V}_{\uvec \mathrm{f}_n} \tilde{d}_2 \mathbf{V}_{\mathrm{f}_n\uvec} \mathbf{K}_{\uvec\uvec}^{-1}
\end{align}

\item Projection step: find $\gamma$ and $\beta$
\begin{align}
\gamma 
&= \gamma^{\setminus n} + \mathbf{K}_{\uvec\uvec}^{-1}  \mathbf{V}^{\setminus n}_{\uvec \mathrm{f}_n} d_1,\\
\beta
&= \beta^{\setminus n} - \mathbf{K}_{\uvec\uvec}^{-1}\mathbf{V}^{\setminus n}_{\uvec \mathrm{f}_n} d_2 \mathbf{V}^{\setminus n}_{\mathrm{f}_n\uvec} \mathbf{K}_{\uvec\uvec}^{-1}
\end{align}

\item Update step: find $g_{n, \mathrm{new}}$ and $v_{n, \mathrm{new}}$
\begin{align}
g_{n, \mathrm{new}} &= -\frac{d_1}{d_2} + \mathbf{K}_{\mathrm{f}_n\uvec}\gamma^{\setminus n},\\
v_{n, \mathrm{new}} &= - d_2^{-1} - \mathbf{V}^{\setminus n}_{\mathrm{f}_n \uvec} \mathbf{V}^{\setminus n, -1}_\uvec \mathbf{V}^{\setminus n}_{\uvec \mathrm{f}_n}
\end{align}
and parameters for the full factor,

\begin{align}
v_n &\leftarrow (v^{-1}_{n, \mathrm{new}} + (1-\alpha)v^{-1}_{n} )^{-1}\\
g_n &\leftarrow v_n (g_{n, \mathrm{new}} v^{-1}_{n, \mathrm{new}} + (1-\alpha) g_{n} v^{-1}_{n})
\end{align}

\end{enumerate}

\end{enumerate}

\section{Power-EP energy for sparse GP regression and classification}

The Power-EP procedure gives an approximate marginal likelihood, which is the negative Power-EP energy, as follows,
\begin{align}
\mathcal{F} = \mathcal{G}(q_*(\uvec)) - \mathcal{G}(p_*(\uvec)) + \frac{1}{\alpha} \sum_{n} \left[ \log \mathcal{Z}_{\mathrm{tilted}, n} + \mathcal{G}(q_*^{\setminus n}(\uvec)) - \mathcal{G}(q_*(\uvec)) \right]
\end{align}

where $\mathcal{G}(q_*(\uvec))$ is the log-normaliser of the approximate posterior, that is,
\begin{align}
\mathcal{G}(q_*(\uvec))
&= \log \int p( f_{ \neq \uvec } | \uvec ) \exp(\theta_\uvec^\intercal \phi(\uvec)) \dd f_{ \neq \uvec } \dd \uvec\\
&= \log \int \exp(\theta_\uvec^\intercal \phi(\uvec)) \dd \uvec\\
&= \frac{M}{2} \log(2\pi) + \frac{1}{2} \log |\mathbf{V}| + \frac{1}{2} \mathbf{m}^\intercal \mathbf{V}^{-1} \mathbf{m},\label{phipost}
\end{align}
where $\mathbf{m}$ and $\mathbf{V}$ are the mean and covariance of the posterior distribution over $\uvec$, respectively. Similarly,
\begin{align}
\mathcal{G}(q_*^{\setminus n}(\uvec)) &= \frac{M}{2} \log(2\pi) + \frac{1}{2} \log |\mathbf{V}_{\mathrm{cav}, n}| + \frac{1}{2} \mathbf{m}_{\mathrm{cav}, n}^\intercal \mathbf{V}_{\mathrm{cav}, n}^{-1} \mathbf{m}_{\mathrm{cav}, n},\label{phicav}\\
\text{and}\; \mathcal{G}(p_*(\uvec)) &= \frac{M}{2} \log(2\pi) + \frac{1}{2} \log |\mathbf{K}_{\mathbf{uu}}|.
\end{align}
Finally, $\log \mathcal{Z}_{\mathrm{tilted}, n}$ is the log-normalising constant of the tilted distribution,
\begin{align}
\log \mathcal{Z}_{\mathrm{tilted}} 
&= \log \int q_\mathrm{cav}(f) p^\alpha(y_n|f) \dd f\\
&= \log \int p( f_{ \neq \uvec } | \uvec ) q_\mathrm{cav}(\uvec) p^\alpha(y_n|f) \dd f_{ \neq \uvec } \dd \uvec\\
&= \log \int p( f_n | \uvec ) q_\mathrm{cav}(\uvec) p^\alpha(y_n|f_n) \dd f_{n} \dd \uvec \label{logZtilted}
\end{align}

Next, we can write down the form of the natural parameters of the approximate posterior and the cavity distribution, based on the approximate factor's parameters, as follows,
\begin{align}
\mathbf{V}^{-1} &= \mathbf{K}_{\mathbf{uu}}^{-1} + \sum_i \mathbf{w}_i \tau_i \mathbf{w}_i^{\intercal}\label{post1}\\
\mathbf{V}^{-1} \mathbf{m} &= \sum_i \mathbf{w}_i \tau_i \tilde{y}_i\label{post2}\\
\mathbf{V}_{\mathrm{cav}, n}^{-1} &= \mathbf{V}^{-1} - \alpha \mathbf{w}_n \tau_n \mathbf{w}_n^{\intercal}\label{cav1}\\
    \mathbf{V_{\mathrm{cav}, n}}^{-1} \mathbf{m}_{\mathrm{cav}, n} &= \mathbf{V}^{-1}\mathbf{m} - \alpha \mathbf{w}_n \tau_n g_n\label{cav2}
\end{align}
Note that $\tau_i \coloneqq v_i^{-1}$. Using \cref{t3} and \cref{cav1} gives,
\begin{align}
\mathbf{V}_{\mathrm{cav}, n} = \mathbf{V} + \frac{\mathbf{V} \mathbf{w}_n \alpha \tau_n \mathbf{w}_n^\intercal \mathbf{V}}{1 - \mathbf{w}_n^\intercal \alpha \tau_n \mathbf{V} \mathbf{w}_n}. \label{cav3}
\end{align}

Using \cref{t4} and \cref{cav1} gives,
\begin{align}
\log \mathrm{det} (\mathbf{V}_{\mathrm{cav}, n}) = \log \mathrm{det} (\mathbf{V}) - \log (1 - \mathbf{w}_n^\intercal \alpha \tau_n \mathbf{V} \mathbf{w}_n). \label{cav4}
\end{align}

Subsituting \cref{cav3} and \cref{cav4} back to \cref{phicav} results in,
\begin{align}
\mathcal{G}(q_*^{\setminus n}(\uvec)) &= \frac{M}{2} \log(2\pi) + \frac{1}{2}\log \mathrm{det} (\mathbf{V})  
+ \frac{1}{2} \mathbf{m}^\intercal \mathbf{V}^{-1} \mathbf{m}\nonumber\\
&\;\; - \frac{1}{2}\log (1 - \mathbf{w}_n^\intercal \alpha \tau_n \mathbf{V} \mathbf{w}_n)
+ \frac{1}{2} \frac {\mathbf{m}^\intercal \mathbf{w}_n \alpha \tau_n \mathbf{w}_n^\intercal \mathbf{m}} {1-\mathbf{w}_n^\intercal\alpha\tau_n\mathbf{V}\mathbf{w}_n} \nonumber\\
&\;\;+ \frac {1}{2} g_n \alpha \tau_n \mathbf{w}_n^\intercal \mathbf{V}_{\mathrm{cav}, n} \mathbf{w}_n \alpha \tau_n g_n 
- g_n\alpha\tau_n \mathbf{w}_n^\intercal \mathbf{V}_{\mathrm{cav}, n} \mathbf{V}^{-1} \mathbf{m}
\end{align}
We now plug the above result back into the approximate marginal likelihood, yeilding,
\begin{align}
\mathcal{F} &= 
\frac{1}{2} \log |\mathbf{V}| + \frac{1}{2} \mathbf{m}^\intercal \mathbf{V}^{-1} \mathbf{m}
- \frac{1}{2} \log |\mathbf{K}_{\mathbf{uu}}|
+ \frac{1}{\alpha}\sum_n \log \mathcal{Z}_{\mathrm{tilted}, n} \nonumber \\
&\;\; + \sum_n \left[ - \frac{1}{2\alpha}\log (1 - \mathbf{w}_n^\intercal \alpha \tau_n \mathbf{V} \mathbf{w}_n) + \frac{1}{2} \frac {\mathbf{m}^\intercal \mathbf{w}_n \tau_n \mathbf{w}_n^\intercal \mathbf{m}} {1-\mathbf{w}_n^\intercal\alpha\tau_n\mathbf{V}\mathbf{w}_n} \right. \nonumber\\
&\;\;\left. + \frac {1}{2} g_n \tau_n \mathbf{w}_n^\intercal \mathbf{V}_{\mathrm{cav}, n} \mathbf{w}_n \alpha \tau_n g_n 
- g_n\tau_n \mathbf{w}_n^\intercal \mathbf{V}_{\mathrm{cav}, n} \mathbf{V}^{-1} \mathbf{m} \right] \label{energy}
\end{align}

\subsection{Regression}
We have shown in the previous section that the fixed point solution of the Power-EP iterations can be obtained analytically for the regression case, $g_n = y_n$ and $\tau_n^{-1} = d_n = \alpha (K_{f_nf_n} - \mathbf{K}_{f_n\uvec} \mathbf{K}_{\uvec\uvec}^{-1} \mathbf{K}_{\uvec f_n}) + \sigma_y^2$. Crucially, we can obtain a closed form expression for $\log \mathcal{Z}_{\mathrm{tilted}, n}$,
\begin{align}
\log \mathcal{Z}_{\mathrm{tilted}, n} = -\frac{\alpha}{2}\log(2\pi\sigma_y^2) + \frac{1}{2}\log(\sigma_y^2) - \frac{1}{2}\log(\alpha v_n + \sigma_y^2) - \frac{1}{2} \frac{(y_n-\mu_n)^2}{v_n + \sigma_y^2/\alpha}
\end{align}
where $\mu_n = \mathbf{w}_n^\intercal \mathbf{m}_\mathrm{cav} = \mathbf{w}_n^\intercal \mathbf{V}_\mathrm{cav} (\mathbf{V}^{-1}\mathbf{m} - \mathbf{w}_n\alpha\tau_n y_n)$ and $v_n = \frac{d_n - \sigma_y^2}{\alpha} + \mathbf{w}_n^\intercal \mathbf{V}_\mathrm{cav} \mathbf{w}_n$.
We can therefore simplify the approximate marginal likelihood $F$ further,
\begin{align}
\mathcal{F} 
&= \frac{1}{2} \log |\mathbf{V}| + \frac{1}{2} \mathbf{m}^\intercal \mathbf{V}^{-1} \mathbf{m} - \frac{1}{2} \log |\mathbf{K}_{\mathbf{uu}}| + \sum_n \left[ -\frac{1}{2}\log(2\pi\sigma_y^2) + \frac{1}{2\alpha}\log\sigma_y^2 - \frac{1}{2\alpha}\log d_n - \frac{y_n^2}{2d_n}\right]\nonumber\\
&=\textcolor{NavyBlue}{-\frac{N}{2}\log(2\pi) -\frac{1}{2} \log |\mathbf{D} + \mathbf{Q}_{\mathbf{ff}}| - \frac{1}{2} \mathbf{y}^{T} (\mathbf{D} + \mathbf{Q}_{\mathbf{ff}})^{-1} \mathbf{y} -\frac{1-\alpha}{2\alpha}\sum_n \log(\frac{d_n}{\sigma_y^2})},
\end{align}
where $\mathbf{Q}_{\mathbf{ff}} = \mathbf{K}_{\fvec\uvec} \mathbf{K}_{\uvec\uvec}^{-1} \mathbf{K}_{\uvec\fvec}$ and $\mathbf{D}$ is a diagonal matrix, $\mathbf{D}_{nn} = d_n$.

When $\textcolor{OliveGreen}{\alpha = 1}$, the approximate marginal likelihood takes the same form as the FITC marginal likelihood,
\begin{align}
\textcolor{OliveGreen}{\mathcal{F} = -\frac{N}{2}\log(2\pi) -\frac{1}{2} \log |\mathbf{D} + \mathbf{Q}_{\mathbf{ff}}| - \frac{1}{2} \mathbf{y}^{T} (\mathbf{D} + \mathbf{Q}_{\mathbf{ff}})^{-1} \mathbf{y}}
\end{align}
where $\mathbf{D}_{nn} = d_n = K_{f_nf_n} - \mathbf{K}_{f_n\uvec} \mathbf{K}_{\uvec\uvec}^{-1} \mathbf{K}_{\uvec f_n} + \sigma_y^2$.

When \textcolor{RubineRed}{$\alpha$ tends to $0$}, we have,
\begin{align}
\lim_{\alpha \to 0} \frac{1-\alpha}{2\alpha} \sum_n \log(\frac{d_n}{\sigma_y^2}) = \frac{1}{2} \sum_n \lim_{\alpha \to 0} \frac{\log(1+\alpha\frac{g_n}{\sigma_y^2})}{\alpha} = \frac{\sum_n h_n}{2\sigma_y^2},
\end{align}
where $h_n = K_{f_nf_n} - \mathbf{K}_{f_n\uvec} \mathbf{K}_{\uvec\uvec}^{-1} \mathbf{K}_{\uvec f_n}$. Therefore, 
\begin{align}
\textcolor{RubineRed}{\mathcal{F} = -\frac{N}{2}\log(2\pi) -\frac{1}{2} \log |\sigma_y^2\mathrm{I} + \mathbf{Q}_{\mathbf{ff}}| - \frac{1}{2} \mathbf{y}^{T} (\sigma_y^2\mathrm{I} + \mathbf{Q}_{\mathbf{ff}})^{-1} \mathbf{y} - \frac{\sum_n h_n}{2\sigma_y^2}},
\end{align}
which is the variational lower bound of Titsias \citep{Tit09a}.

\subsection{Classification}
In contrast to the regression case, the approximate marginal likelihood for classification cannot be simplified due to the non-Gaussian likelihood. Specifically, $\log \mathcal{Z}_{\mathrm{tilted}, n}$ is not analytically tractable, except when $\alpha=1$ and the classification link function is the Gaussian CDF. However, this quantity can be evaluated numerically, using sampling or Gauss-Hermite quadrature, since it only involves a one-dimensional integral.

We now consider the case when \textcolor{RubineRed}{$\alpha$ tends to $0$} and verify that in such case the approximate marginal likelihood becomes the variational lower bound. We first find the limits of individual terms in \cref{energy}:
\begin{align}
\lim_{\alpha \to 0} - \frac{1}{2\alpha}\log (1 - \mathbf{w}_n^\intercal \alpha \tau_n \mathbf{V} \mathbf{w}_n) &= \frac{1}{2}\mathbf{w}_n^\intercal\tau_n\mathbf{V}\mathbf{w}_n \label{cla1}\\
\frac{1}{2} \frac {\mathbf{m}^\intercal \mathbf{w}_n \tau_n \mathbf{w}_n^\intercal \mathbf{m}} {1-\mathbf{w}_n^\intercal\alpha\tau_n\mathbf{V}\mathbf{w}_n} \bigg|_{\alpha=0} &= \frac{1}{2} \mathbf{m}^\intercal \mathbf{w}_n \tau_n \mathbf{w}_n^\intercal \mathbf{m} \label{cla2}\\
\frac {1}{2} g_n \tau_n \mathbf{w}_n^\intercal \mathbf{V}_{\mathrm{cav}, n} \mathbf{w}_n \alpha \tau_n g_n \bigg|_{\alpha=0} &= 0 \label{cla3}\\
- g_n\tau_n \mathbf{w}_n^\intercal \mathbf{V}_{\mathrm{cav}, n} \mathbf{V}^{-1} \mathbf{m}  \bigg|_{\alpha=0} &= - g_n\tau_n \mathbf{w}_n^\intercal \mathbf{m}. \label{cla4}
\end{align}
We turn our attention to $\log \mathcal{Z}_{\mathrm{tilted}, n}$. First, we expand $p^\alpha(y_n|f_n)$ using \cref{expx}:
\begin{align}
p^\alpha(y_n|f_n) 
&= \exp(\alpha\log p(y_n|f_n))\\
&= 1 + \alpha\log p(y_n|f_n) + \xi(\alpha^2).
\end{align}
Substituting this result back into $\log\mathcal{Z}\mathrm{tilted}/\alpha$ gives,
\begin{align}
\frac{1}{\alpha} \log \mathcal{Z}_{\mathrm{tilted}} 
&= \frac{1}{\alpha} \log \int p( f_n | \uvec ) q_\mathrm{cav}(\uvec) p^\alpha(y_n|f_n) \dd f_{n} \dd \uvec\\
&= \frac{1}{\alpha} \log \int p( f_n | \uvec ) q_\mathrm{cav}(\uvec) [1 + \alpha\log p(y_n|f_n) + \xi(\alpha^2)] \dd f_{n} \dd \uvec\\
&= \frac{1}{\alpha} \log \left[1 + \alpha \int p( f_n | \uvec ) q_\mathrm{cav}(\uvec) \log p(y_n|f_n) \dd f_{n} \dd \uvec + \alpha^2\xi(1)\right]\\
&= \frac{1}{\alpha} \left[\alpha \int p( f_n | \uvec ) q_\mathrm{cav}(\uvec) \log p(y_n|f_n) \dd f_{n} \dd \uvec + \alpha^2\xi(1)\right]\\
&= \int p( f_n | \uvec ) q_\mathrm{cav}(\uvec) \log p(y_n|f_n) \dd f_{n} \dd \uvec + \alpha\xi(1).
\end{align}
Therefore,
\begin{align}
\lim_{\alpha \to 0} \frac{1}{\alpha} \log \mathcal{Z}_{\mathrm{tilted}} = \int p( f_n | \uvec ) q(\uvec) \log p(y_n|f_n) \dd f_{n} \dd \uvec. \label{cla5}
\end{align}
Putting these results into \cref{energy}, we obtain,
\begin{align}
\mathcal{F} &= 
\frac{1}{2} \log |\mathbf{V}| + \frac{1}{2} \mathbf{m}^\intercal \mathbf{V}^{-1} \mathbf{m}
- \frac{1}{2} \log |\mathbf{K}_{\mathbf{uu}}|\nonumber\\
&\;\;\; + \sum_n \frac{1}{2}\mathbf{w}_n^\intercal\tau_n\mathbf{V}\mathbf{w}_n + \frac{1}{2} \mathbf{m}^\intercal \mathbf{w}_n \tau_n \mathbf{w}_n^\intercal \mathbf{m} - g_n\tau_n \mathbf{w}_n^\intercal \mathbf{m} + \int p( f_n | \uvec ) q(\uvec) \log p(y_n|f_n) \dd f_{n} \dd \uvec\nonumber\\
&=
\frac{1}{2} \log |\mathbf{V}| + \frac{1}{2} \mathbf{m}^\intercal \mathbf{V}^{-1} \mathbf{m}
- \frac{1}{2} \log |\mathbf{K}_{\mathbf{uu}}| + \frac{1}{2} \mathbf{m}^\intercal (\mathbf{V}^{-1} - \mathbf{K}_\mathbf{uu}^{-1}) \mathbf{m} - \mathbf{m}^\intercal\mathbf{V}^{-1} \mathbf{m}\nonumber\\
&\;\;\;\; + \sum_n \frac{1}{2}\mathbf{w}_n^\intercal\tau_n\mathbf{V}\mathbf{w}_n  + \int p( f_n | \uvec ) q(\uvec) \log p(y_n|f_n) \dd f_{n} \dd \uvec\nonumber\\
&=\textcolor{RubineRed}{\frac{1}{2} \log |\mathbf{V}| - \frac{1}{2} \mathbf{m}^\intercal \mathbf{K}_\mathbf{uu}^{-1} \mathbf{m}
- \frac{1}{2} \log |\mathbf{K}_{\mathbf{uu}}| + \sum_n \frac{1}{2}\mathbf{w}_n^\intercal\tau_n\mathbf{V}\mathbf{w}_n + \sum_n \int p( f_n | \uvec ) q(\uvec) \log p(y_n|f_n) \dd f_{n} \dd \uvec}. \label{claenergy}
\end{align}

We now write down the evidence lower bound of the global variational approach of Titsias \citep{Tit09a}, as applied to the classification case \citep{HenMatGha15},
\begin{align}
\mathcal{F}_\mathrm{VFE} 
&= -\mathrm{KL}(q(\uvec) || p(\uvec)) + \sum_n \int p( f_n | \uvec ) q(\uvec) \log p(y_n|f_n) \dd f_{n} \dd \uvec \end{align}
where
\begin{align}
-\mathrm{KL}(q(\uvec) || p(\uvec)) 
&= - \frac{1}{2} \mathrm{trace}(\mathbf{K}_\mathbf{uu}^{-1}\mathbf{V}) 
	- \frac{1}{2} \mathbf{m}^\intercal \mathbf{K}_\mathbf{uu}^{-1} \mathbf{m} 
	+ \frac{M}{2} - \frac{1}{2}\log|\mathbf{K}_\mathbf{uu}| + \frac{1}{2} \log|\mathbf{V}| \nonumber \\
&= - \frac{1}{2} \mathrm{trace}([\mathbf{V}^{-1} - \sum_n \mathbf{w}_n \tau_n \mathbf{w}_n] \mathbf{V}) 
	- \frac{1}{2} \mathbf{m}^\intercal \mathbf{K}_\mathbf{uu}^{-1} \mathbf{m} 
	+ \frac{M}{2} - \frac{1}{2}\log|\mathbf{K}_\mathbf{uu}| + \frac{1}{2} \log|\mathbf{V}| \nonumber \\
&= \frac{1}{2} \mathrm{trace}(\sum_n \mathbf{w}_n \tau_n \mathbf{w}_n\mathbf{V}) 
	- \frac{1}{2} \mathbf{m}^\intercal \mathbf{K}_\mathbf{uu}^{-1} \mathbf{m} 
	- \frac{1}{2}\log|\mathbf{K}_\mathbf{uu}| + \frac{1}{2} \log|\mathbf{V}|.
\end{align}
Therefore, $\mathcal{F}_\mathrm{VFE}$ is \textcolor{RubineRed}{identical} to the limit of the approximate marginal likelihood provided by power-EP as shown in \cref{claenergy}.

\section{The surrogate regression viewpoint\label{app:surrogate}}

It was written in the main text that it is instructive to view the approximation using pseudo-points as forming a surrogate exact Gaussian process regression problem such that the posterior and the marginal likelihood of this surrogate problem are close to that of the original intractable regression/classification problem. This approximation view is useful and could potentially be used for other intractable probabilistic model, despite that we have not used this view in the practical implementation of the algorithms/PEP procedure discussed in this paper. In this section, we detail the surrogate model and how the parameters of this model can be tuned to match the approximate posterior and approximate marginal likelihood. 

We consider the exact GP regression problem with $M$ surrogate observations $\tilde{\yvec}$ that are formed by linear combininations the pseudo-outputs and additive surrogate Gaussian noise, $\tilde{\yvec} = \tilde{\mathbf{W}}\uvec + \tilde{\Sigma}^{1/2}\epsilon$. The exact posterior and log marginal likelihood can be obtained for this model as follows,
\begin{align}
\tilde{p}(\uvec|\yvec) &= \norm^{-1}(\uvec; \tilde{\mathbf{W}} \tilde{\Sigma}^{-1}\tilde{\yvec}, \kuu^{-1} + \tilde{\mathbf{W}}^{\intercal} \tilde{\Sigma}^{-1} \tilde{\mathbf{W}}) \\
\log p(\tilde{\yvec}) &= - \frac{M}{2}\log(2\pi) - \frac{1}{2} (\log | \kuu^{-1} + \tilde{\mathbf{W}}^{\intercal} \tilde{\Sigma}^{-1} \tilde{\mathbf{W}}| + \log |\kuu^{-1}| + \log |\tilde{\Sigma}|) \nonumber \\ & \quad - \frac{1}{2} \tilde{\yvec}^\intercal \tilde{\Sigma}^{-1} \tilde{\yvec} - \frac{1}{2} \tilde{\yvec}^\intercal \tilde{\Sigma}^{-1} \tilde{\mathbf{W}} (\kuu^{-1} + \tilde{\mathbf{W}}^{\intercal} \tilde{\Sigma}^{-1} \tilde{\mathbf{W}})^{-1} \tilde{\mathbf{W}}^\intercal \tilde{\Sigma}^{-1} \tilde{\yvec},
\end{align}
where we have used the matrix inversion lemma and the matrix determinant lemma in the equations above, and that $\norm^{-1}$ denotes the Gaussian distribution with natural parameters. 

The aim is to show that we can use the above quantities is to match a given approximate posterior $q(\uvec) = \norm^{-1}(\uvec; \Svec^{-1}\mvec, \Svec^{-1})$ and an approximate marginal likelihood $\mathcal{F}$, that is, $\tilde{p}(\uvec|\yvec)= q(\uvec)$ and $\log p(\tilde{\yvec}) = \mathcal{F}$. Substituting the above results into the constraints leading to the following simplified constraints:
\begin{align}
\tilde{\mathbf{W}} \tilde{\Sigma}^{-1}\tilde{\yvec} &= \mvec\\
\tilde{\mathbf{W}}^{\intercal} \tilde{\Sigma}^{-1} \tilde{\mathbf{W}} &= \Rvec = \kuu^{-1} - \Svec^{-1} \\
\tilde{\yvec}^\intercal \tilde{\Sigma}^{-1} \tilde{\yvec} + \log |\tilde{\Sigma}| &= c,
\end{align}
where $c$ is a constant. Assume that $\Rvec$ is invertible, we can simplified the above results further,
\begin{align}
\tilde{\Sigma}^{-1/2}\tilde{\yvec} &= \Rvec^{-1/2} \mvec\\
\tilde{\Sigma}^{-1/2} \tilde{\mathbf{W}}  &= \Rvec^{\intercal/2} \\
\log |\tilde{\Sigma}| &= d,
\end{align}
where $d$ is a constant. We can choose $\tilde{\Sigma}$, e.g.~a diagonal matrix, that satisfies the third equality above. Given $\tilde{\Sigma}$, obtaining $\tilde{\yvec}$ and $\tilde{\mathbf{W}}$ from the first two equalities is trivial.

\section{Extra experimental results\label{app:extra_exp}}

\subsection{Comparison between various $\alpha$ values on a toy regression problem}

\begin{landscape}
\begin{figure}[!ht]
\centering
\includegraphics[width=1\linewidth]{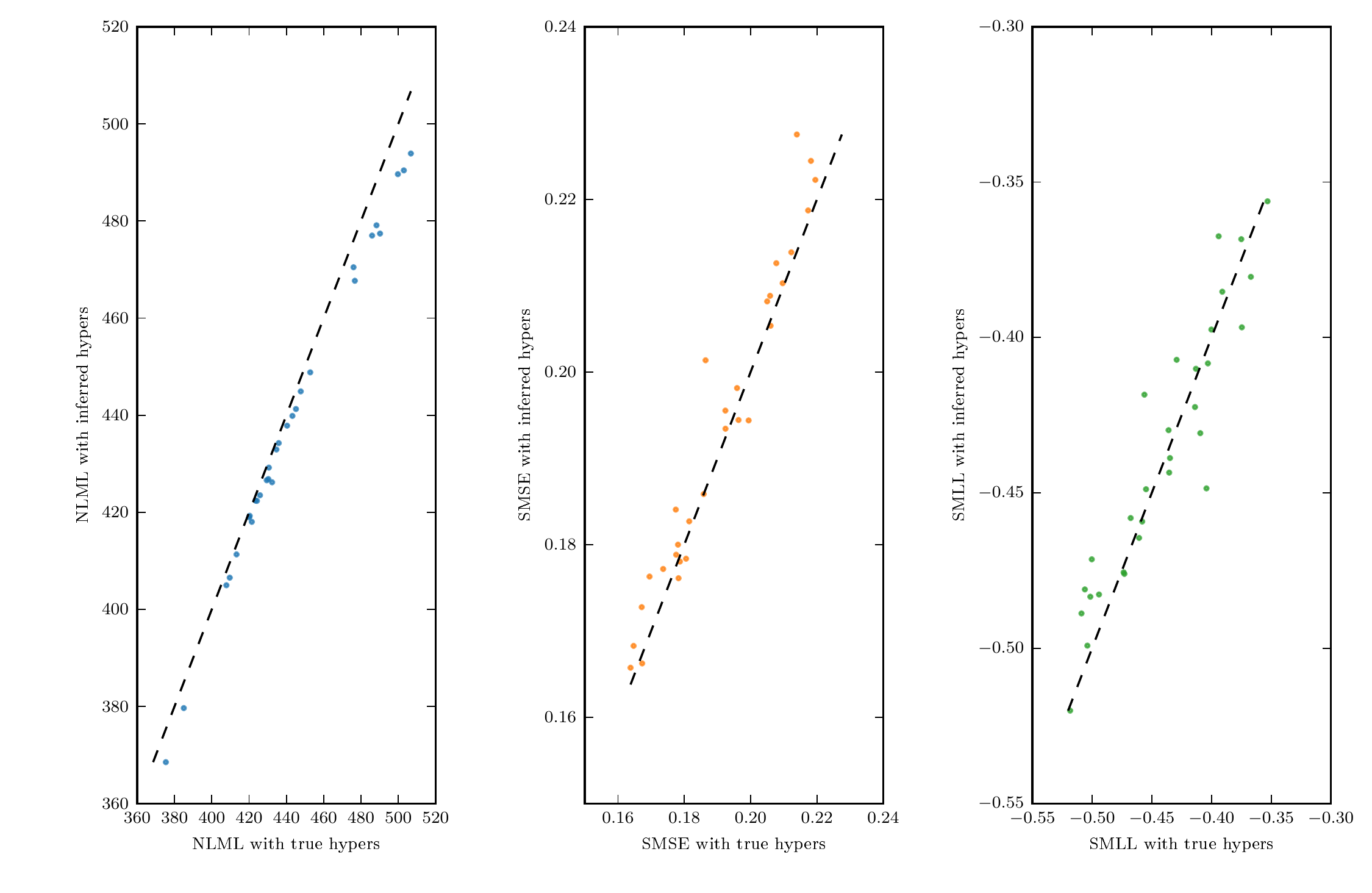}
\caption{Results on a toy regression problem: Negative log-marginal likelihood, mean squared error and mean log-loss on the test set for full Gaussian process regression on synthetic datasets with {\it true} hyper-parameters and hyper-parameters obtained by type-2 ML. Each dot is one trial, i.e.~one synthetic dataset. The results demonstrate that type-2 maximum likelihood on hyper-parameters works well, despite being a little confident on the log-marginal likelihood on the train set.}
\end{figure}

\begin{figure}[!ht]
\centering
\includegraphics[width=1\linewidth]{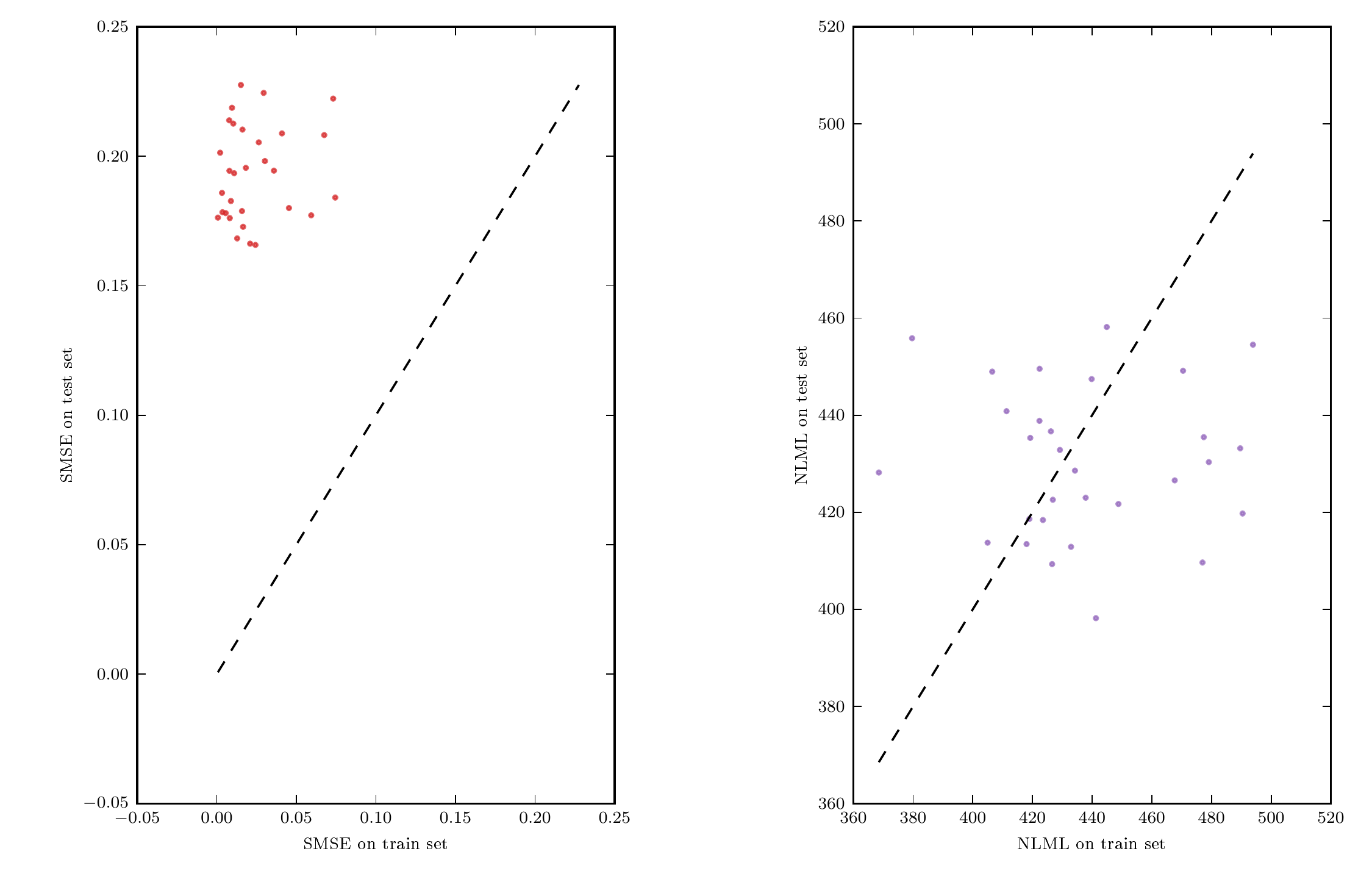}
\caption{Results on a toy regression problem with 500 training points: Mean squared error and log-likelihood on train and test sets on synthetic datasets with hyper-parameters obtained by type-2 ML. In this example, the test error is higher than the training error, as measured by the mean squared error, because the test points and training points are relatively far apart, making the prediction task on the training set easier (interpolation) than on the test set (extrapolation). This is consistent with the results with more training points, shown in \cref{fig:toy_1000}.\label{fig:toy_500}}
\end{figure}

\begin{figure}[!ht]
\centering
\includegraphics[width=1\linewidth]{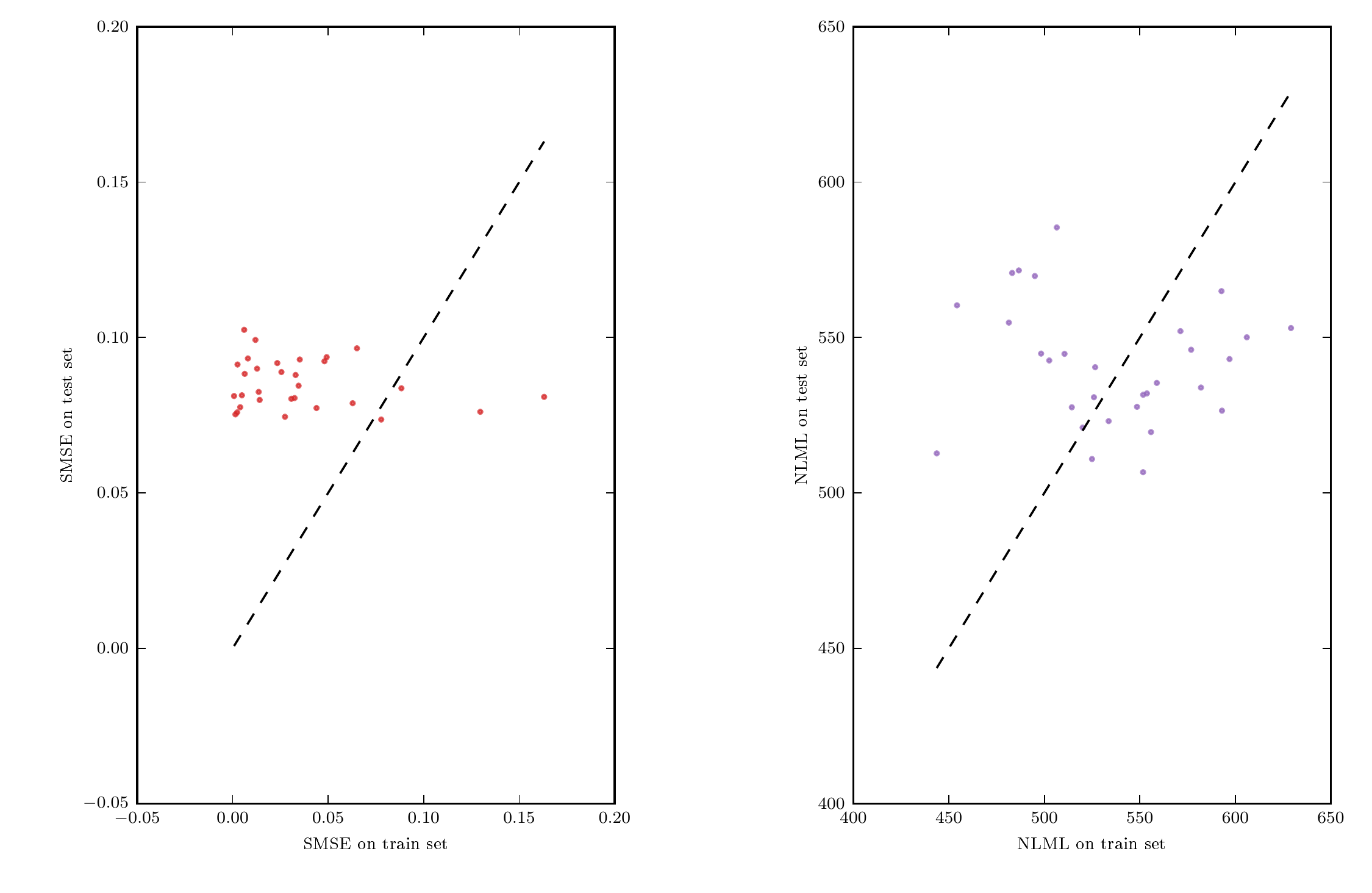}
\caption{Results on a toy regression problem with 1000 training points: Mean squared error and log-likelihood on train and test sets on synthetic datasets with hyper-parameters obtained by type-2 ML. See \cref{fig:toy_500} for a discussion.\label{fig:toy_1000}}
\end{figure}

\begin{figure}[!ht]
\centering
\includegraphics[trim={2cm 2cm 2cm 2cm},clip,width=1\linewidth]{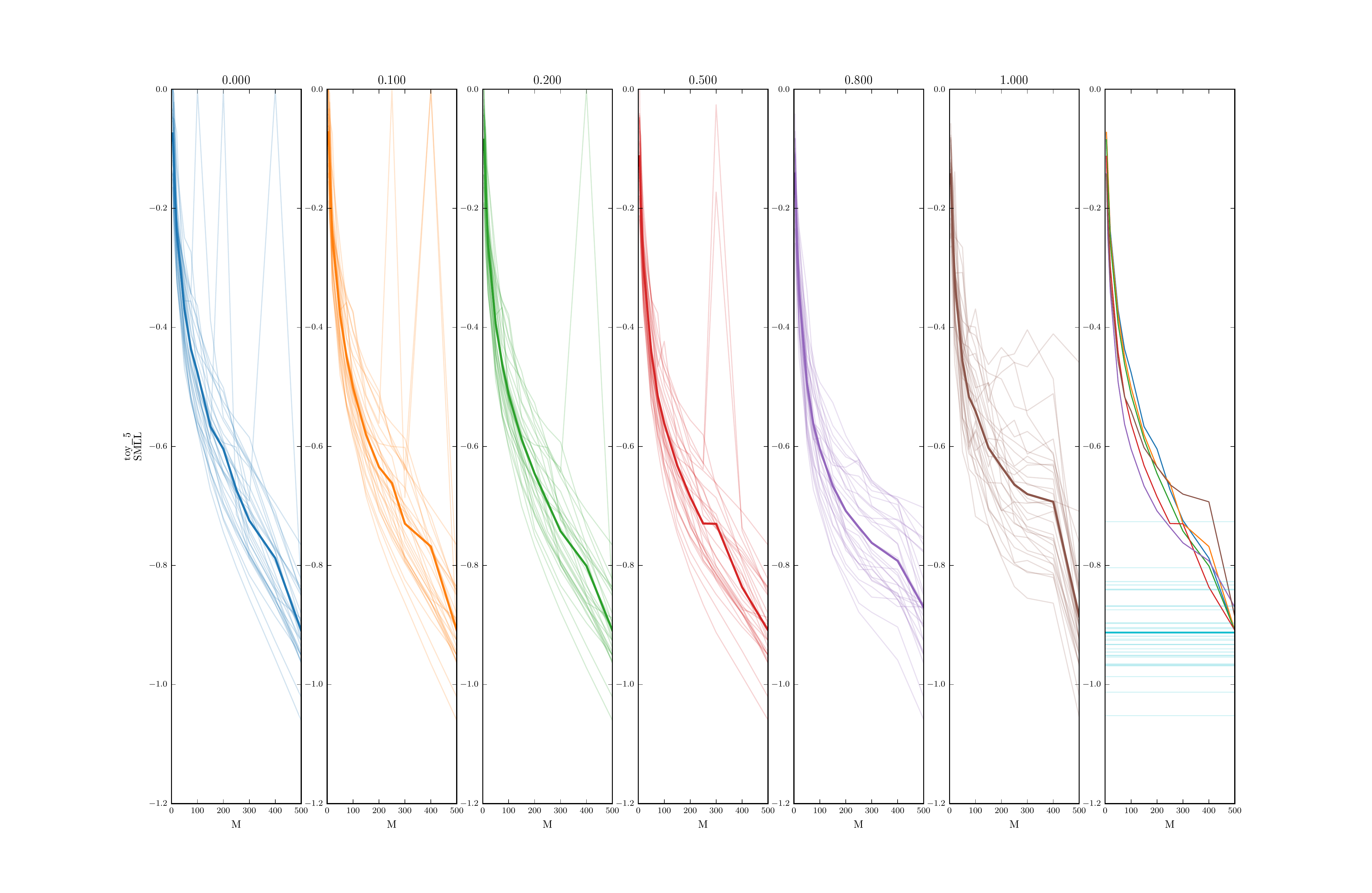}
\caption{Results on a toy regression problem: Standardsised mean log-loss on the test set for various values of $\alpha$ and various number of pseudo-points $M$. Each trace is for one split, bold line is the mean. The rightmost figure shows the mean for various $\alpha$, and the results using GP regression.}
\end{figure}

\begin{figure}[!ht]
\centering
\includegraphics[trim={2cm 2cm 2cm 2cm},clip,width=1\linewidth]{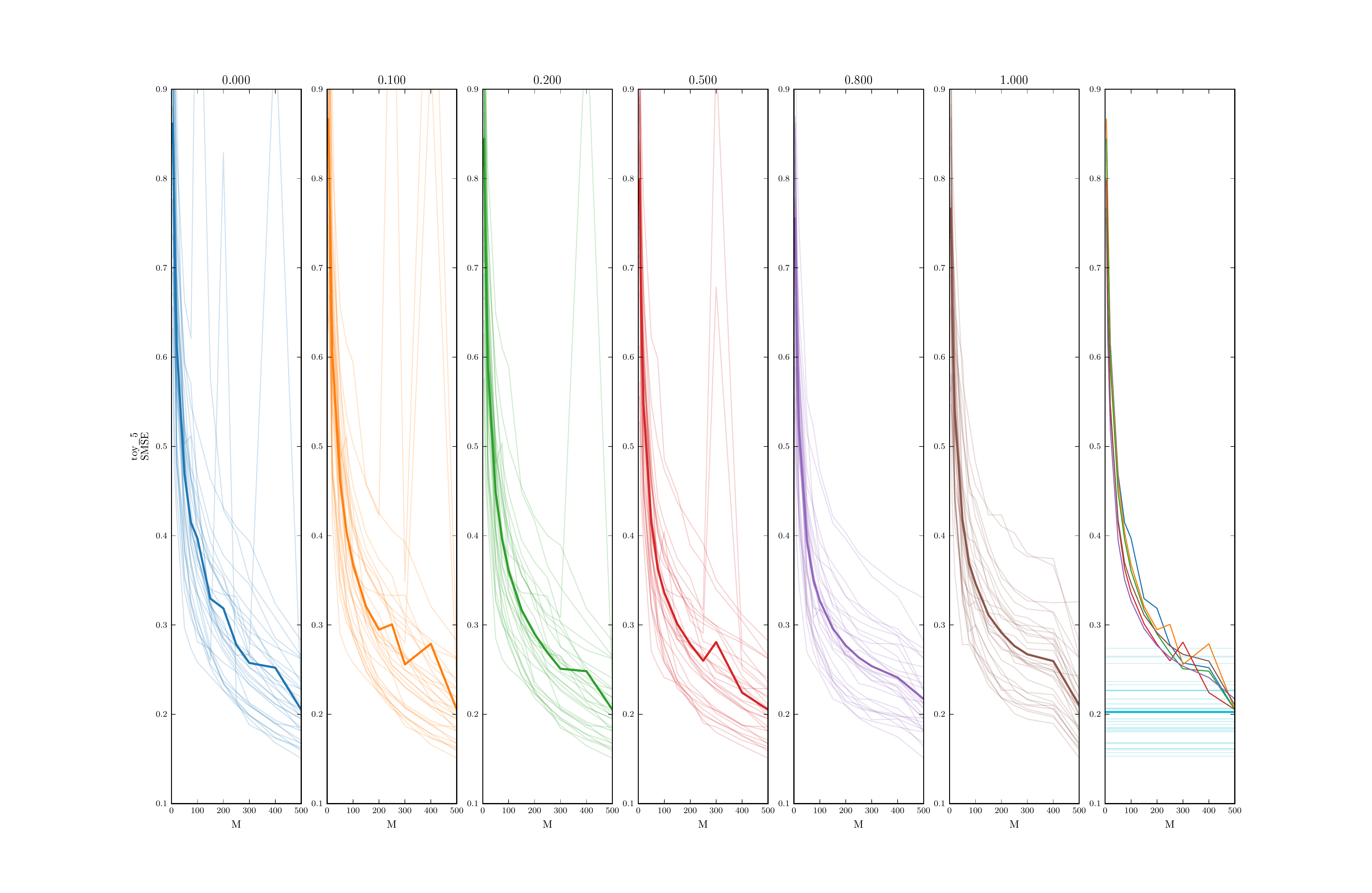}
\caption{Results on a toy regression problem: Standardsised mean squared error on the test set for various values of $\alpha$ and various number of pseudo-points $M$. Each trace is for one split, bold line is the mean. The rightmost figure shows the mean for various $\alpha$, and the results using GP regression.}
\end{figure}

\begin{figure}[!ht]
\centering
\includegraphics[trim={2cm 2cm 2cm 2cm},clip,width=1\linewidth]{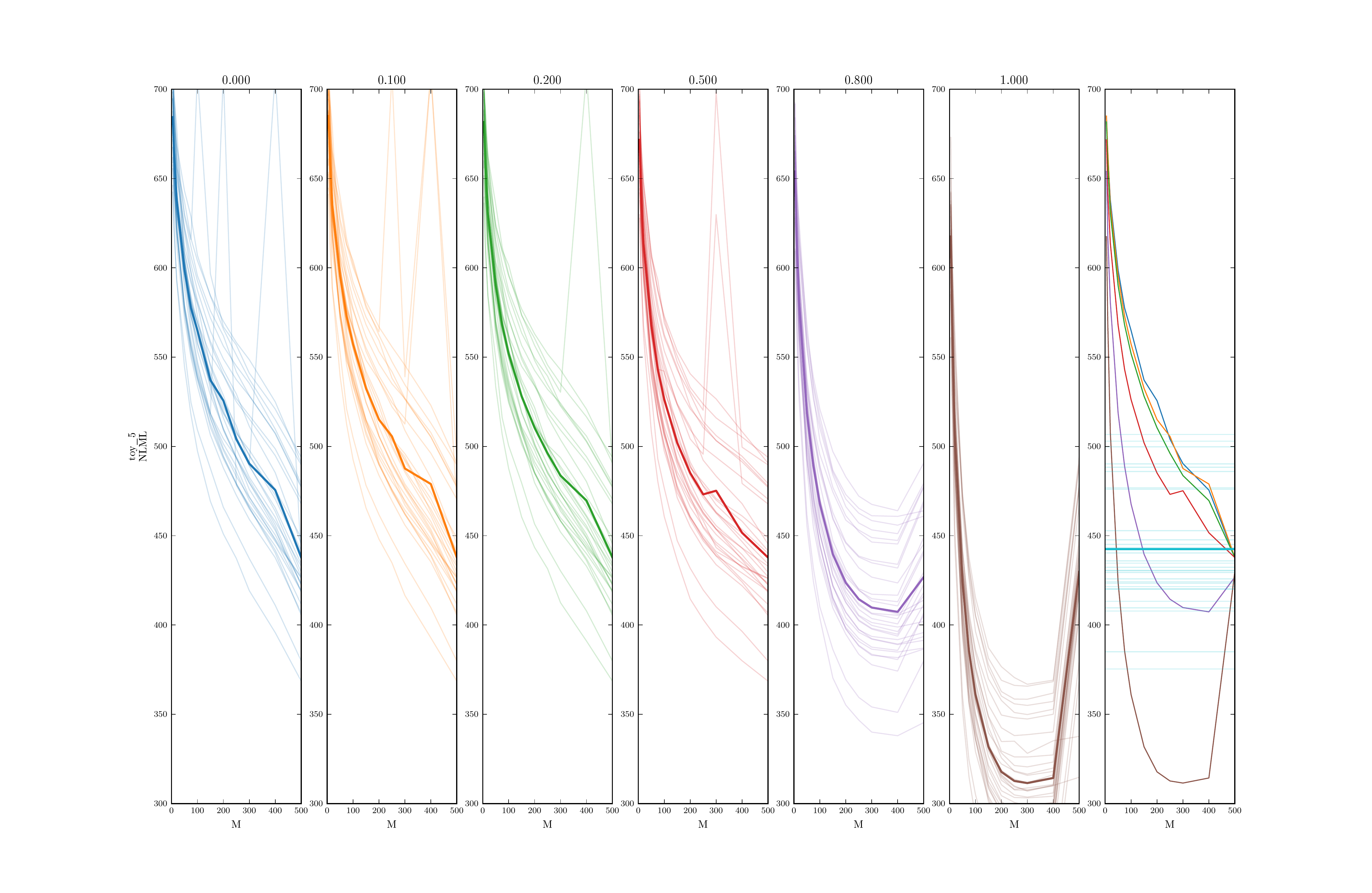}
\caption{Results on a toy regression problem: The negative log marginal likelihood of the training set after training for various values of $\alpha$ and various number of pseudo-points $M$. Each trace is for one split, bold line is the mean. The rightmost figure shows the mean for various $\alpha$, and the results using GP regression. Power EP with $\alpha$ close to 1 over-estimates the marginal-likelihood.}
\end{figure}

\end{landscape}

\subsection{Real-world regression}
We include the details of the regression datasets in \cref{tab:regdatasets} and several comparisons of $\alpha$ values in \cref{fig:reg_nlml_table,fig:reg_nlml_mean_var,fig:reg_smse_table,fig:reg_smse_mean_var,fig:reg_smll_table,fig:reg_smll_mean_var}.

\begin{table}[!ht]
\centering
 \begin{tabular}{||c c c||} 
 \hline
 Dataset & N train/test & D \\ [0.5ex] 
 \hline\hline
	boston & 455/51 & 14 \\
	concrete & 927/103 & 9 \\
	energy & 691/77 & 9 \\
	kin8nm & 7373/819 & 9 \\
	naval & 10741/1193 & 18 \\
	yacht & 277/31 & 7 \\
	power & 8611/957 & 5 \\
	red wine & 1439/160 & 12 \\[1ex] 
 \hline
\end{tabular}
\caption{Regression datasets}
\label{tab:regdatasets}
\end{table}

\begin{landscape}

\begin{figure}[!ht]
    \centering
    \includegraphics[trim={1cm 1cm 1cm 1cm},clip,width=\linewidth]{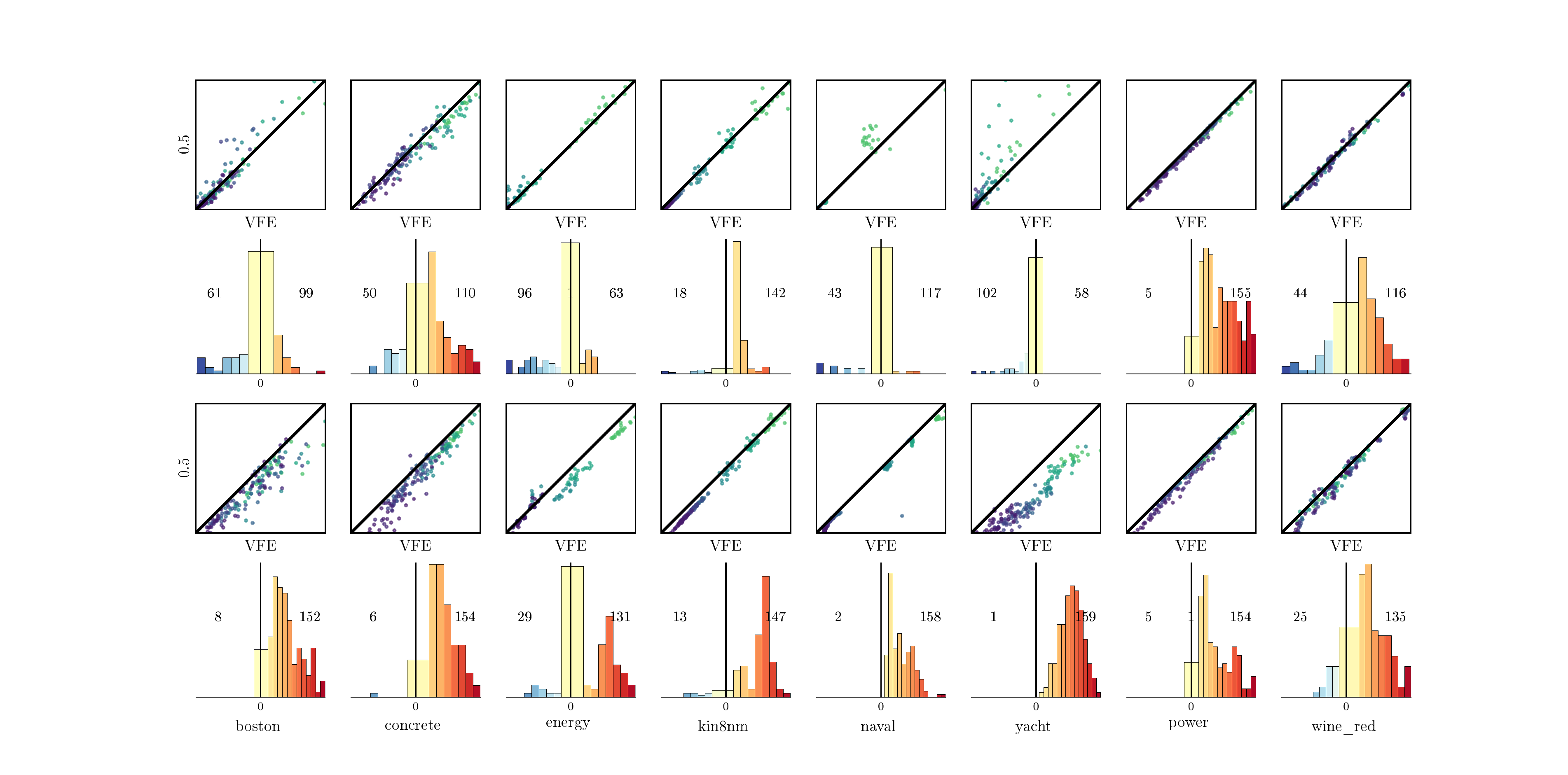}
	\caption{A comparison between Power-EP with $\alpha=0.5$ and VFE on several regression datasets, on two metrics SMSE (top two rows) and SMLL (bottom two rows). The scatter plots show the performance of Power-EP ($\alpha=0.5$) vs VFE. Each point is one split and points with lighter colours are runs with big M. Points that stay below the diagonal line show $\alpha=0.5$ is better than VFE. The plots right underneat the scatter plots show the histogram of the difference between methods. Red means $\alpha=0.5$ is better than VFE. \label{fig:reg_nll_subset0}}
\end{figure}

\begin{figure}[!ht]
    \centering
    \includegraphics[trim={1cm 1cm 1cm 1cm},clip,width=\linewidth]{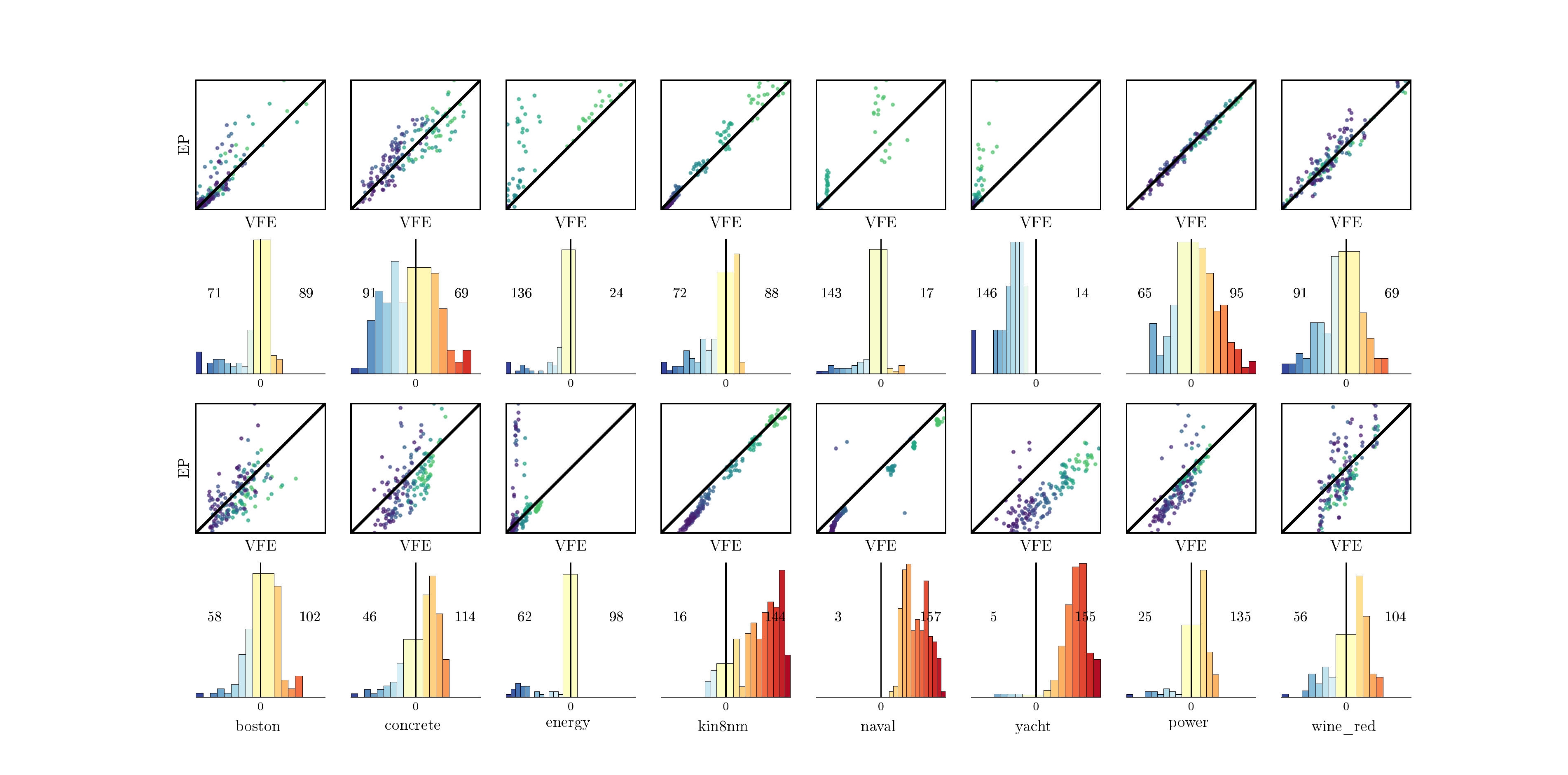}
	\caption{A comparison between EP and VFE on several regression datasets, on two metrics SMSE (top two rows) and SMLL (bottom two rows). See \cref{fig:reg_nll_subset0} for more details about the plots.\label{fig:reg_nll_subset1}}
\end{figure}

\begin{figure}[!ht]
    \centering
    \includegraphics[trim={1cm 1cm 1cm 1cm},clip,width=\linewidth]{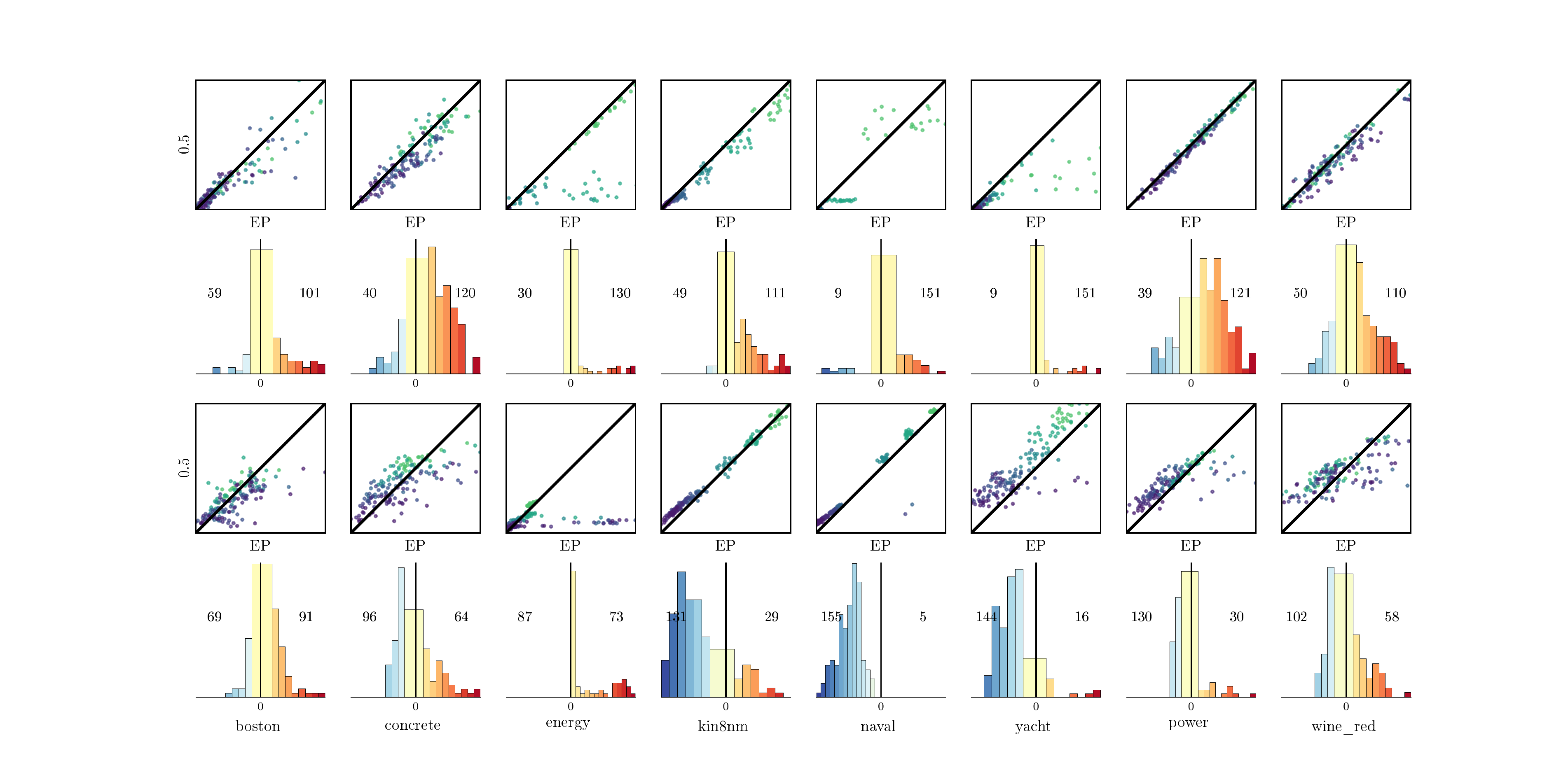}
	\caption{A comparison between Power-EP with $\alpha=0.5$ and EP on several regression datasets, on two metrics SMSE (top two rows) and SMLL (bottom two rows). See \cref{fig:reg_nll_subset0} for more details about the plots.\label{fig:reg_nll_subset2}}
\end{figure}

\begin{figure}[!ht]
\centering
\includegraphics[trim={2cm 2cm 2cm 2cm},clip,width=1.6\textwidth]{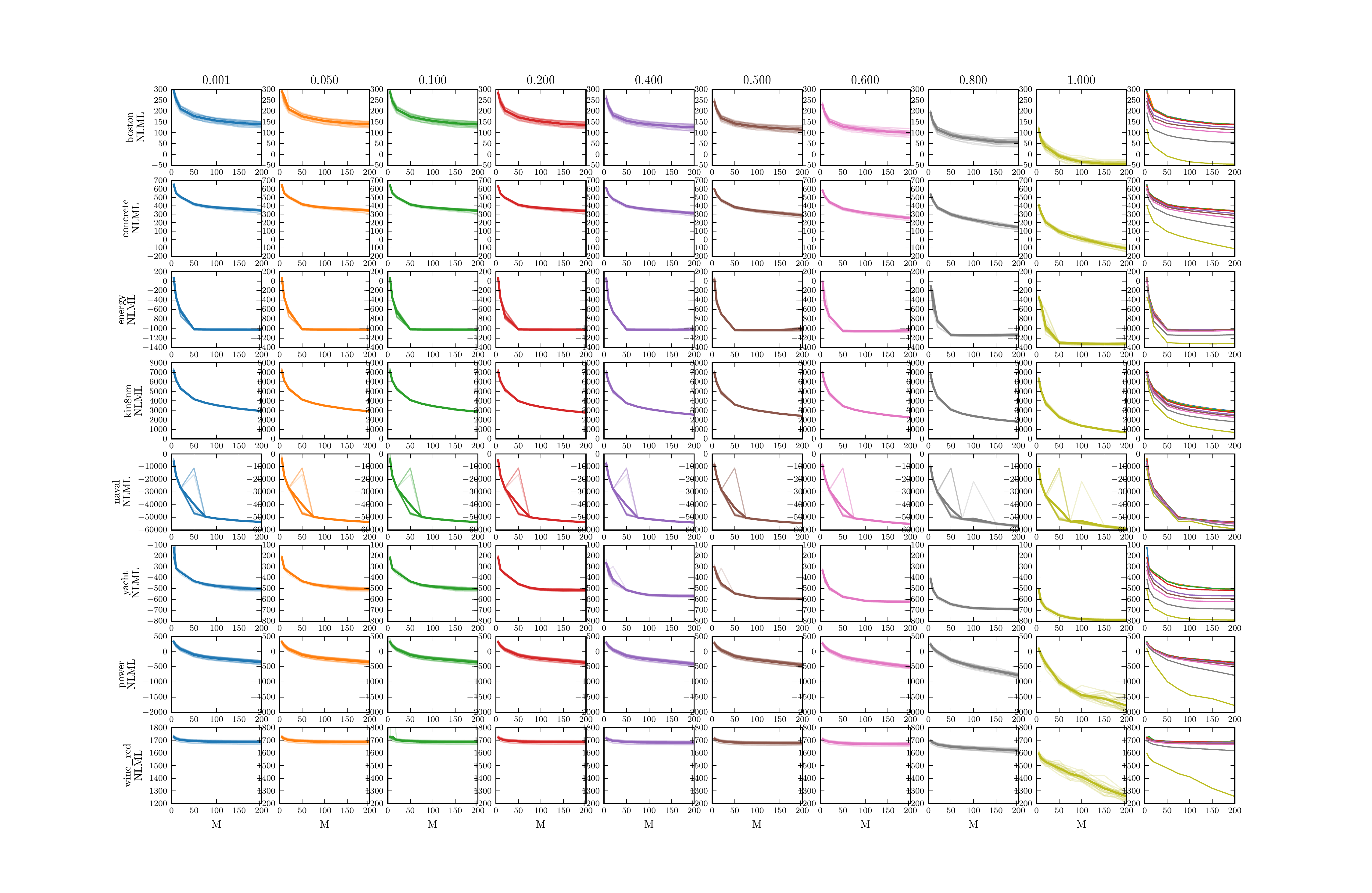}
\caption{Results on real-world regression problems: Negative training log-marginal likelihood for different datasets, various values of $\alpha$ and various number of pseudo-points $M$. Each trace is for one split, bold line is the mean. The rightmost figures show the mean for various $\alpha$ for comparison. Lower is better [however, lower could mean overestimation].\label{fig:reg_nlml_table}}
\end{figure}

\begin{figure}[!ht]
\centering
\includegraphics[trim={2cm 2cm 2cm 2cm},clip,width=1.6\textwidth]{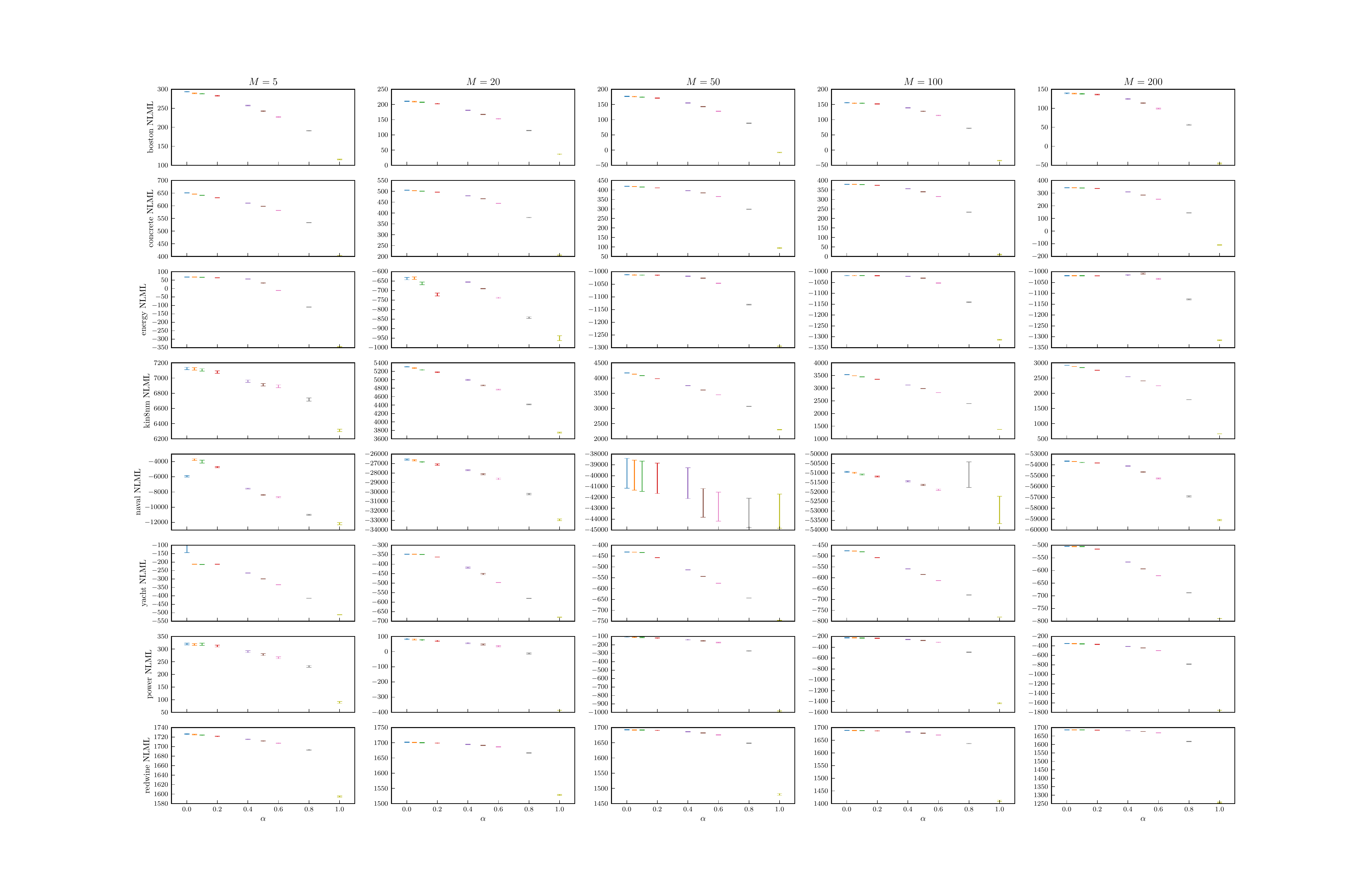}
\caption{Results on real-world regression problems: Negative training log-marginal likelihood for different datasets, various values of $\alpha$ and various number of pseudo-points $M$, averaged over 20 splits. Lower is better [however, lower could mean overestimation].\label{fig:reg_nlml_mean_var}}
\end{figure}

\begin{figure}[!ht]
\centering
\includegraphics[trim={2cm 2cm 2cm 2cm},clip,width=1.6\textwidth]{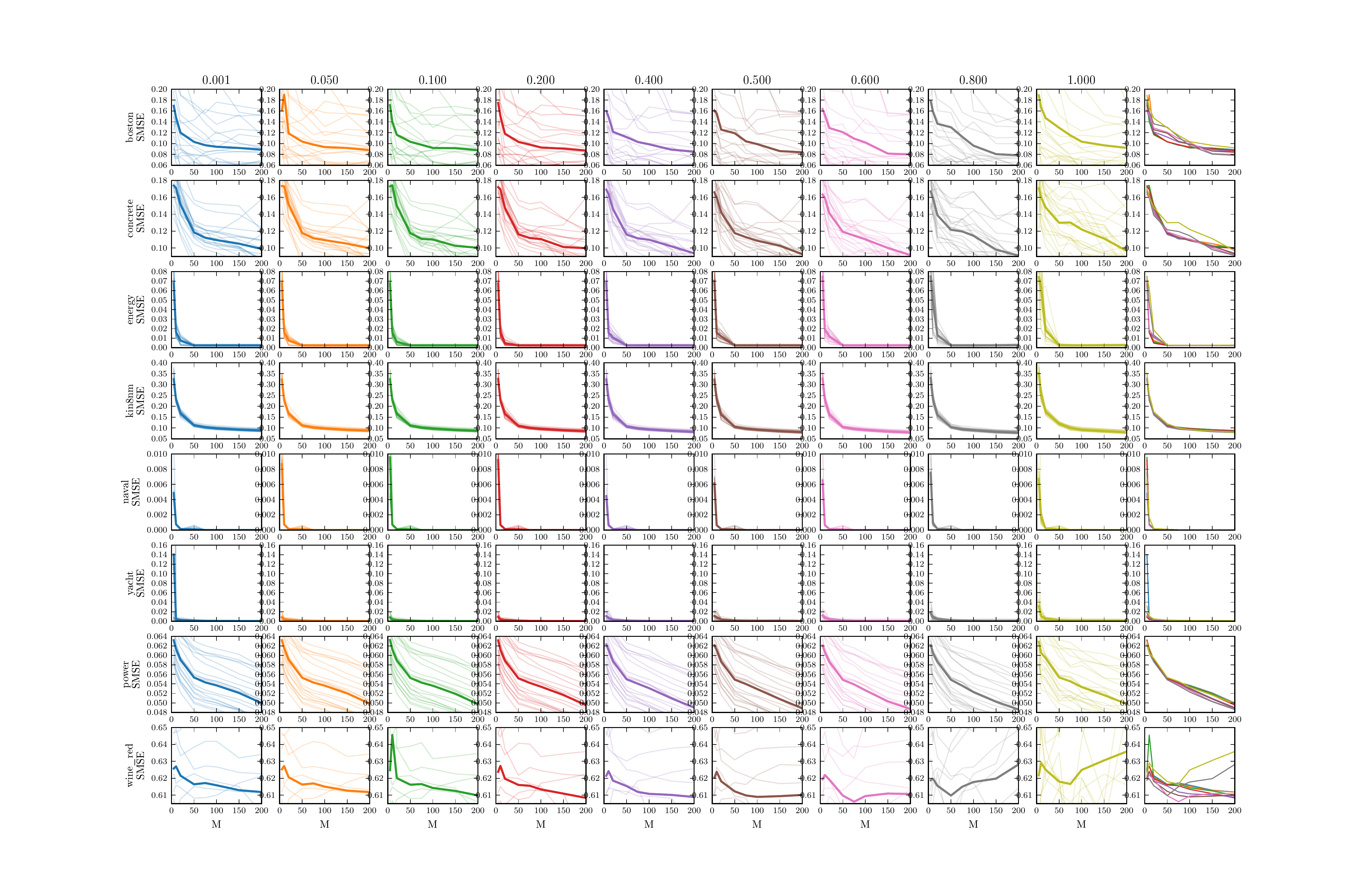}
\caption{Results on real-world regression problems: Standardised mean squared error on the test set for different datasets, various values of $\alpha$ and various number of pseudo-points $M$. Each trace is for one split, bold line is the mean. The rightmost figures show the mean for various $\alpha$ for comparison. Lower is better.\label{fig:reg_smse_table}}
\end{figure}

\begin{figure}[!ht]
\centering
\includegraphics[trim={2cm 2cm 2cm 2cm},clip,width=1.6\textwidth]{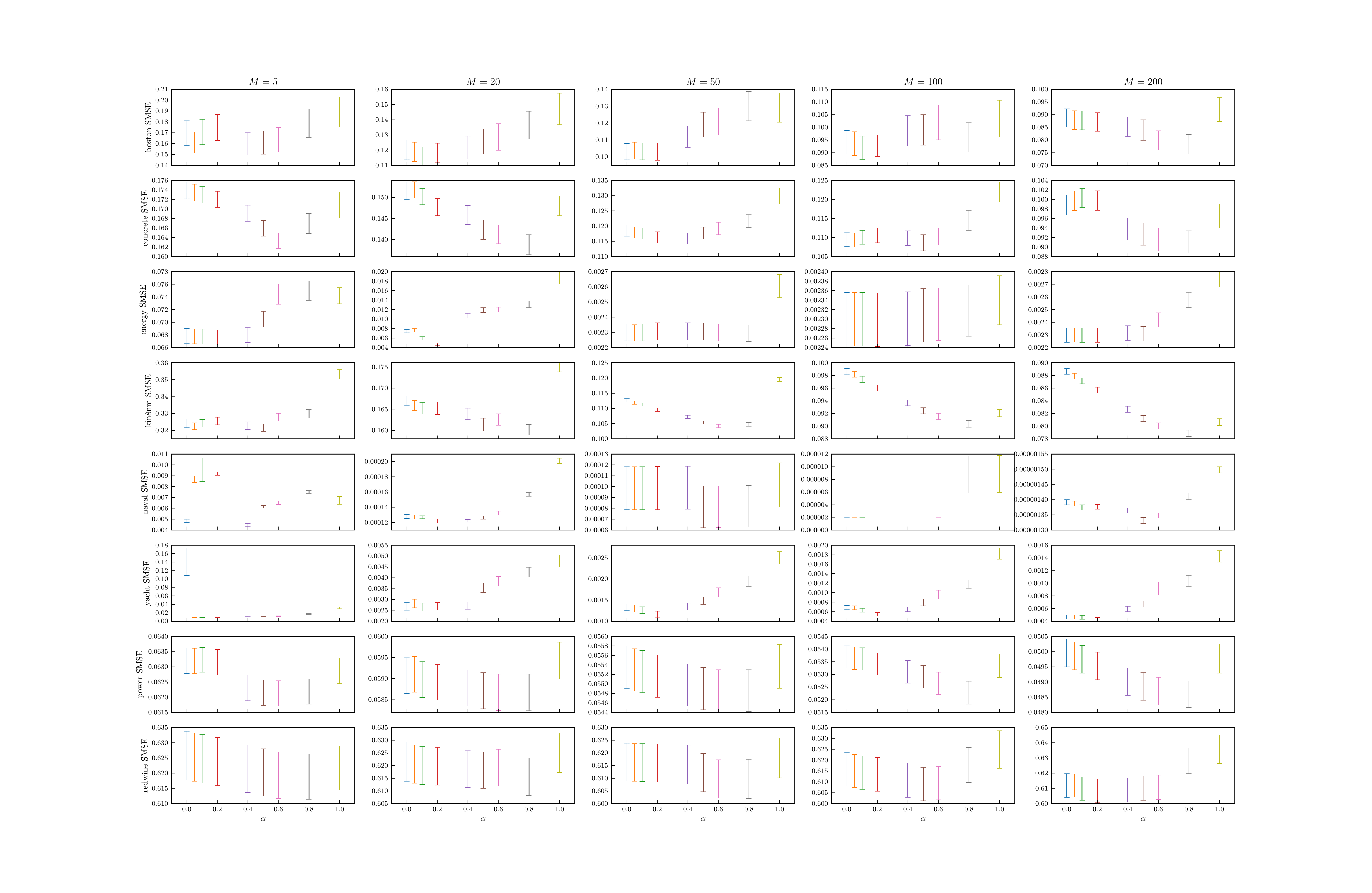}
\caption{Results on real-world regression problems: Standardised mean squared error on the test set for different datasets, various values of $\alpha$ and various number of pseudo-points $M$, averaged over 20 splits. Lower is better.\label{fig:reg_smse_mean_var}}
\end{figure}

\begin{figure}[!ht]
\centering
\includegraphics[trim={2cm 2cm 2cm 2cm},clip,width=1.6\textwidth]{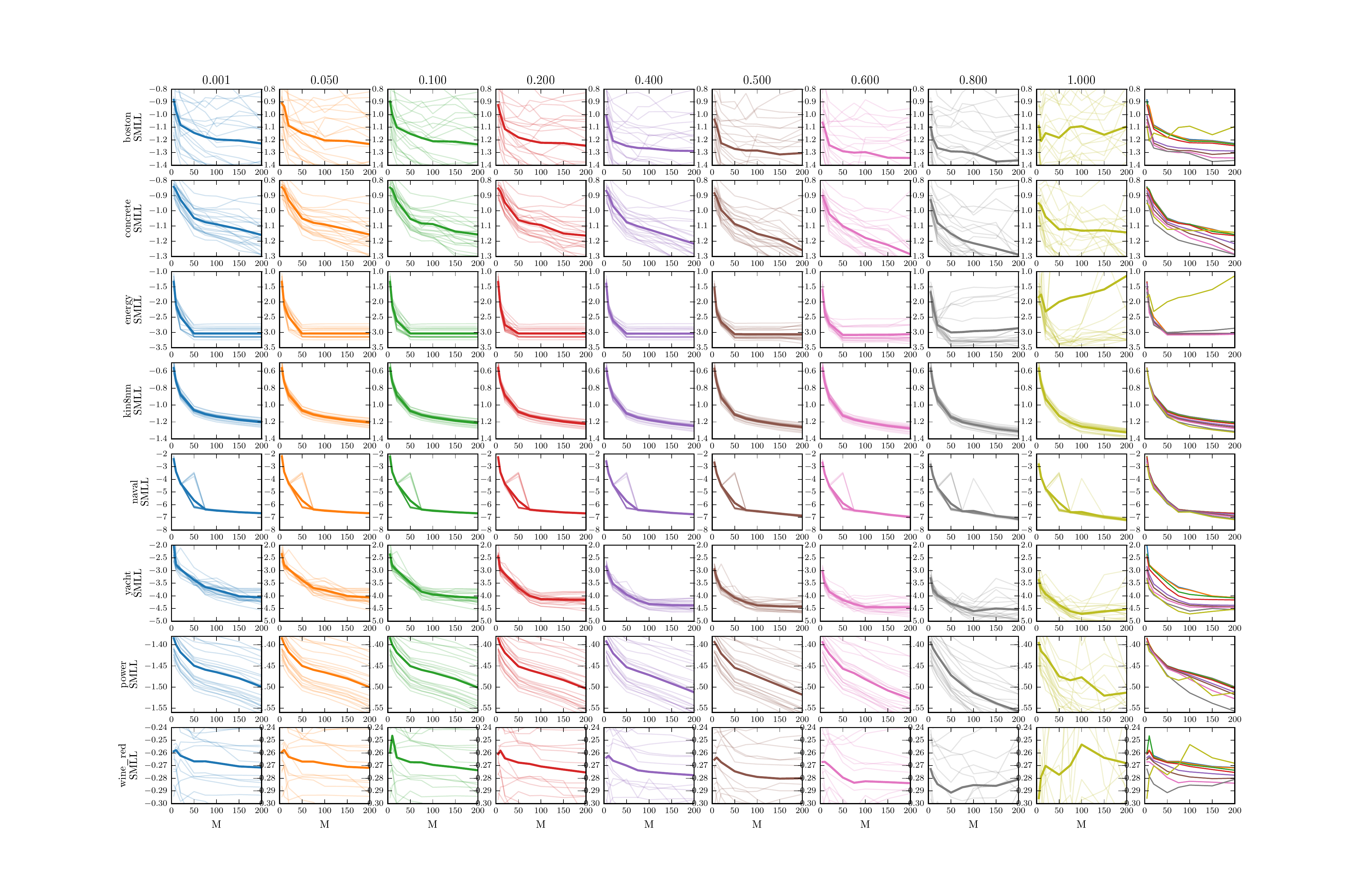}
\caption{Results on real-world regression problems: Standardised mean log loss on the test set for different datasets, various values of $\alpha$ and various number of pseudo-points $M$. Each trace is for one split, bold line is the mean. The rightmost figures show the mean for various $\alpha$ for comparison. Lower is better.\label{fig:reg_smll_table}}
\end{figure}

\begin{figure}[!ht]
\centering
\includegraphics[trim={2cm 2cm 2cm 2cm},clip,width=1.6\textwidth]{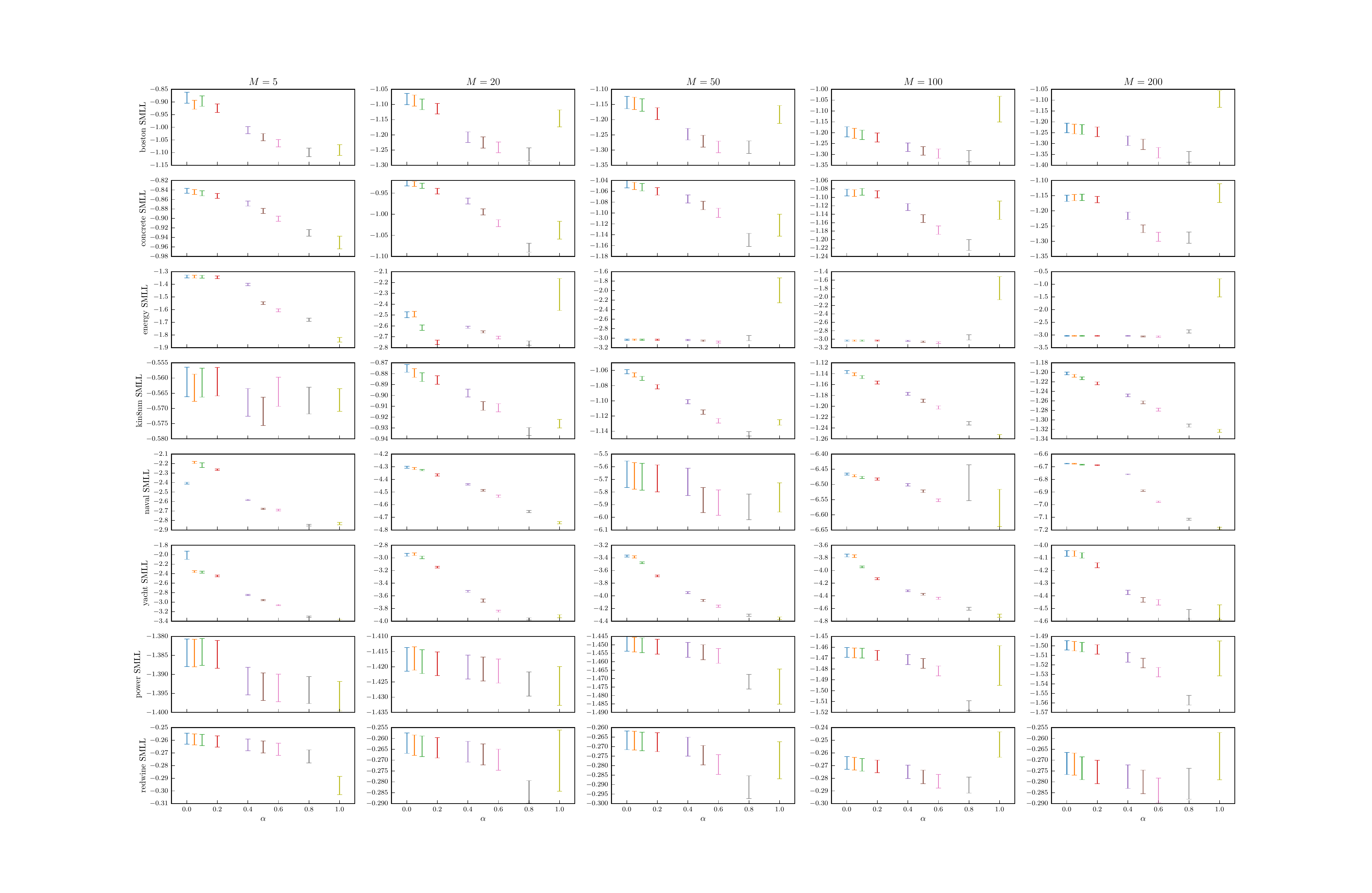}
\caption{Results on real-world regression problems: Standardised mean log loss on the test set for different datasets, various values of $\alpha$ and various number of pseudo-points $M$, averaged over 20 splits. Lower is better.\label{fig:reg_smll_mean_var}}
\end{figure}

\end{landscape}

\subsection{Real-world classification}
\label{app:classification}
\begin{figure}[!ht]
\centering
\includegraphics[trim={2cm 2cm 2cm 2cm},clip,width=1\textwidth]{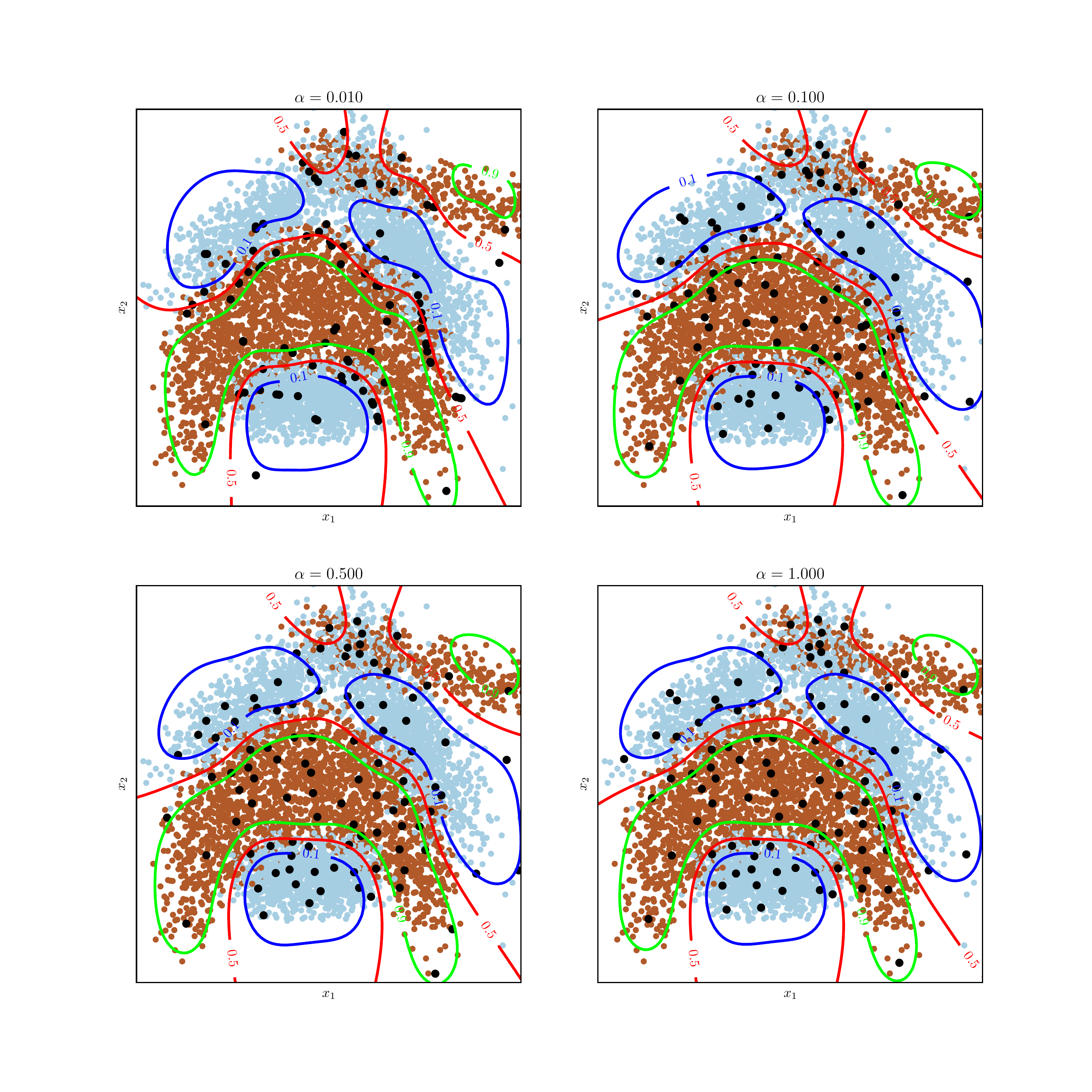}
\caption{The locations of pseudo data points vary with $\alpha$.\label{fig:banana}. Best viewed in colour.}
\end{figure}

It was demonstrated in \citep{HerHer16, HenMatGha15} that, once optimised, the pseudo points tend to concentrate around the decision boundary for VFE, and spread out to cover the data region in EP. \Cref{fig:banana} illustrates the same effect as $\alpha$ goes from close to 0 (VFE) to 1 (EP).

We include the details of the classification datasets in \cref{tab:cladatasets} and several comparisons of $\alpha$ values in \cref{fig:cla_error_table,fig:cla_error_mean_var,fig:cla_nll_table,fig:cla_nll_mean_var}.

\begin{table}[!ht]
\centering
 \begin{tabular}{||c c c c||} 
 \hline
 Dataset & N train/test & D & N positive/negative\\ [0.5ex] 
 \hline\hline
 	australian & 621/69 & 15 & 222/468\\
	breast & 614/68 & 11 & 239/443\\
	crabs & 180/20 & 7 & 100/100\\
	iono & 315/35 & 35 & 126/224\\
	pima & 690/77 & 9 & 500/267\\
	sonar & 186/21 & 61 & 111/96\\[1ex] 
 \hline
\end{tabular}
\caption{Classification datasets}
\label{tab:cladatasets}
\end{table}

\begin{landscape}

\begin{figure}[!ht]
    \centering
    \includegraphics[trim={1cm 1cm 1cm 1cm},clip,width=\linewidth]{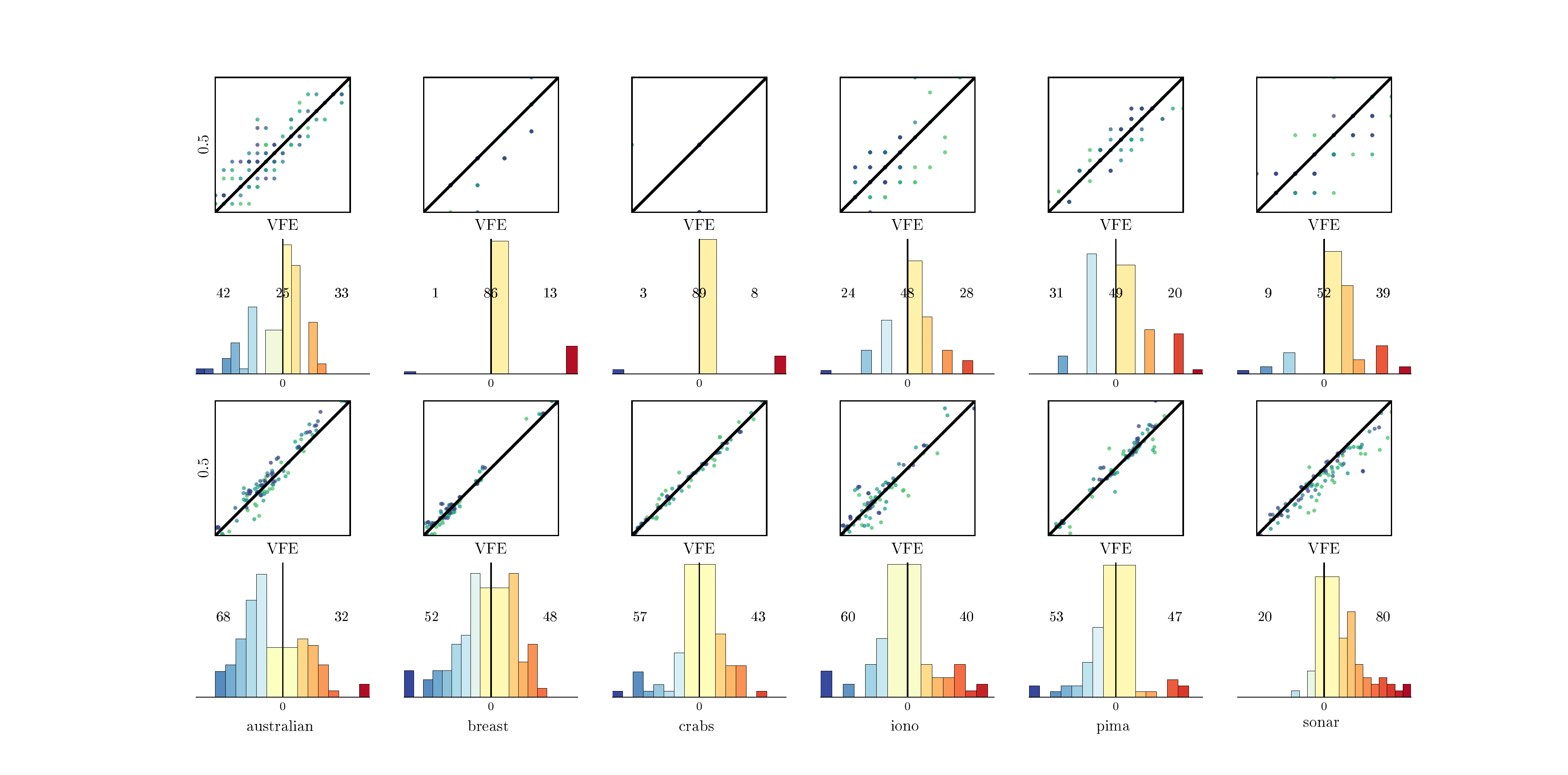}
	\caption{A comparison between Power-EP with $\alpha=0.5$ and VFE on several classification datasets, on two metrics: classification error (top two rows) and NLL (bottom two rows). See \cref{fig:reg_nll_subset0} for more details about the plots.\label{fig:cla_nll_subset0}}
\end{figure}

\begin{figure}[!ht]
    \centering
    \includegraphics[trim={1cm 1cm 1cm 1cm},clip,width=\linewidth]{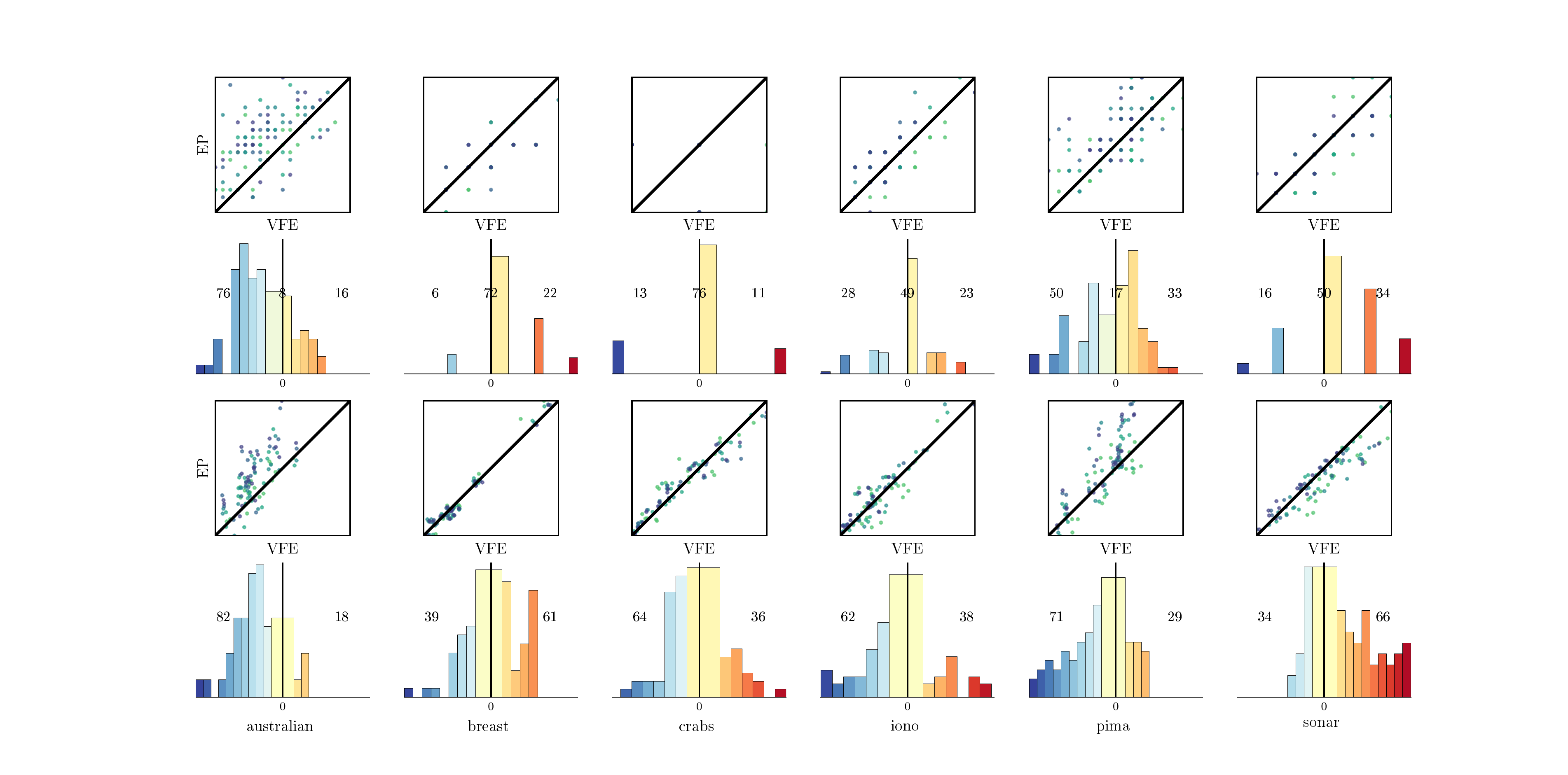}
	\caption{A comparison between EP and VFE on several classification datasets, on two metrics: classification error (top two rows) and NLL (bottom two rows). See \cref{fig:reg_nll_subset0} for more details about the plots.\label{fig:cla_nll_subset1}}
\end{figure}

\begin{figure}[!ht]
    \centering
    \includegraphics[trim={1cm 1cm 1cm 1cm},clip,width=\linewidth]{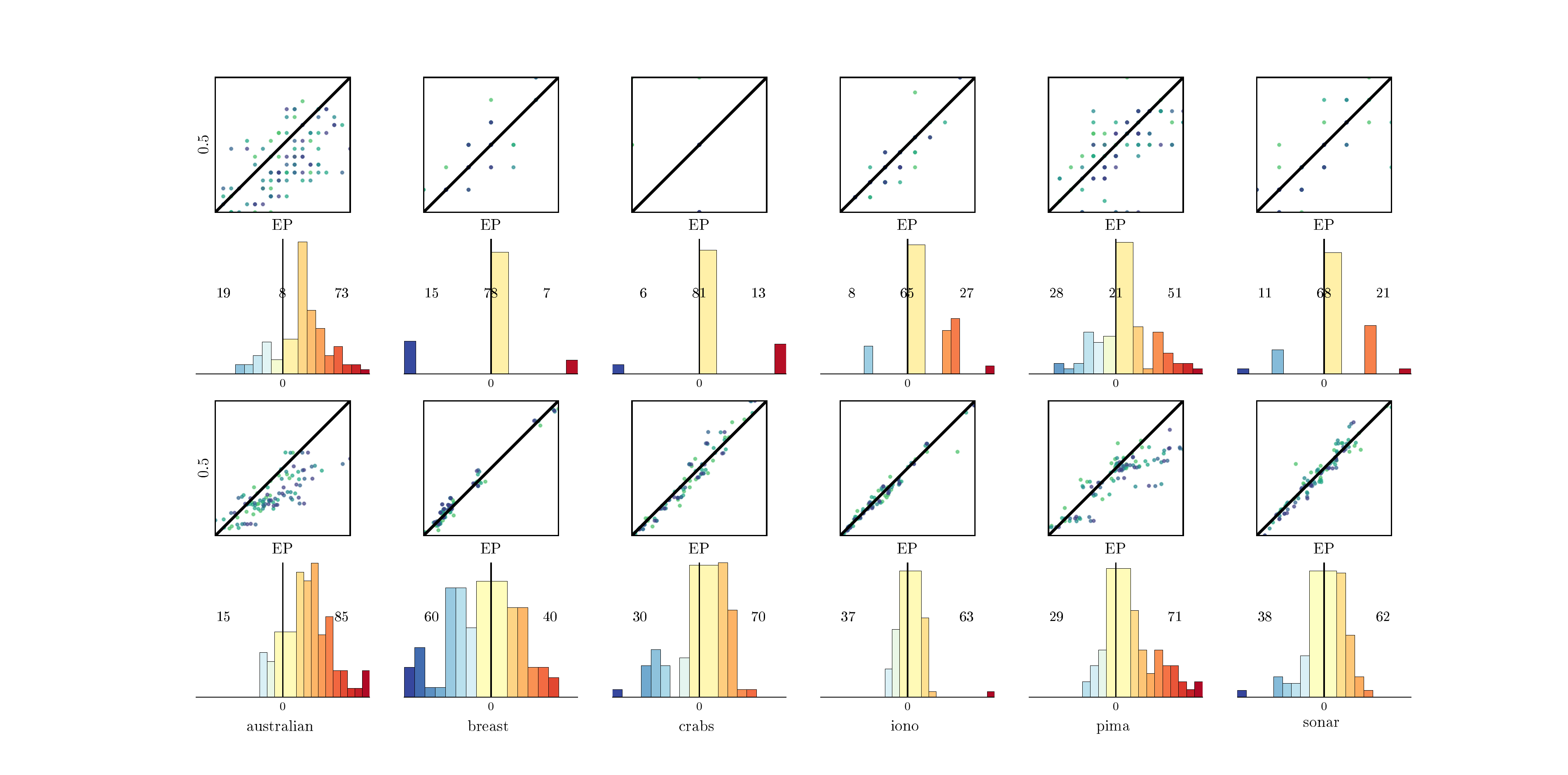}
	\caption{A comparison between Power-EP with $\alpha=0.5$ and EP on several classification datasets, on two metrics: classification error (top two rows) and NLL (bottom two rows). See \cref{fig:reg_nll_subset0} for more details about the plots.\label{fig:cla_nll_subset2}}
\end{figure}

\begin{figure}[!ht]
\centering
\includegraphics[trim={2cm 2cm 2cm 2cm},clip,width=1.6\textwidth]{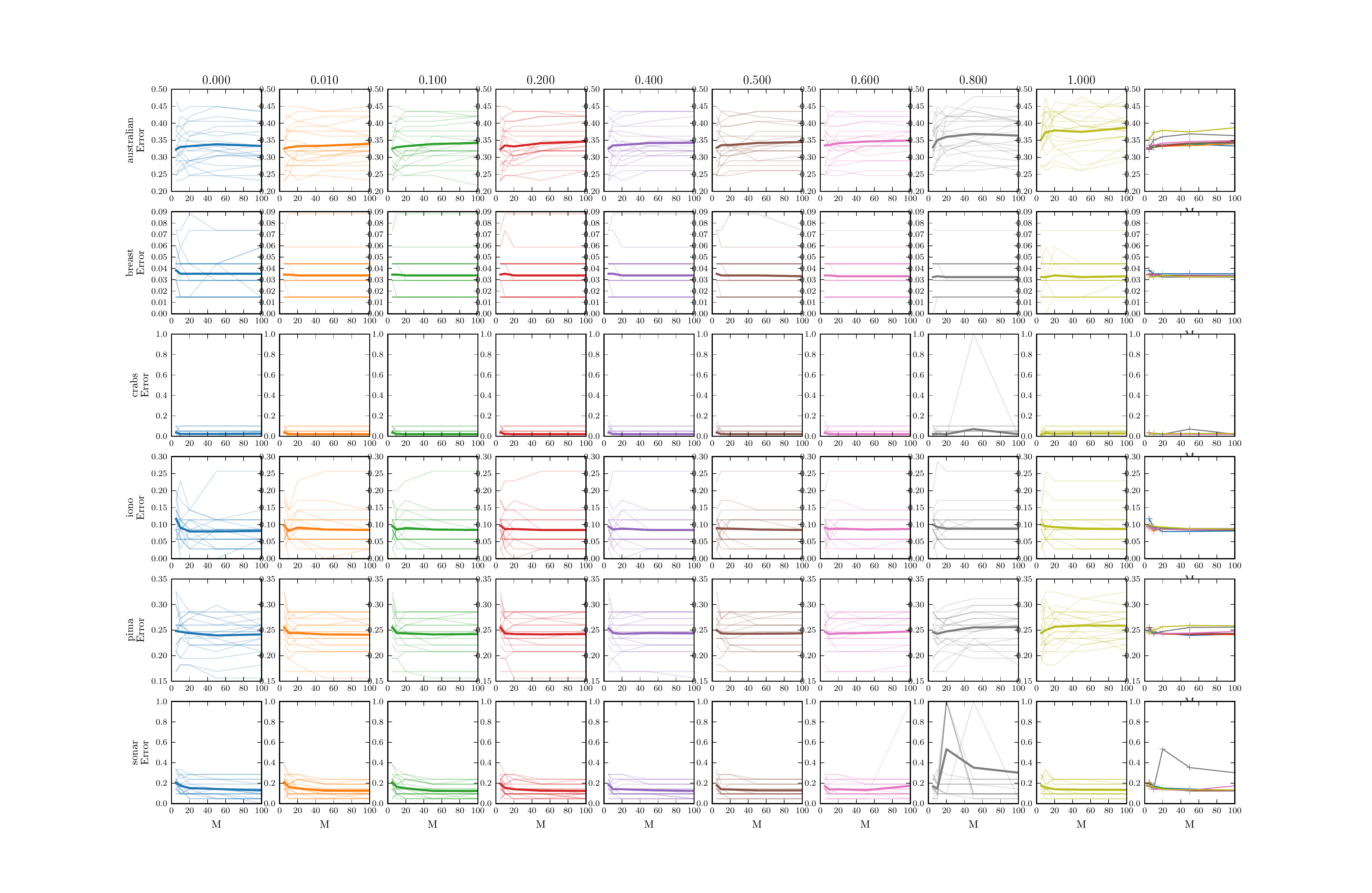}
\caption{Results on real-world classification problems: Classification error rate on the test set for different datasets, various values of $\alpha$ and various number of pseudo-points $M$. Each trace is for one split, bold line is the mean. The rightmost figures show the mean for various $\alpha$ for comparison. Lower is better.\label{fig:cla_error_table}}
\end{figure}

\begin{figure}[!ht]
\centering
\includegraphics[trim={2cm 2cm 2cm 2cm},clip,width=1.6\textwidth]{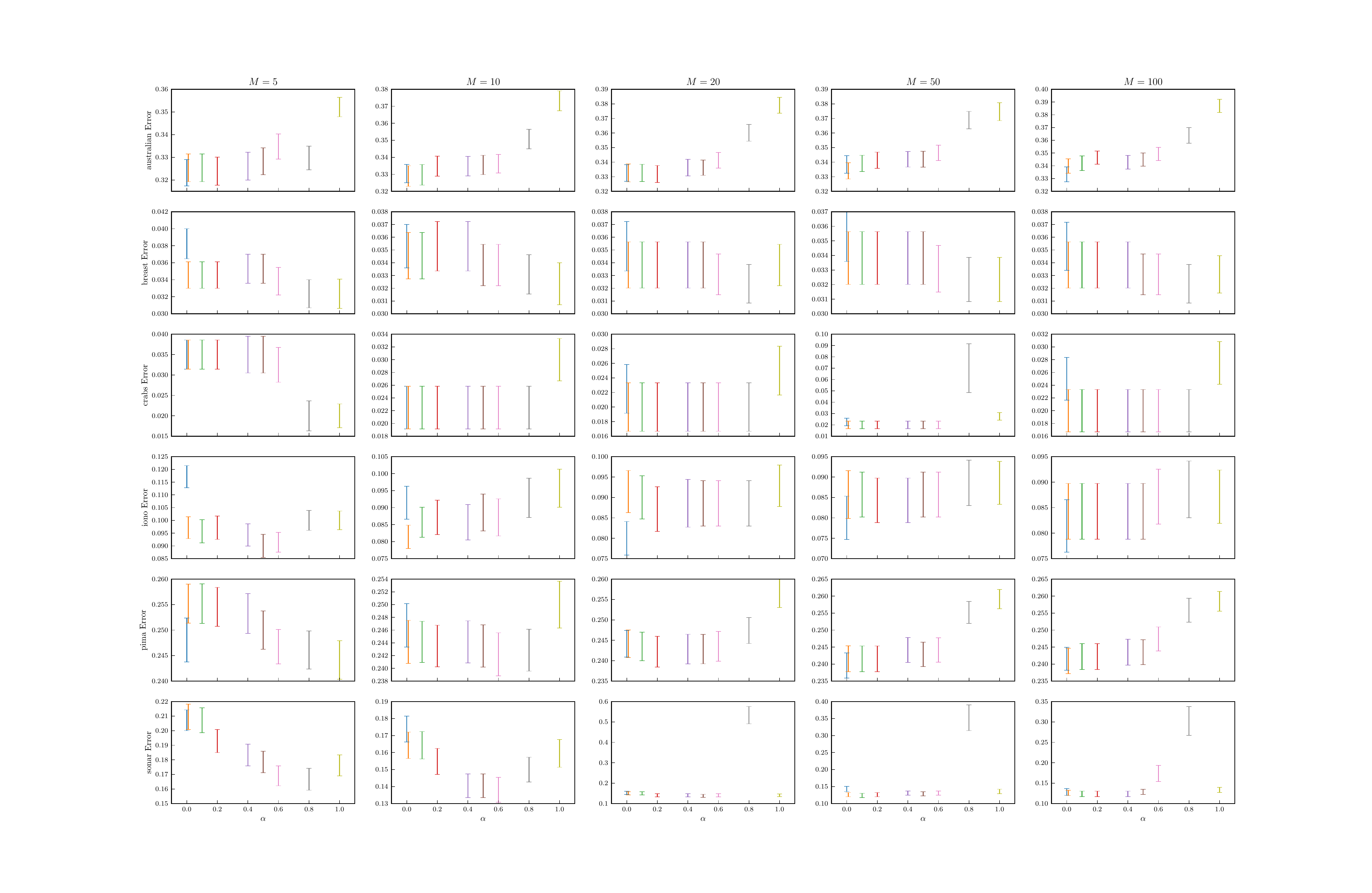}
\caption{Results on real-world classification problems: Classification error rate on the test set for different datasets, various values of $\alpha$ and various number of pseudo-points $M$, averaged over 20 splits. Lower is better.\label{fig:cla_error_mean_var}}
\end{figure}

\begin{figure}[!ht]
\centering
\includegraphics[trim={2cm 2cm 2cm 2cm},clip,width=1.6\textwidth]{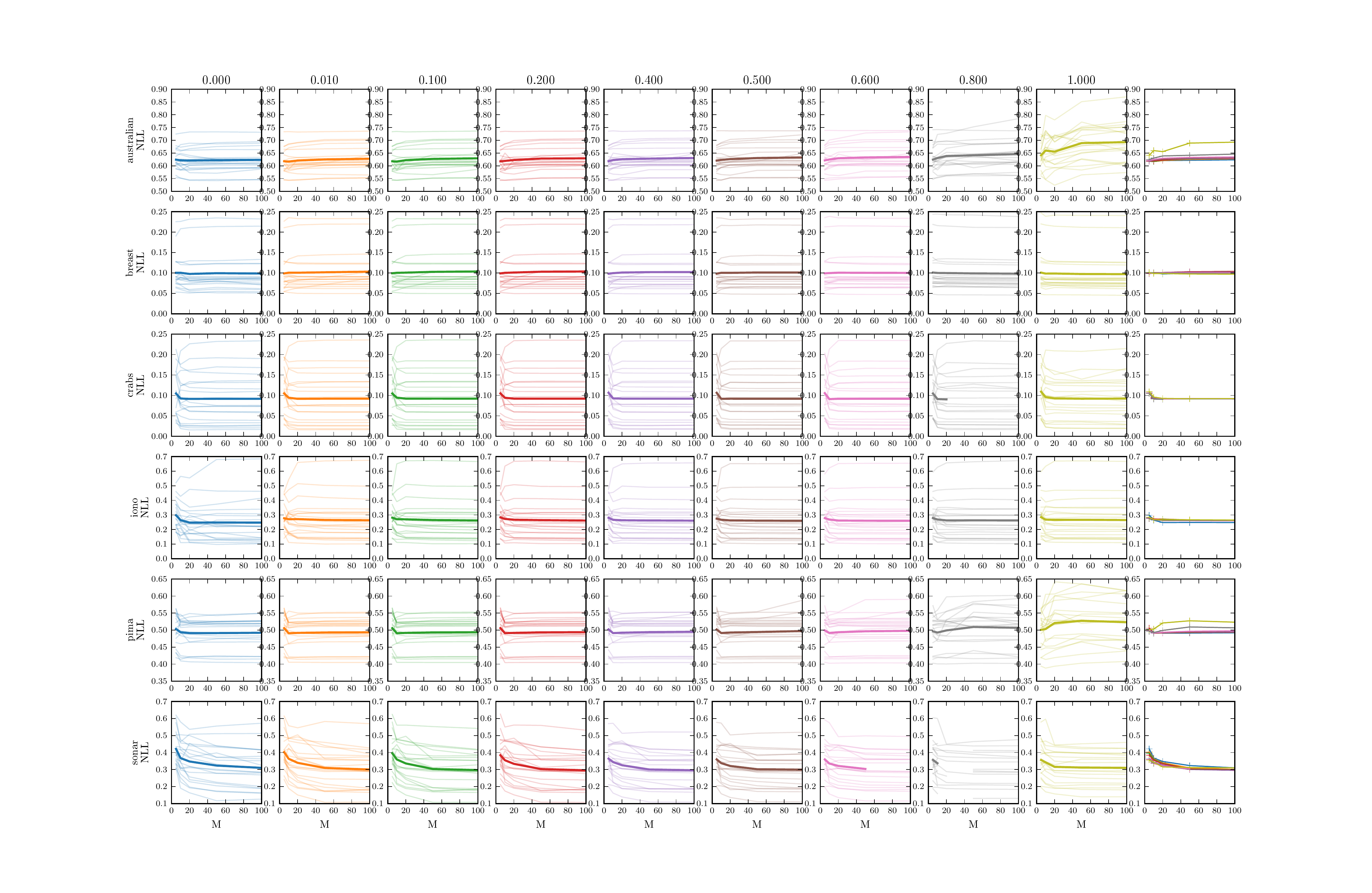}
\caption{Results on real-world classification problems: Average negative log-likelihood on the test set for different datasets, various values of $\alpha$ and various number of pseudo-points $M$. Each trace is for one split, bold line is the mean. The rightmost figures show the mean for various $\alpha$ for comparison. Lower is better.\label{fig:cla_nll_table}}
\end{figure}

\begin{figure}[!ht]
\centering
\includegraphics[trim={2cm 2cm 2cm 2cm},clip,width=1.6\textwidth]{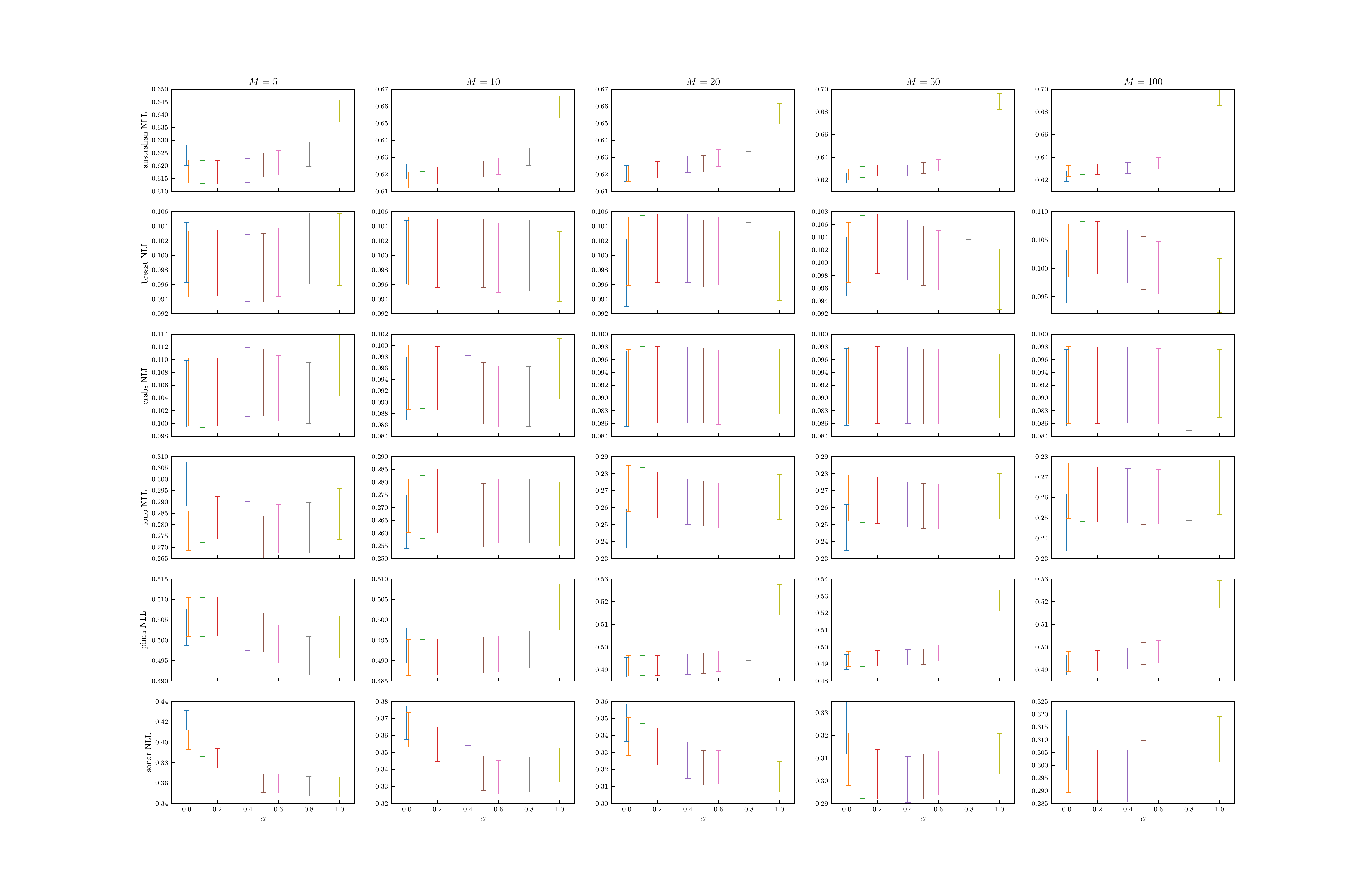}
\caption{Results on real-world classification problems: Average negative log-likelihood on the test set for different datasets, various values of $\alpha$ and various number of pseudo-points $M$, averaged over 20 splits. Lower is better.\label{fig:cla_nll_mean_var}}
\end{figure}
\end{landscape}

\subsection{Binary classification on even/odd MNIST digits}
\begin{figure*}[!ht]
	\centering
	\begin{minipage}[b]{.48\textwidth}
		\includegraphics[width=\textwidth]{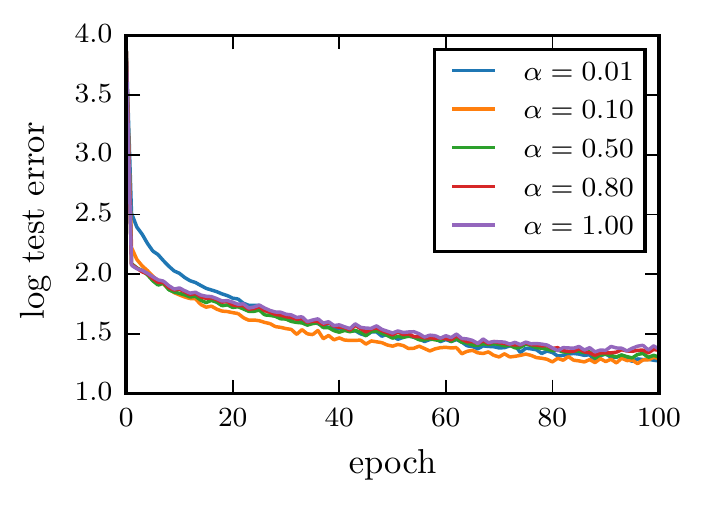}
	\end{minipage}\quad
	\begin{minipage}[b]{.48\textwidth}
		\includegraphics[width=\textwidth]{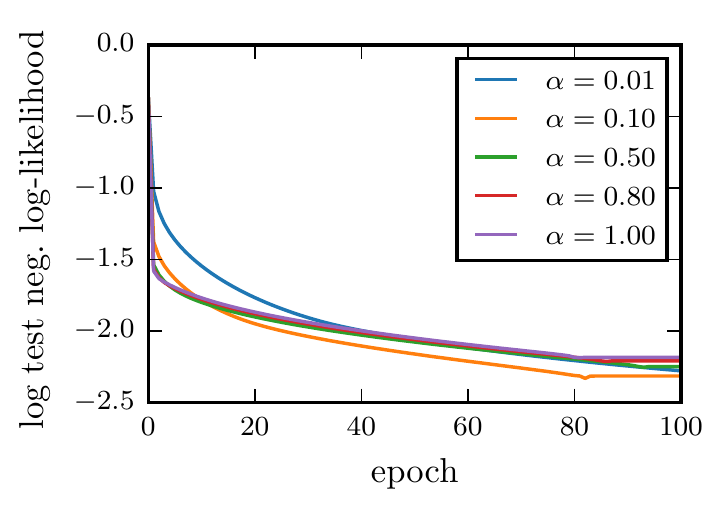}
	\end{minipage}
	\caption{The test error and log-likelihood of the MNIST binary classification task (M=100).\label{fig:mnist100}}
\end{figure*}

\begin{figure*}[!ht]
	\centering
	\begin{minipage}[b]{.48\textwidth}
		\includegraphics[width=\textwidth]{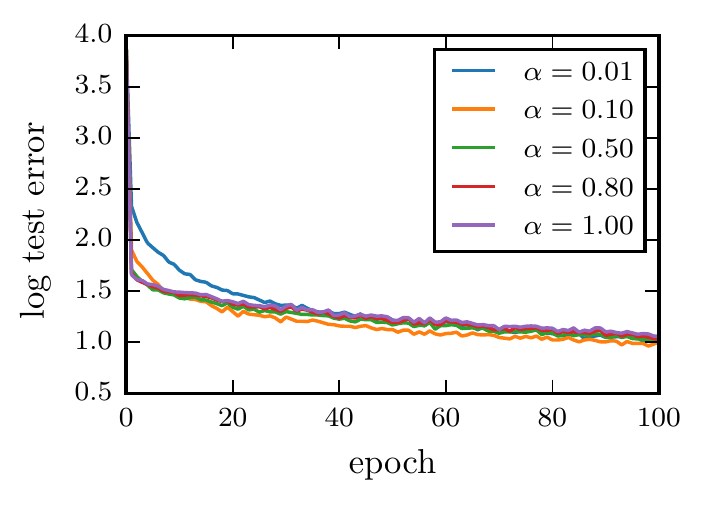}
	\end{minipage}\quad
	\begin{minipage}[b]{.48\textwidth}
		\includegraphics[width=\textwidth]{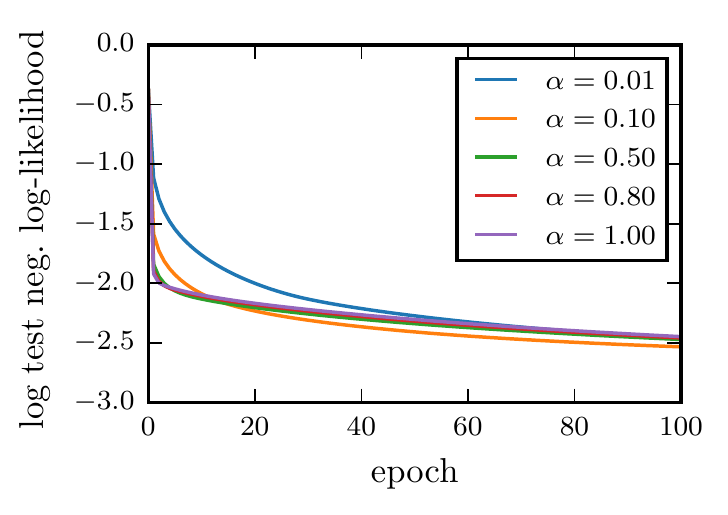}
	\end{minipage}
	\caption{The test error and log-likelihood of the MNIST binary classification task (M=200).\label{fig:mnist200}}
\end{figure*}

\subsection{When $M=N$ and $\alpha=1$, do we recover EP for GPC \citep[sec.~3.6]{RasWil05}?}

\begin{figure}[htbp!]
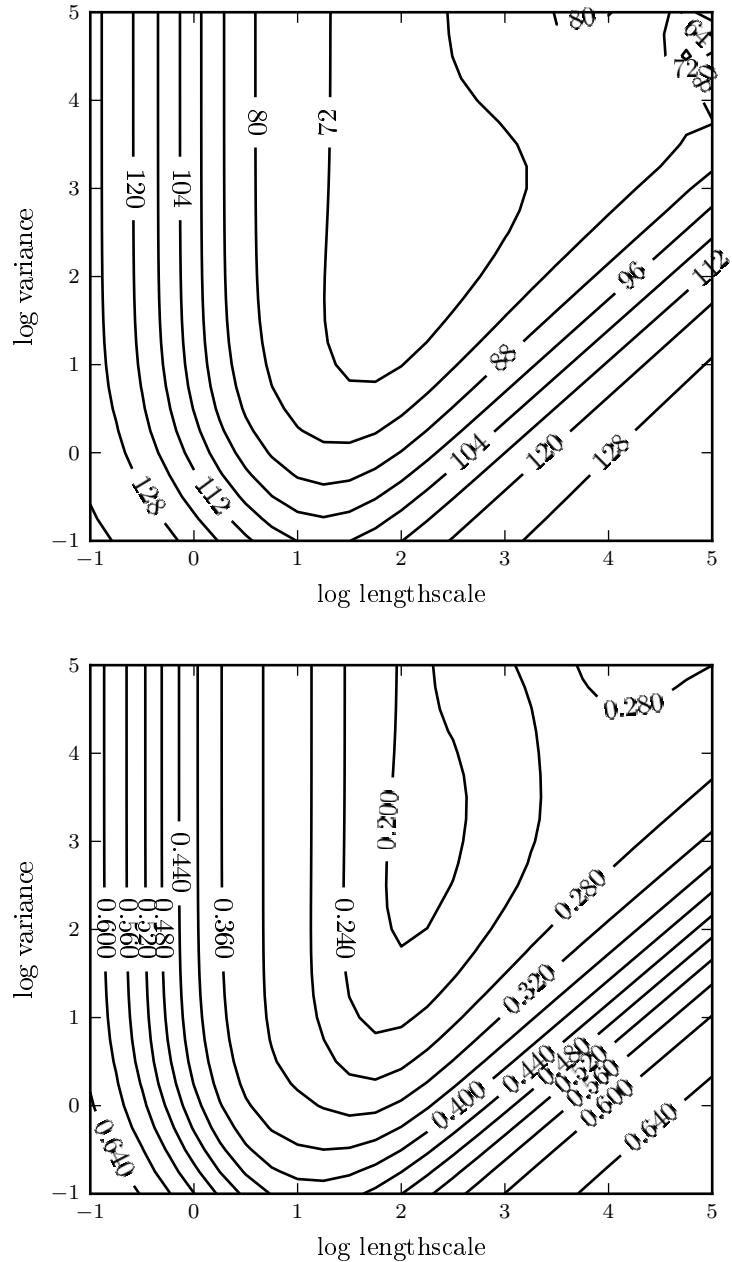

\centering
\begin{tabular}{c}
\includegraphics{{{iono_energy_alpha_1.000}}}\\
\includegraphics{{{iono_test_nll_alpha_1.000}}}
\end{tabular}
\caption{EP energy on the train set [TOP] and the average negative log-likelihood on the test set[BOTTOM] when $M=N$.}
\label{fig:gpc_energy}
\end{figure}

The key difference between the EP method in this manuscript when $M=N$ and the pseudo-inputs and the training inputs are identical, and the standard EP method as described by \citep[sec.~3.6]{RasWil05} is the factor representation. While \cite{RasWil05} used a one dimensional un-normalised Gaussian distribution that touches only {\it one} function value $f_n$ to approximate each exact factor, the approximate factor used in the EP scheme described in the main text touches {\it all} $M$ pseudo-points, hence {\it all} $N$ function values when the pseudo-inputs are placed at the training inputs. However, in practice both methods give virtually identical results. \Cref{fig:gpc_energy} shows the approximate log marginal likelihood and the negative test log-likelihood, given by running the EP procedure described in the main text on the \texttt{ionosphere} dataset. We note that these results are similar to that of the standard EP method \cite[see][]{KusRas05}.

\vskip 0.2in

{
\bibliography{refs}
}

\end{document}